\documentclass{article} 
\usepackage{style/iclr2024_conference,times}
\iclrfinalcopy

\usepackage{url}
\usepackage[utf8]{inputenc} 
\usepackage[T1]{fontenc}    
\usepackage{url}            
\usepackage{booktabs}       
\usepackage{enumitem}
\usepackage{natbib}
\usepackage{amsfonts}       
\usepackage{nicefrac}       
\usepackage{microtype}      
\usepackage{xcolor}         
\usepackage{enumitem}
\usepackage{xspace}
\usepackage{amsthm}
\usepackage{amsmath,amsfonts,bm}
\makeatletter
\newtheorem*{rep@theorem}{\rep@title}
\newcommand{\newreptheorem}[2]{%
\newenvironment{rep#1}[1]{%
 \def\rep@title{#2 \ref{##1}}%
 \begin{rep@theorem}}%
 {\end{rep@theorem}}}
\makeatother

\newtheorem{theorem}{Theorem}
\newtheorem{proposition}{Proposition}

\newreptheorem{theorem}{Theorem}

\definecolor{myred}{RGB}{215,48,39}
\definecolor{mygreen}{RGB}{26,152,80}
\newcommand{\cmark}{\textcolor{mygreen}{\ding{51}}}
\newcommand{\xmark}{\textcolor{myred}{\ding{55}}}
\newcommand{\halfmark}{\textcolor{gray}{\checkmark\kern-1.1ex\raisebox{.7ex}{\rotatebox[origin=c]{125}{--}}}}
\usepackage[framemethod=TikZ]{mdframed}
\mdfdefinestyle{MyFrame}{%
    linecolor=black,
    outerlinewidth=.3pt,
    roundcorner=5pt,
    innertopmargin=1pt, 
    innerbottommargin=1pt, 
    innerrightmargin=1pt,
    innerleftmargin=1pt,
    backgroundcolor=black!0!white}

\mdfdefinestyle{MyFrame2}{%
    linecolor=white,
    outerlinewidth=1pt,
    roundcorner=1pt,
    innertopmargin=0,
    innerbottommargin=0,
    innerrightmargin=7pt,
    innerleftmargin=7pt,
    backgroundcolor=black!3!white}

\mdfdefinestyle{MyFrameEq}{%
    linecolor=white,
    outerlinewidth=0pt,
    roundcorner=0pt,
    innertopmargin=0pt,
    innerbottommargin=0pt,
    innerrightmargin=7pt,
    innerleftmargin=7pt,
    backgroundcolor=black!3!white}

\newcommand{\RNum}[1]{\uppercase\expandafter{\romannumeral #1\relax}}

\newcommand{\R}{\mathcal{R}}

\newcommand{\vertiii}[1]{{\left\vert\kern-0.25ex\left\vert\kern-0.25ex\left\vert #1 
    \right\vert\kern-0.25ex\right\vert\kern-0.25ex\right\vert}}
\newcommand{\vertiiii}[1]{{\vert\kern-0.25ex\vert\kern-0.25ex\vert #1 
    \vert\kern-0.25ex\vert\kern-0.25ex\vert}}

\usepackage{mathtools}

\newcommand{\xhdr}[1]{{\noindent\bfseries #1}.}
\newcommand{\cut}[1]{}

\newcommand{\angstrom}{\textup{\AA}}

\newcommand{\removelatexerror}{\let\@latex@error\@gobble}

\newcommand\rev[0]{}

\def\eqref#1{Eq.~\ref{#1}}

\def\1{\bm{1}}

\def\eps{{\epsilon}}

\DeclareMathAlphabet{\mathsfit}{\encodingdefault}{\sfdefault}{m}{sl}
\SetMathAlphabet{\mathsfit}{bold}{\encodingdefault}{\sfdefault}{bx}{n}

\def\gC{{\mathcal{C}}}

\def\gL{{\mathcal{L}}}
\def\gM{{\mathcal{M}}}
\def\gN{{\mathcal{N}}}

\def\gQ{{\mathcal{Q}}}

\def\gT{{\mathcal{T}}}
\def\gU{{\mathcal{U}}}

\def\gW{{\mathcal{W}}}

\def\sP{{\mathbb{P}}}

\def\R{{\mathbb{R}}}

\newcommand{\sethree}{\mathrm{SE(3)}}

\newcommand{\sothree}{\mathrm{SO(3)}}
\newcommand{\sethreen}{\sethree^{\scriptscriptstyle N}}
\newcommand{\sethreenzero}{\sethree^{\scriptscriptstyle N}_{\scriptscriptstyle 0}}
\newcommand{\sotwo}{\mathrm{SO(2)}}
\newcommand{\sothreelie}{\mathfrak{so}(3)}
\newcommand{\sethreelie}{\mathfrak{se}(3)}
\newcommand{\igso}{\mathcal{IG}_\mathrm{SO(3)}}

\newcommand{\cfm}[1]{$\sethree$-CFM\xspace}
\newcommand{\otcfm}[1]{$\sethree$-OT-CFM\xspace}
\newcommand{\sfm}[1]{$\sethree$-SFM\xspace}
\newcommand{\foldflow}[1]{\textsc{FoldFlow}\xspace}
\newcommand{\foldflowbase}[1]{\textsc{FoldFlow-Base}\xspace}
\newcommand{\foldflowot}[1]{\textsc{FoldFlow-OT}\xspace}
\newcommand{\foldflowsfm}[1]{\textsc{FoldFlow-SFM}\xspace}

\def\dt{\,\mathrm{d}t}

\newcommand{\ddd}{\mathrm{d}}
\usepackage{todonotes}
\usepackage{amssymb,fge}
\usepackage{thm-restate}
\usepackage{wrapfig}
\usepackage{bbm}
\usepackage{mathrsfs}
\usepackage{soul}
\usepackage{array}
\usepackage{multirow}
\usepackage{algorithm}
\usepackage[commentColor=black,beginLComment=/*~, endLComment=~*/]{algpseudocodex}
\usepackage{subcaption}
\usepackage{pifont}

\usepackage[final,pagebackref=true]{hyperref} 
\usepackage{cleveref}
\renewcommand*{\backrefalt}[4]{%
    \ifcase #1 \footnotesize{(Not cited.)}%
    \or        \footnotesize{(Cited on page~#2)}%
    \else      \footnotesize{(Cited on pages~#2)}%
    \fi}
\newcolumntype{P}[1]{>{\centering\arraybackslash}p{#1}}

\hypersetup{
	colorlinks=true,       
	linkcolor=blue,        
	citecolor=blue,        
	filecolor=magenta,     
	urlcolor=blue         
}

\def\setstretch#1{\renewcommand{\baselinestretch}{#1}}
\makeatletter
\providecommand{\section}{}
\renewcommand{\section}{%
  \@startsection{section}{1}{\z@}%
                {-1.0ex \@plus -0.5ex \@minus -0.2ex}%
                { 1.0ex \@plus  0.3ex \@minus  0.2ex}%
                {\large\sc\raggedright}%
}
\providecommand{\subsection}{}
\renewcommand{\subsection}{%
  \@startsection{subsection}{2}{\z@}%
                {-0.75ex \@plus -0.5ex \@minus -0.2ex}%
                { 0.75ex \@plus  0.2ex}%
                {\normalsize\sc\raggedright}%
}
\providecommand{\subsubsection}{}
\renewcommand{\subsubsection}{%
  \@startsection{subsubsection}{3}{\z@}%
                {-0.5ex \@plus -0.5ex \@minus -0.2ex}%
                { 0.5ex \@plus  0.2ex}%
                {\normalsize\sc\raggedright}%
}
\providecommand{\paragraph}{}
\renewcommand{\paragraph}{%
  \@startsection{paragraph}{4}{\z@}%
                {0.3ex \@plus 0.2ex \@minus 0.2ex}%
                {-1em}%
                {\normalsize\bf}%
}
\makeatother

\title{$\sethree$ Stochastic Flow Matching for Protein Backbone Generation}

\author{%
 Avishek (Joey) Bose$^{1,2,3}$\thanks{Equal Contribution. Corresponding authors: \{\texttt{joey.bose,tara,alex\}@dreamfold.ai}},
Tara Akhound-Sadegh$^{1,2,3}$\footnotemark[1],
Guillaume Huguet$^{4,2,3}$,
Kilian Fatras$^{1,2,3}$,\\
\textbf{Jarrid Rector-Brooks$^{4,2,3}$,
Cheng-Hao Liu$^{1,2,3}$,
Andrei Cristian Nica$^{3}$,
Maksym Korablyov$^{3}$,}\\
\textbf{Michael Bronstein$^{5,3}$,
Alexander Tong$^{4,2,3}$}\\
 $^1$McGill University, $^2$Mila, $^3$Dreamfold, $^4$Université de Montréal, $^5$University of Oxford
}

\begin{document}

\maketitle

\begin{abstract}
\looseness=-1
The computational design of novel protein structures has the potential to impact numerous scientific disciplines greatly. Toward this goal, we introduce \foldflow, a series of novel generative models of increasing modeling power based on the flow-matching paradigm over $3\mathrm{D}$ rigid motions---i.e. the group $\sethree$---enabling accurate modeling of protein backbones. We first introduce \foldflowbase, a simulation-free approach to learning deterministic continuous-time dynamics and matching invariant target distributions on $\sethree$. We next accelerate training by incorporating Riemannian optimal transport to create \foldflowot, leading to the construction of both more simple and stable flows. Finally, we design \foldflowsfm, coupling both Riemannian OT and simulation-free training to learn stochastic continuous-time dynamics over $\sethree$. Our family of \foldflow, generative models offers several key advantages over previous approaches to the generative modeling of proteins: they are more stable and faster to train than diffusion-based approaches, and our models enjoy the ability to map any invariant source distribution to any invariant target distribution over $\sethree$. Empirically, we validate \foldflow, on protein backbone generation of up to $300$ amino acids leading to 
high-quality designable, diverse, and novel samples. 

\end{abstract}

\section{Introduction}
\vspace{-5pt}
\label{sec:introduction}


\looseness=-1
Proteins are one of the basic building blocks of life. Their complex geometric structure enables specific inter-molecular interactions that allow for crucial functions within organisms, such as acting as catalysts in chemical reactions, transporters for molecules, and providing immune responses. 
Normally, such functions arise as a result of evolution. 
With the emergence of computational techniques, it has become possible to rationally design novel proteins with desired structures that program their functions. Such methods are now seen as the future of drug design and can lead to solutions to long-standing global health challenges. 
Some recent examples include 
rationally designed 
protein binders for receptors related to influenza~\citep{strauch2017computational}, COVID-19~\citep{cao2021denovocovid,gainza2023novo}, and cancer~\citep{silva2019novo}. 
%

\cut{
Proteins, the workhorses of the cell, are large biomolecules constructed from amino acid sequences intricately folded into unique three-dimensional structures. The complex geometric structure of proteins enables specific intermolecular interactions that allow for crucial functions within organisms, such as acting as catalysts in chemical reactions, transporters for molecules, structural supports, and even signalling molecules themselves. Designing the structure of proteins enables the programming of their functions, which can lead to solutions to long-standing global health challenges. For instance, the novel design of protein shapes complementary to given target proteins has produced strong binders for receptors related to influenza~\citep{strauch2017computational}, COVID-19~\citep{cao2021denovocovid}, and cancer~\citep{silva2019novo}. Consequently, there is a pressing need to develop new computational approaches to protein engineering that not only have the ability to navigate the large search spaces that are common in proteins but also respect the exact laws of nature that govern protein dynamics in the form of symmetries. 
}
\begin{figure}[htb]
    \vspace{-5pt}
    \centering
    \begin{minipage}{0.6\textwidth}
    \includegraphics[width=1\linewidth]{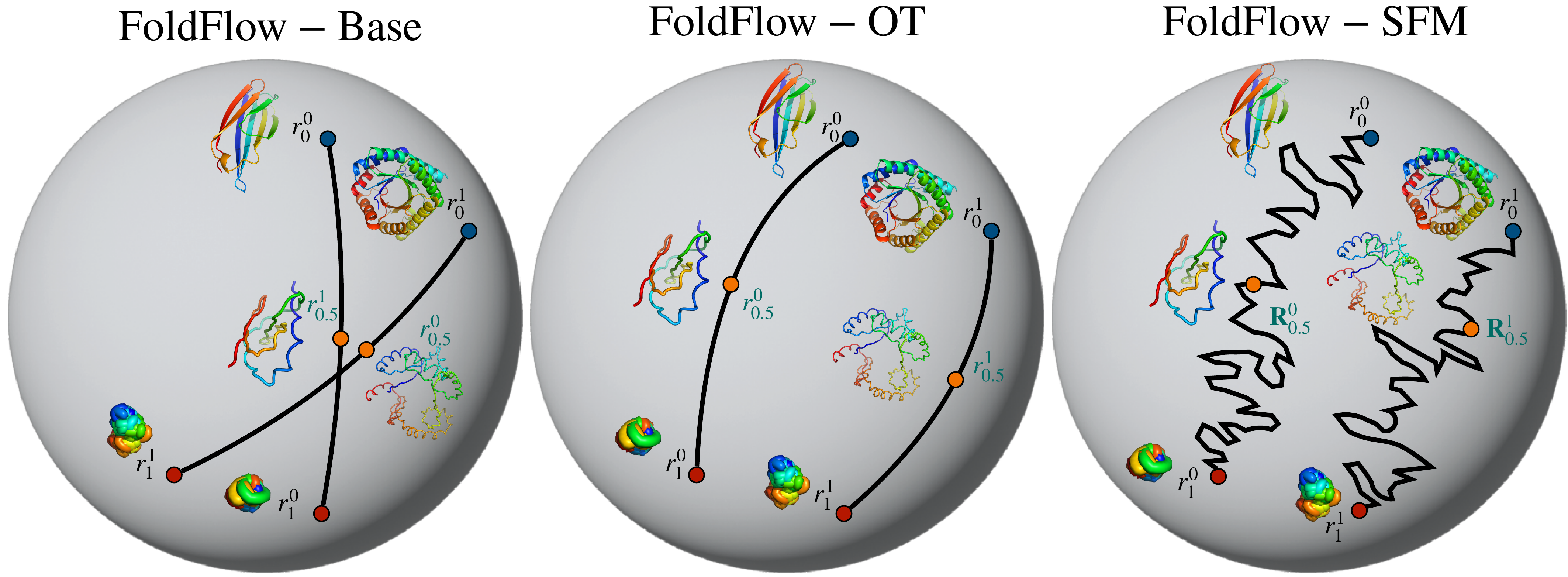}
    
\end{minipage}
\begin{minipage}{0.39\textwidth}
\resizebox{\textwidth}{!}{
\begin{tabular}{l|P{0.7cm}P{0.3cm}P{1cm}P{1cm}}
\toprule
Method  & Any source & OT & Stochastic & Score free \\
\midrule
RFDiffusion    & \xmark  & \xmark    & \cmark & \xmark \\
FrameDiff      & \xmark  & \xmark    & \cmark & \xmark \\
\foldflowbase, & \cmark & \xmark     & \xmark & \cmark  \\
\foldflowot,   & \cmark & \cmark     & \xmark & \cmark  \\
\foldflowsfm,  & \cmark & \cmark$^*$ & \cmark & \cmark  \\
\bottomrule
\end{tabular}
}
\end{minipage}
    \caption{\small \textbf{Left:} Conditional probability paths learned by \foldflowbase,(left), \foldflowot, (mid), and \foldflowsfm,(right). We visualize the rotation trajectory of a single residue by the action of $\sothree$ on its homogenous space $\mathbb{S}^2$. \textbf{Right:} Table with the properties of each model: whether they can map from a general source distribution, perform optimal transport, are stochastic, or require the score of the $\igso$ density.}
    \label{fig:simulation_on_sphere}
    \vspace{-5pt}
\end{figure}


\looseness=-1
In protein engineering, the term \emph{de novo} design refers to a setting when a new protein is designed to satisfy pre-specified structural and functional properties~\citep{huang2016coming}. 
Chemically, a protein is a sequence of amino acids ({\em residues}) linked into a chain that folds into a complex 3D structure under the influence of electrostatic forces. 
The protein backbone can be seen as $N$ rigid bodies (corresponding to $N$ residues) that contain four heavy atoms $\text{N} - \text{C}_{\alpha} - \text{C} - \text{O}$. Mathematically, each residue can be associated with a frame that 
adheres to the symmetries of orientation-preserving rigid transformations (3D rotations and translations), forming the {\em special Euclidean group} $\sethree$~\citep{jumper2021highly}; the entire protein backbone is described by the 
group product $\sethreen$. 
The problem of protein design can be formulated as sampling from the distribution over this group, which is a perspective used in our paper. 
Recently, generative models have been generalized to Riemannian manifolds~\citep{mathieu2020riemannian,de2022riemannian}. However, they are not purpose-built to exploit the rich geometric structure of $\sethreen$. Furthermore, several approaches require numerically expensive steps like simulating a Stochastic Differential Equation (SDE) during training or using the Riemannian divergence in the objective~\citep{huang2022riemannian,leach2022denoising,ben2022matching}. 


\cut{
Proteins are the basic building blocks of life. The complex geometric structure imparted by the
sequence of amino acids, within a protein, is chiefly responsible for critical tasks in living organisms.
Important examples of protein functions include acting as enzymes that catalyze chemical reactions,
regulation of physiological processes, and developing antibodies in immune responses to fight
pathogens. Indeed, the fabrication of novel proteins has the potential to enable significant progress
in biological drug design with the goal of addressing long-standing global health challenges such
as cancer (Silva et al., 2019), influenza (Strauch et al., 2017), diabetes, and COVID-19 (Cao et al.,
2022). Consequently, there is a pressing need to develop new computational approaches to protein
engineering that not only have the ability to navigate the large search spaces that are common in
proteins but also respect the exact laws of nature that govern molecular dynamics.
}

\xhdr{Our approach}
\looseness=-1
We introduce \foldflow,, a family of continuous normalizing flows (CNFs) tailored for distributions on $\sethreen$ (\cref{fig:simulation_on_sphere}). We use the framework of Conditional Flow Matching (CFM), a \emph{simulation-free} approach to learning CNFs by directly regressing time-dependent vector fields that generate probability paths ~\citep{lipman_flow_2022,tong2023improving}. 
In particular, we introduce three new CFM-based models that learn $\sethreen$-invariant distributions to generate protein backbones. In contrast to the previous  $\sethree$ diffusion approach of \citet{yim2023se}, our \foldflow, is able to start from an informative prior. This enables new applications of generative models for protein design such as equilibrium conformation generation~\citep{zheng2023towards}.

\looseness=-1
\xhdr{Main contributions} Our first model \foldflowbase, extends the Riemannian flow matching approach~\citep{chen2023riemannian}  by introducing a closed-form expression of the ground truth conditional vector field for $\sothree$ needed in the loss computation---thus greatly increasing speed and stability of training. Next, in \foldflowot,, we accelerate the training of our base model by constructing shorter and simpler flows using Riemannian Optimal Transport (OT) by proving the existence of a Monge map on $\sethreen$. Finally, we present our most complex simulation-free model, \foldflowsfm,, which learns a stochastic bridge on $\sethreen$. Empirically, we validate our proposed models by learning to generate protein backbones of up to $300$ residues. We observe that all \foldflow, models outperform the current SOTA non-pretrained diffusion model in FrameDiff~\citep{yim2023se} for \emph{in-silico} designability with \foldflowot, being the most designable. Moreover, for novelty \foldflowsfm, is competitive with the current gold-standard RFDiffusion~\citep{watson_novo_2023} with a fraction of the compute and data resources. We highlight the importance of novel and designable proteins as a key goal in AI-powered drug discovery where useful drug candidates are necessarily beyond the available training set~\citep{schneider2018automating, schneider2020rethinking, marchand2022computational}. Finally, we show the utility of \foldflow, on equilibrium conformation generation by learning to simulate molecular dynamics trajectories starting from a reference empirical distribution in comparison to an uninformed prior. Our code can be found at \url{https://github.com/DreamFold/FoldFlow}.



\cut{
\begin{itemize}[noitemsep,topsep=0pt,parsep=0pt,partopsep=0pt,label={\large\textbullet},leftmargin=*]
    \item \looseness=-1 We introduce three new simulation-free generative models in \foldflowbase,, \foldflowot,, and \foldflowsfm, that map between any source and target distribution on $\sethreen$. 
    \item We prove the existence of a Monge map on an invariant subgroup $\sethreenzero$ and build the Riemannian OT problem on $\sethreenzero$ which power \foldflowot,, resulting in more stable flows.
    \item We prove \foldflowsfm, is a valid flow and use it to construct a simulation-free SDE on $\sethree$.
     \item We provide several new numerical techniques, such as a closed-form expression of the target conditional vector field for $\sothree$, which leads to fast, stable, and efficient training on $\sothree$. 
    \item Empirically, we validate our proposed models by learning $\sethreen$-invariant distributions for protein backbones. All three of our models outperform the current SOTA non-pretrained model in FrameDiff-Improved~\citep{yim2023se} in \emph{in-silico} designability, diversity, and novelty. 
 \end{itemize}
 }
\vspace{-5pt}
\section{Background and preliminaries}
\vspace{-5pt}
\label{sec:background}
\subsection{Riemannian manifolds and Lie groups}
\vspace{-5pt}
\label{sec:riemannian_manifolds}
\looseness=-1
\xhdr{Riemannian manifolds} Informally, an $n$-dimensional {\em manifold} $\gM$ is a topological space locally equivalent (homeomorphic) to $\mathbb{R}^n$. 
This implies that one has the notion of `neighbourhood' but not of `distance' or `angle' on $\gM$. The manifold is said to be {\em smooth} if it additionally has a $C^\infty$ differential structure. 
At every point $x\in \gM$, one can attach a {\em tangent space} $\gT_x$. The disjoint union of tangent spaces forms the {\em tangent bundle}. 
A {\em Riemannian manifold}\footnote{We tacitly assume $\gM$ to be orientable, connected, and complete and admit a volume form denoted as $dx$. } $(\gM, g)$ is additionally equipped with an inner product ({\em Riemannian metric}) $g_x: \gT_x\gM \times \gT_x\gM\rightarrow\R$ on the
tangent space $\gT_x\gM$ at each $x\in\gM$. The Riemannian metric $g$ allows to define key geometric quantities on $\gM$ such as distances, volumes, angles, and length minimizing curves ({\em geodesics}). 
We consider functions defined on $\gM$ and the tangent bundle, referred to as {\em scalar-} and {\em vector fields}, respectively. The {\em Riemannian gradient} is an operator $\nabla_g: C^\infty(\gM) \rightarrow \mathfrak{X}(\gM)$ between the respective functional spaces. Given a smooth scalar field $f\in C^\infty(\gM)$, its gradient $\nabla_g f \in \mathfrak{X}(\gM)$ is the local direction of its steepest change.  
%
%
%


\xhdr{Lie groups}
\looseness=-1
A {\em Lie group} is a group that is also a differentiable manifold, in which the group operations $\circ: G \times G \to G$ of multiplication and inversion are smooth maps.
It has a left action $L_h: G \to G$ defined by $x \mapsto h \circ x$ that is a topological isomorphism and whom the derivative is also an isomorphism between the tangent spaces on $G$. Since a group has an identity element, its tangent space is of special interest and is known as the {\em Lie algebra} $\mathfrak{G}$. The Lie algebra is a vector space with an associated bilinear operation called the {\em Lie bracket} that is anticommutative and satisfies the Jacobi identity. Lie algebras elements can be mapped to the Lie group via the {\em exponential map} $\exp: \mathfrak{G} \rightarrow G$ which has an inverse called the logarithmic map $\log: G \to \mathfrak{G}$. 
For matrix Lie groups where the group action is the matrix multiplication, the $\exp$ and $\log$ maps correspond to the matrix exponential and matrix logarithm. 
The orientation-preserving rigid motions form the matrix Lie group $\sethree \cong \sothree \ltimes (\R^3, +)$, a semidirect product of rotations and translations (see ~\S\ref{app:sothree}~\S\ref{app:sethree}, and~\citet{hall2013lie} for details). 


\subsection{Flow matching on Riemannian manifolds}
\vspace{-5pt}
Analogous to Euclidean spaces, probability densities can be defined on Riemannian manifolds 
as continuous non-negative functions $\rho: \gM \to \R_{+}$ that integrate to $\int_{\gM} \rho(x) dx = 1$. 


\xhdr{Probability paths on Riemannian manifolds}
Let $\sP(\gM)$ be the space of probability distributions on $\gM$. A \emph{probability path} $\rho_t: [0, 1] \to \sP(\gM)$  is an interpolation in probability space between two distributions $\rho_0, \rho_1 \in \sP(\gM)$ indexed by a continuous parameter $t$. A one-parameter diffeomorphism $\psi_t: \gM \to \gM$ is known as a \emph{flow} on $\gM$ and is defined as the solution of the following ordinary differential equation (ODE): $ \frac{d}{dt} \psi_t(x) = u_t \left(\psi_t(x) \right)$, with initial conditions $\psi_0(x) = x$, for $u:[0,1]\times\gM\to\gM$ a time-dependent smooth vector field. We say the flow $\psi_t$ generates $\rho_t$ if it pushes forward $\rho_0$ to $\rho_1$ by following the time-dependent vector field $u_t$---i.e. $\rho_t=[\psi_t]_\#(\rho_0)$. As $\psi_t$ is a diffeomorphism, $\rho_t$ verifies the famous \textit{continuity equation} and the density can be calculated using the instantaneous change of variables formula for Riemannian manifolds~\citep{mathieu2020riemannian}.

\xhdr{Riemannian flow matching} 
\looseness=-1
Given a probability path $\rho_t$ that connects $\rho_0$ to $\rho_1$, and its associated flow $\psi_t$, we can learn a CNF by directly regressing the vector field $u_t$ with a parametric one $v_{\theta} \in  \mathfrak{X}(\gM)$. This technique is termed \emph{flow matching}~\citep[FM]{lipman_flow_2022} and leads to a simulation-free training objective as long as $\rho_t$ satisfies the boundary conditions $\rho_0 = \rho_{\text{data}}$ and $\rho_1 = \rho_{\text{prior}}$. Unfortunately, the vanilla flow matching objective is intractable as we generally do not have access to the closed-form of $u_t$ that generates $\rho_t$. Instead, we can opt to regress $v_{\theta}$ against a conditional vector field $u_t(x_t | z)$, generating a conditional probability path $\rho_t(x_t | z)$, and use it to recover the target unconditional path: $\rho_t(x_t) = \int_{\gM} \rho_t(x_t | z) q(z) dz$. The vector field $u_t$ can also be recovered by marginalizing of conditional vector fields:
$u_t(x) := \int_\gM  u_t(x | z)\frac{ \rho_t(x_t | z) q(z)}{\rho_t(x)}  dz$. The Riemannian CFM objective~\citep{chen2023riemannian} is then,
\begin{equation}\label{eq:CFM}
\gL_{\rm rcfm}(\theta) = \mathbb{E}_{t, q(z), \rho_t(x_t | z)} \|v_\theta(t, x_t) - u_t(x_t | z)\|_g^2, \quad t \sim \mathcal{U}(0,1).
\end{equation}
\looseness=-1
As FM and CFM objectives have the same gradients~\citep{tong2023improving}, at inference, we can generate by sampling from $\rho_1$, and using $v_{\theta}$ to propagate the ODE backward in time.

\subsection{Protein Backbone Parametrization}
\label{sec:backbone_parametrization}
 \begin{wrapfigure}[8]{r}{0.43\textwidth}
    \centering%
    \vspace{-40pt}
    \includegraphics[width=0.35\textwidth]{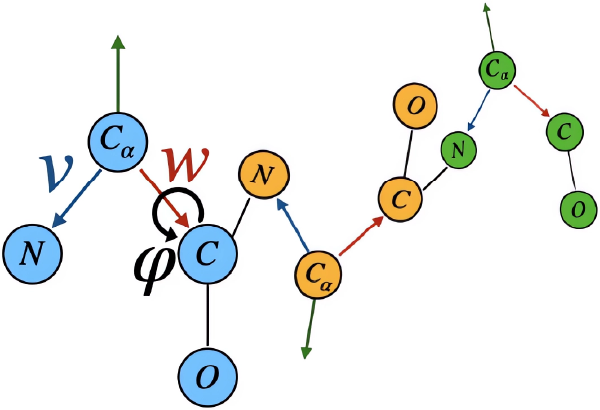}
    \vspace{-5pt}
    \caption{\small Protein backbone parametrization.}
    \vspace{-10pt}
    \label{fig:protein_frams}
\end{wrapfigure}
\looseness=-1
Our protein backbone parameterization follows the seminal work of AlphaFold2~\citep[AF2]{jumper2021highly} in that we associate a frame with each residue in the amino acid sequence. For a protein of length $N$ this results in $N$ frames that are $\sethree$-equivariant.
Each frame maps a rigid transformation starting from idealized coordinates of four heavy atoms $\text{N}^*,\text{C}^*_{\alpha}, \text{C}^*,\text{O}^* \in \R^3$, with $\text{C}^*_{\alpha} = (0,0,0)$ being centered at the origin, and is a measurement of experimental bond angles and lengths~\citep{engh2012structure}.
Thus, residue $i\in [N]$ is represented as an action of $x^i = (r^i,s^i) \in \sethree$ applied to the idealized frame $[\text{N},\text{C}_{\alpha},\text{C},\text{O}]^i = x^i \circ [\text{N}^*,\text{C}^*_{\alpha},\text{C}^*,\text{O}^*]$. To construct the backbone oxygen atom $\text{O}$, we rotate about the axis given by the bond between $\text{C}_{\alpha}$ and $\text{C}$ using an additional rotation angle $\varphi$. Finally, we denote the full 3D coordinates of all heavy atoms as $\text{A} \in \R^{N \times 4 \times 3}$.
An illustration for this backbone parametrization is provided in \cref{fig:protein_frams} with rotations being parametrized as $r = v \times w$.


\section{\foldflow, for condition flow matching on $\sethree$}
\vspace{-5pt}
\label{sec:se3_flow_matching}
\looseness=-1
We seek to learn an $\sethreen$ invariant density $\rho_t$ by training a flow using the objective in \cref{eq:CFM}. To do so we can pushforward an $\sethreen$-invariant source distribution $\rho_1$ to the empirical distribution of proteins $\rho_0$ using an equivariant flow. One way to guarantee the existence of a translation-invariant measure is to construct a subspace that is invariant to global translations. This can be achieved by simply subtracting the center of mass of all inputs to the flow~\citep{rudolph2021same,yim2023se}. Formally, this leads to an invariant measure on $\sethreenzero$ which is a subgroup of $\sethreen$.
We then note that $\sethreenzero$ is a product group and thus the Riemannian metric extends in a natural way to the product space: the exponential and logarithmic maps decompose across each manifold, and the geodesic distance in $\sethreenzero$ is simply the sum of each individual distance in the product. 
As such, a flow on $\sethreenzero$ can be built from separate flows for each residue in the backbone, on $\sethree$, after centering. %

\xhdr{Decomposing $\sethree$ into $\sothree$ and $\R^3$}
\looseness=-1
As Lie groups are manifolds, they can be equipped with a metric to obtain a Riemannian structure. In the case of $\sethree \cong \sothree \ltimes (\R^3, +)$ there are multiple possible choices, but a natural one is $\langle \mathfrak{x}, \mathfrak{x}' \rangle_{\sethree} = \langle \mathfrak{r}, \mathfrak{r}' \rangle_{\sothree} + \langle s , s' \rangle_{\R^3}$ (see~\S\ref{app:sethree}). Moreover, the disintegration of measures implies that every $\sethree$-invariant measure can be broken down to a $\sothree$-invariant measure and a measure proportional to the Lebesgue measure on $\R^3$~\citep{pollard2002user}. Thus, we may simply build independent flows on $\sothree$ and $\R^3$. In this section, we focus on designing \foldflow, models on $\sothree$, as CFMs on $\R^d$ are well-studied in \citet{albergo_building_2023,lipman_flow_2022,tong2023improving} (we provide a complete description in~\S\ref{app:fm_euc}). We use the notation $\rho_t$, $q$, and $\pi$ for densities whose support is determined by its context.

\subsection{\foldflowbase,}
\vspace{-5pt}
\label{sec:so3_flow_matching}
\looseness=-1
To construct a flow on $\sothree$ that connects the target distribution $\rho_0$ to a source distribution $\rho_1$, we must first choose a parametrization of the group elements. The most familiar 
and natural 
parametrization 
is by orthogonal matrices with unit determinant (see~\S\ref{app:sothree} for a discussion on other parametrizations e.g. rotation-vector). The Lie algebra $\sothreelie$ contains skew-symmetric matrices $\mathfrak{r}$ that are tangent vectors at the identity of $\sothree$. The last important component that we require is a choice of Riemannian metric for $\sothree$. A canonical bi-invariant metric for $\sothree$ can be derived from the Killing form (see~\S\ref{app:sothree}), and is given by: $\langle \mathfrak{r}, \mathfrak{r}' \rangle_\sothree = \mathrm{tr}(\mathfrak{r} \mathfrak{r}'^T)/2$.

\cut{
There are several other parameterizations of $\sothree$ besides the rotation matrices, such as Euler angles, quaternions, and axis-angle. The axis-angle representation represents a rotation via an axis (indicating the direction) and an angle of rotation. This means that any $r \in \sothree$ can be represented by a vector, $\mathbf{a}$, in $\mathbb{R}^3$, whose magnitude, $\omega_a$ is the angle of rotation around that vector. The axis-angle representation is connected to the lie-algebra $\sothreelie$ via the hat operation, taking the rotation vector and converting it to an equivalent (skew-symmetric) matrix. Therefore, any matrix $r \in \sothree$, we can write $r = \exp{(\hat{\mathbf{a}})}$, for $\mathbf{a}$ a rotation vector.

\tara{figure out notation for rotvec a and matrix x}

As smooth manifolds, all lie groups can be equipped with a Riemannian metric, providing a way to measure distances between the group elements. 
A canonical bi-invariant metric for $\sothree$ can be derived from the Killing form \tara{cite?}, and is given by:

\begin{equation}
\label{eq:norm_so3}
    \langle \mathfrak{r}, \mathfrak{r}' \rangle_\sothree = \| \log( \mathfrak{r}^T \mathfrak{r}') \|_F,
\end{equation}
for $\mathfrak{r}, \mathfrak{r}' \in \sothreelie$, and $\log$ is the matrix logarithm (the inverse of the matrix exponential map).
}
\cut{
The special Euclidean group in $3$ dimensions, $\sethree$, is also a lie group, whose elements are rigid transformations (rotation followed by a translation). Formally, this group can be written as the semidirect product of the group of rotations and the group of translations in three dimensions, $\sethree = \sothree \rtimes \mathbb{R}^3$.

\tara{how much do we want on the metric for SO(3)? math is done in se3 diffusion}
}

\xhdr{$\sothree$ conditional vector fields and flows}
\looseness=-1
We seek to construct a conditional vector field $u_t(r_t | z)$, lying on the tangent space $\mathcal{T}_{r_t} \sothree$, that transports $r_0 \sim \rho_0$ to $r_1 \sim \rho_1$. Following,~\citet{tong2023improving} we set the conditioner to $z = (r_0, r_1)$. Next, we construct a flow $\psi_t$ that connects $r_0$ to $r_1$. We follow the most natural strategy which is to build the flow using the geodesic between $r_0$ and $r_1$. For general $\gM$, the geodesic interpolant between two points, indexed by $t$, has the following form:
\begin{equation}
\label{eq:sampling_r_t}
    r_t = \exp_{r_0}(t \log_{r_0}(r_1)). 
\end{equation}
For rotation matrices, \cref{eq:sampling_r_t} involves computing the $\exp$ and $\log$ maps which are both infinite matrix power series. Unfortunately, controlling the approximation error of $\log_{r_0}$ map is computationally expensive as the de facto inverse scaling method for computing matrix logarithms requires estimating and calculating fractional matrix powers~\citep{al2012improved}. Instead, we use a numerical trick by converting $r_1$ to its axis-angle representation which gives a vector representation of $\mathfrak{r}_1 \in \sothreelie$ and, by definition, lives at the tangent space at the identity and is equivalent to $\log_{e}(r_1)$. Next, we can parallel transport $\mathfrak{r}_1$ to the tangent space of $r_0$ since Lie algebras of all tangent spaces are isomorphic and $\sothree$ carries a free action which gives us the desired end result $\log_{r_0}(r_1)$.


\looseness=-1
Given $r_t$, we can build constant velocity vector fields by directly leveraging the ODE associated with the conditional flow: $\frac{d}{dt} \psi_t(r) = \dot{r}_t$~\citep{chen2023riemannian}. As a result, computing $u_t(r_t | z)$ boils down to computing the point $r_t$ along the ODE and taking its time derivative.
In practice, taking the time-derivative to compute $u_t = \dot{r}_t$ amounts to using autograd to compute the gradient during a forward pass. We can overcome this unnecessary overhead without relying on automatic differentiation but instead by exploiting the geometry of the problem. Specifically, we calculate the $\sothreelie$ element corresponding to the relative rotation between $r_0$ and $r_t$, given by $r^\top_tr_0$. We divide by $t$ to get a vector which is an element of $\sothreelie$ and corresponds to the skew-symmetric matrix representation of the velocity vector pointing towards the target $r_1$. Finally, we parallel-transport the velocity vector to the tangent space $\gT_{r_t}\sothree$ using left matrix multiplication by $r_t$. These operations can be concisely written as $\log_{r_t}(r_0)$, where we use our numerical trick to calculate the matrix logarithm. The closed form expression of the loss to train the $\sothree$ component of \foldflowbase, is thus 
\begin{mdframed}[style=MyFrameEq]
\begin{equation}\label{eq:foldflowbase}
\gL_{\foldflowbase,-\sothree}(\theta)=\mathbb{E}_{t\sim \mathcal{U}(0,1), q(r_0,r_1), \rho_t(r_t | r_0, r_1)} \left \|v_\theta(t, r_t) - \log_{r_t}(r_0)/t\right\|_{\sothree}^2. 
\end{equation}
\end{mdframed}
\looseness=-1

In \cref{eq:foldflowbase} the conditioning distribution $q(z) = q(r_0, r_1)$ is the independent coupling $q(r_0, r_1) = \rho_0 \rho_1$, where $\rho_1 = \mathcal{U}(\sothree)$ and is left-invariant w.r.t. to the Haar measure.
Also, note that the vector field in \cref{eq:foldflowbase} is on the tangent space $v_{\theta} \in \gT_{r_t}\sothree$ and the norm is induced by the metric on $\sothree$.


\subsection{\foldflowot,}
The conditional vector field $u_t(r_t | z)$ generates the conditional probability path $\rho_t(r_t|z)$ which deterministically evolves $\rho_0$ to $\rho_1$. However, there is no reason to believe the conditional probability path is \emph{optimal} in the sense that it is a length-minimizing curve, under an appropriate metric, in the space of distributions $\mathbb{P}(\sothree)$. We seek to rectify this by constructing conditional probability paths that are not only shorter and straighter, but also more stable from an optimization perspective. This is motivated by previous research~\citep{tong2023improving, pooladian2023multisample} which has shown optimal transport to lead to faster training with a lower variance training objective in Euclidean spaces.

\looseness=-1
To this end, we propose \foldflowot,, a model that accelerates \foldflowbase, by constructing conditional probability paths using \emph{Riemannian optimal transport}. 
The interpolation measure $\rho_t$ connects $\rho_0 \to \rho_1$ and is built from Riemannian OT which solves the Monge optimal transport problem:
\begin{equation}\label{eq:monge}
\operatorname{OT}(\rho_0, \rho_1) = \inf_{\Psi: \Psi_\#\rho_0=\rho_1} \int_{\sethreenzero} \frac12 c(x, \Psi(x))^2\,d\rho_0(x).
\end{equation}
Here $c$  is the geodesic cost induced by the metric (cf. \cref{eq:sethree_distance} in \S\ref{app:sethree}) and $\Psi$ a pushforward map: $\rho_0 \to \rho_1$. A related problem, called the OT-Kantorovich formulation, relaxes the Monge problem by looking for a joint probability distribution $\pi$ minimizing the displacement cost of transporting $\rho_0$ to $\rho_1$ (see~\S\ref{app:riemannianOT}). 
The uniqueness of the Monge map over $\sethreenzero$ is guaranteed under some assumptions on the measures $\rho_0, \rho_1$ as stated in the following proposition and proven in~\S\ref{app:proof_prop_1}.
\begin{mdframed}[style=MyFrame2]
\label{monge_kantorovich_se3}
\begin{restatable}{proposition}{newriemannianotprop}
    Let us consider $\sethreenzero$ with the product distance $d_{\sethreenzero}$ and two compactly supported probability distributions $\rho_0, \rho_1 \in \sP(\sethreenzero)$. In addition, suppose that $\rho_0$ is absolutely continuous with respect to Riemannian volume form (\emph{i.e.,} $\rho_0 \ll dx$). Then for the distance $c = \frac12 d_{\sethreenzero}^2$, the Kantorovich and Monge problems admit a unique solution that is connected as follows $\pi = (id \times \Psi)_\#\rho_0$, where $\Psi$ is almost uniquely determined everywhere $\rho_0$. Furthermore, we have that $\Psi(x) = \exp_x(\nabla \phi(x))$ for some $d_{\sethreenzero}^2$-concave function $\phi$. 
\end{restatable}
\end{mdframed}
\vspace{-10pt}
\looseness=-1
Following this proposition, we define the McCann interpolants as $\rho_t(x) = (\exp_{x}(-t\nabla \phi (x)))_\#\rho_0$. 
While it is possible to approximate the Monge map and McCann interpolants using $c$-concave functions, it imposes practical limitations on the architecture of the flow~\citep{cohen2021riemannian}. 
Instead, we use the correspondence between the Monge and Kantorovich problems and rely on the optimal transport plan $\pi$. Formally, we draw two samples from $q(z) = q(x_0, x_1):= \pi(x_0, x_1)$ and we compute for a given frame $ \rho_t(r_t|r_0, r_1) = \delta(\exp_{r_0}(t \log_{r_0}(r_1)))$, where $\delta$ is a Dirac. Since our choice of metric for $\sethree$ factorizes into metrics on $\sothree$ and $\R^3$, we can use independent losses on rotations and translations---similar to \foldflowbase,---and repeat this over $N$ frames, as long as each geometric quantity in $\pi$ is coupled properly. Defining $\bar{\pi}(r_0, r_1)$ as the projection of $\pi(x_0, x_1)$ on $\sothree$, we present the $\sothree$ loss for a single frame in \foldflowot, as 
\begin{mdframed}[style=MyFrameEq]
\begin{equation}\label{eq:foldflowot}
\gL_{\foldflowot,-\sothree}(\theta) = \mathbb{E}_{t\sim \mathcal{U}(0,1), \bar{\pi}(r_0, r_1), \rho_t(r_t | r_0, r_1)} \left \|v_\theta(t, r_t) - \log_{r_t}(r_0)/t\right\|_{\sothree}^2.  
\end{equation}
\end{mdframed}
\looseness=-1
\cut{
Instead of taking a uniform latent distribution $q$, we follow \cite{tong2023improving, pooladian2023multisample} and chose $q$ to be an optimal transport map. Let $(\gM, d)$ be our Riemannian manifold equipped with the metric $d$. For two probability distributions $\rho_0, \rho_1 \in \sP(\gM)$, the Monge optimal transport problem is defined as
\begin{equation}\label{eq:monge}
\operatorname{OT}(\rho_0, \rho_1) = \inf_{\Psi: \Psi_\#\rho_0=\rho_1} \int_{\gM} \frac12 d(x, \Psi(x))^2\,d\rho_0(x).
\end{equation}
When $\gM$ is a smooth compact manifold with no boundary and $\rho_0$ has a density, \cite[Proposition 9]{McCann2001PolarFO} shows that the map $T$ exists and is unique. Intuitively, the Monge map is the map that minimizes the displacement between $\rho_0$ and $\rho_1$. It is equal to $\Psi(x) = \exp_{x}(-\nabla \phi (x))$, where $\phi$ is a $c$-concave function. In the case of the $\sethreen$ manifold, the Monge map also exists:

\begin{mdframed}[style=MyFrame2]
\begin{proposition}
    Let us consider $(\sethreen, d_{\sethreen})$ and two probability distributions $\rho_0, \rho_1 \in \sP(\sethreen)$. In addition, suppose that $\rho_0$ is absolutely continuous with respect to Riemannian volume (\emph{i.e.,} $\rho_0 \ll d \text{Vol}$). Then, there exists a unique map $\Psi$ pushing $\rho_0$ to $\rho_1$ minimizing the Monge problem for the distance $c = \frac12 d_{\sethreen}^2$.
\end{proposition}
\end{mdframed}

\begin{proof}
    Use Villani's Theorem for non-compact manifold
\end{proof}

Furthermore, one can move the distribution $\rho_0$ to $\rho_1$ in time using the McCann interpolants. It is defined as $\rho_t(x) = \exp_{x}(-t\nabla \phi (x))$. The Monge map and McCann interpolants were approximated by \cite{cohen2021riemannian} who parameterized the space of optimal transport maps by $c$-concave functions for some manifolds. Instead, we choose to approximate the Monge map by an optimal transport plan that solves the Kantorovich optimal transport problem:
\begin{equation}\label{eq:ot}
W(\rho_0, \rho_1)^2_2 = \inf_{\pi \in U(\rho_0, \rho_1)} \int_{\gM^2} \frac12 d(x, y)^2\,d\pi(x, y),
\end{equation}
where $U(\rho_0, \rho_1)$ is the set of joint probability measures on $\gM\times\gM$ whose marginals are $\rho_0$ and $\rho_1$. Therefore, we compute the Monge map and McCann interpolants by drawing two samples from the transport plan $(x_0,x_1) \sim \pi, x_t = \rho_t(x|x_0, x_1) = \exp_{x_0}(-t \log_{x_0} x_1)$ as described in \cite[Section 2.3]{sarrazin2023linearized}
}

\vspace{-10pt}
\subsection{\foldflowsfm,}
\label{sec:foldflowsfm}
\looseness=-1
We finally present \foldflowsfm,, which builds on the foundations of both \foldflowbase, and \foldflowot,. Departing from the deterministic dynamics of the previous models, we aim to build a \textit{stochastic} flow over $\sethreen_0$ by replacing these deterministic bridges with guided stochastic bridges. Previous research has shown that, compared to ODEs, SDEs have the important benefit of being more robust to noise in high dimensions~\citep{tong2023simulationfree, shi_diffusion_2023, liu_sb_2023}. For proteins in $\sethreen_0$, this means the generative process has greater empirical calibre to sample \emph{outside} the support of the training distribution---crucial for generating designable proteins that are also \emph{novel}.
In Euclidean space, we can build a translation invariant flow on $(\R^3)^N_0$ by using a (reverse time) Brownian bridge as the conditional flow between the points, 
\begin{equation}
\label{eqn:translation_sde}
    \ddd \rm{S}_t = \frac{\rm{S}_t - s_0}{t} \dt + \gamma(t) \ddd \rm{W}_t, \quad \rm{S}_1 = s_1.
\end{equation} 
This flow, also known as Doob's h-transform \citep{doob_classical_1984}, is easy to sample from in a simulation-free manner and correctly maps between arbitrary samplable marginals in expectation. Specifically, we can build a simulation-free bridge by sampling from the conditional probability $\rho_t(x_t | x_0, x_1) = \mathcal{N}(x_t; t x_1 + (1-t) x_0, \gamma^2 t (1-t))$~\citep{shi_diffusion_2023,albergo_stochastic_2023}. See~\S\ref{app:flow_matching_rd} for more details.

\xhdr{Brownian Bridge on $\sothree$}
On $\sothree$, we aim to model the dynamics between rotations matrices by a guided diffusion bridge~\citep{jensen2022bridge,liu2022learning}, leading to the following dynamics,
\begin{equation}
\label{eq:r_bridge}
    \ddd\rm{R}_t = \frac{ \log_{\rm{R}_t} r_0}{t} \dt + \gamma(t) \ddd \rm{B}_t, \quad \rm{R}_1 = r_1,
\end{equation}
for $\rm{B}_t$ the Brownian motion on $\sothree$ (see~\S\ref{app:brown_bridge_so3} for further technical details on this SDE). Despite the close resemblance of this SDE to the translation one in~\cref{eqn:translation_sde}, the corresponding Brownian bridge---to the best of our knowledge---does not have a closed-form expression for $\rho_t(r_t | r_0, r_1)$. Thus, to sample from $\rho_t(r_t | r_0, r_1)$ correctly, we start at $r_1$ and simulate the bridge backward in time using \cref{alg:sfm_inference} in~\S\ref{app:sec:sde_train_inf}. Given this form of the conditional bridge, we can make use of a flow-matching loss to optimize a flow between source and target distributions on $\sothree$ as follows: 




\begin{mdframed}[style=MyFrameEq]
\begin{equation}\label{eq:sfm-sothree}\hspace{-0.5cm}
\gL_{\textsc{SFM}-\sothree}(\theta) = \mathbb{E}_{t\sim \mathcal{U}(0,1), \bar{\pi}(r_0, r_1), \rho_t(\tilde{r}_t | r_0, r_1)} \left \|v_\theta(t, \tilde{r}_t) - \log_{\tilde{r}_t}(r_0)/t \right\|_{\sothree}^2.
\end{equation}
\end{mdframed}
Here $\tilde{r}_t$ is a sample from the bridge between $r_0$ and $r_1$. When $\pi(x_0, x_1)$ is a valid coupling between $\rho_0$ and $\rho_1$---and thus $\bar{\pi}(r_0, r_1)$ on $\sothree$---this objective is equivalent in expectation to matching directly the (computationally intractable) marginal loss $\gL_{\textsc{USFM}} = \mathbb{E}_{t\sim \mathcal{U}(0,1), \rho_t(\tilde{r}_t)} \left \|v_\theta(t, \tilde{r}_t) - u(t, \tilde{r}_t) \right\|_{\sothree}^2$.
The correctness of this approach is established in the next proposition and proved in~\S\ref{app:proof_proptwo}.
\begin{mdframed}[style=MyFrame2]
\begin{restatable}{proposition}{proptwo}
    Given $\rho_t(x) > 0$, $\forall x \in \sethreen_0$, the conditional and unconditional \foldflowsfm, losses have equal gradients \emph{w.r.t.} $\theta$:
    $\nabla_\theta \gL_{\textsc{USFM}}(\theta) = \nabla_\theta \gL_{\textsc{SFM}}(\theta)$. 
\end{restatable}
\end{mdframed}
\vspace{-10pt}
This result allows us to learn a stochastic flow from any source to any target distribution supported on $\sethreen_0$, only requiring samples from both distributions. However, it does require simulation of an SDE to sample from the conditional probability $\rho_t(r_t | r_0, r_1)$, limiting scalability.

\xhdr{An Efficient Simulation-free Approximation} Unfortunately, sampling from the correct conditional bridge requires simulation and is thus computationally expensive for training. In practice, we use a simulation-free approximation that closely matches the true conditional probability path on $\sothree$, $\rho_t(\tilde{r}_t | r_0, r_1)$. Specifically, we approximate $\rho_t$ with the simulation-free alternative, 
\begin{equation}\label{eq:igso_approx}
    \hat{\rho}_t(\tilde{r}_t|r_0, r_1) = \igso\Big(\tilde{r}_t; \exp_{r_0}(t \log_{r_0}(r_1)), \gamma^2(t) t(1-t)\Big),
\end{equation}
\looseness=-1
where $\igso$ denotes the isotropic Gaussian distribution on $\sothree$. This distribution can be seen as an analog of the Gaussian distribution in $\R^d$. It is the heat kernel on $\sothree$ \citep{Nikolayev1900} and it can be seen as the limit of small i.i.d.\ rotations in $3\rm D$~\citep{qiu2013isotropic}. Additionally, it has some of the desirable properties of the normal distribution, such as being closed under convolution \citep{Nikolayev1900}. We provide both the training and sampling algorithms for \foldflowsfm, in~\S\ref{app:experiment_setup}, details on $\igso$ in~\S\ref{app:isgo}, and results on the approximation error in~\S\ref{app:numerical_approximation_of_bridges}.

\section{Modeling Protein Backbones using \foldflow,}
\label{sec:implementation}


To model protein backbones using \foldflow, models we parameterize the velocity prediction $v_\theta(t, x_t)$ as a function that consumes a protein $x_t$ on the conditional path at time $t$ and predicts the starting point $\hat{x}_0$. Specifically, the predicted velocity is $v_\theta(t,x_t) = \nabla_g d(\hat{x}_0, x_t)^2/t$, with $\hat{x}_0 = w_\theta(t, x_t)$. 
This choice of parameterization has two principal benefits. (1) It allows the usage of specialized architectures specifically designed for structure prediction, and (2) it allows for auxiliary protein-specific losses to be placed directly on the $\hat{x}_0$ to improve performance.

\looseness=-1
\xhdr{Architecture} \looseness=-1 Following by~\citet{anand2022protein,yim2023se} (FrameDiff) we use the structure module of AF2 to model $w_\theta$. This begins with a time-dependent node and edge embeddings $N_\theta(t, x_t)$ and $E_\theta(t, x_t)$, followed by layers of invariant point attention. We use a small MLP head on top of the node embeddings to predict the torsion angle of the oxygen $\varphi$ as $\hat{\varphi} = \text{MLP}(N_\theta(t, x_t))$. 

\xhdr{Full Loss} Our \foldflow, models are trained to optimize a flow-matching loss on $\sothree$ and $\R^3$ for each residue $i\in [N]$ in the backbone. These are denoted $\gL_{\foldflow,-\sothree}$ and $\gL_{\foldflow,-\R^3}$ (see~\S\ref{app:fm_euc} for the complete expression in $\R^3$) respectively. In addition to the flow-matching losses, we also include auxiliary losses from \cite{yim2023se} which enforce good predictions at the atomic level in the $\R^3$ atomic representation $A$. These include a direct regression on the backbone (bb) positions $\gL_{\rm bb}$ and a loss on the pairwise atomic distance in a local neighbourhood $\gL_{2\rm D}$,
\begin{equation}
    \mathcal{L}_{\rm aux} = \mathbb{E}_{ \gQ} \left[\mathcal{L}_{\rm bb} + \mathcal{L}_{2\rm D} \right], \quad \mathcal{L}_{\rm bb} = \frac{1}{4N} \sum \|A_0 - \hat{A}_0\|^2, \quad \mathcal{L}_{2\rm D} = \frac{\| \mathbf{1}\{D < 6 \angstrom \} (D - \hat{D}) \|^2}{\sum \mathbf{1}_{D < 6 \angstrom} - N}
    \nonumber
\end{equation}
\looseness=-1
Here $\gQ(t, x_0, x_1, \tilde{x}_t):= \mathcal{U}(0,1) \otimes \bar{\pi}(x_0, x_1) \otimes \rho_t(\tilde{x}_t | x_0, x_1)$ is the factorized joint distribution, $\mathbf{1}$ is the indicator function, $A$ is in Angstroms ($\angstrom$), $D$ is an $N \times N \times 4 \times 4$ tensor containing the pairwise distances between the four heavy atoms, i.e.\ $D_{ijab} = \|A_{ia} - A_{jb}\|$, and $\hat{D}$ is defined similarly from $\hat{A}$. We only apply auxiliary losses for $t < 0.25$, with scaling $\lambda_{\rm aux}$ for a final loss of:
\begin{mdframed}[style=MyFrameEq]
\begin{equation}
    \gL_{\text{\foldflow,}}(\theta) = \gL_{\foldflow,-\sothree}
    + 
   \gL_{\foldflow,-\R^3}
    + 
    \mathbf{1} \{ t < 0.25 \} \lambda_{\rm aux}\gL_{\rm aux}.
\end{equation}
\end{mdframed}

\begin{figure}[tbp]
\captionsetup[figure]{aboveskip=-2pt,belowskip=-2pt}
    \centering
    \includegraphics[width=1\textwidth]{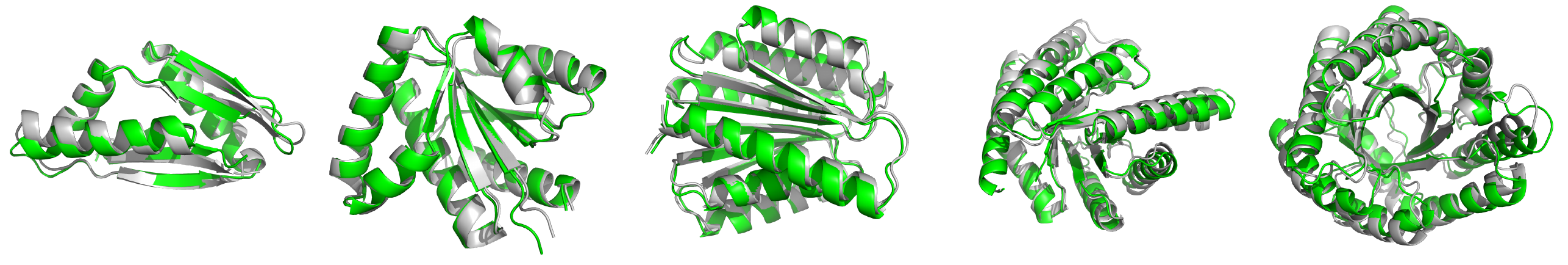}
    \vspace{-25pt}
    \caption{\small \foldflowsfm, generated structures in green compared to ProteinMPNN $\longrightarrow$ ESMFold refolded structures in grey. Samples with RMSD $<2 \angstrom$ for lengths 100, 150, 200, 250, 300 from left to right.}
    \label{fig:sfm-refold}
    \vspace{-10pt}
\end{figure}
\vspace{-5pt}
\section{Experiments}
\label{sec:experiments}
\begin{wraptable}{r}
{0.4\textwidth}
\vspace{-4.5mm}
    \centering
    \caption{\small Mean and std of the 1-  and 2-Wasserstein distances, computed against 5000 points in the test set, over 5 seeds.}
    \resizebox{0.4\textwidth}{!}{
    \begin{tabular}{lrrrrr}
        \toprule
        ($\mu \pm \sigma$)  &  $\mathcal{W}_1$ ($\times 10^{-2})$ &  $\mathcal{W}_2$ ($\times 10^{-1}$) \\
        \midrule
        \foldflowbase, & $5.39 \pm 0.88$ & $1.52 \pm 0.27$\\
        \foldflowot,   & $4.96 \pm 0.27$ & $1.25 \pm 0.12$\\
        \foldflowsfm,  & $4.92 \pm 1.56$ & $1.26 \pm 0.49$\\
        Simulated SDE  & $5.13 \pm 1.36$ & $1.33 \pm 0.44$\\
        \bottomrule
        \end{tabular}
        }
    \vspace{-5pt}
    \label{tab:toy_results}
\end{wraptable}
\subsection{$\sothree$ Synthetic data}
\looseness=-1
We evaluate all our \foldflow, models on synthetic multimodal densities on $\sothree$ as done by  \citet{brofos2021manifold} (see~\S\ref{app:toy_vfield} for details). We report the Wasserstein distance between generated and ground truth samples in \cref{tab:toy_results} and visualize the generated samples in~\cref{fig:toy_results} in~\S\ref{app:toy_para}. We find that all our proposed methods correctly model all the modes of the ground truth distribution. However, \foldflowbase, exhibits mode shrinkage in relation to the ground truth. \foldflowot,, \foldflowsfm,, and the simulated SDE results in comparable performance, with the OT-based method being the best. Importantly, this shows that our simulation-free approximation of the SDE does not hinder model performance, and combined with its significant speedup justifies its use in protein experiments.

\subsection{Protein Backbone Design}
\label{sec:exp_protein_backbone}
\begin{table}[htb]
    \vspace{-10pt}
    \centering
    \caption{\looseness=-1\small Comparison of Designability (fraction of proteins with scRMSD < $2.0\angstrom$ and mean scRMSD), Diversity (avg. pairwise TMscore), Novelty (max. TM-score to PDB and fraction of proteins with averaged max. TMscore $<0.5$ and scRMSD < $2.0\angstrom$). Designability and Novelty metrics include standard errors.$^*$RFDiffusion and Genie have larger training sets that likely overestimate novelty with respect to our dataset. 
    }
    \vspace{-10pt}
    \resizebox{0.95\textwidth}{!}{
    \begin{tabular}{lrrrrrr}
        \toprule
          \multicolumn{1}{c}{} & \multicolumn{2}{c}{Designability}  & \multicolumn{2}{c}{Novelty} & \multicolumn{1}{c}{Diversity ($\downarrow$)} & \multicolumn{1}{c}{iters / sec ($\uparrow$)}\\
          \cmidrule(lr){2-3} \cmidrule(lr){4-5}
          & Fraction ($\uparrow$) & scRMSD ($\downarrow$) & Fraction ($\uparrow$) &  avg. max TM ($\downarrow$)  &  &\\
        \midrule
        RFDiffusion                 & 0.969 $\pm$ 0.023 & 0.650 $\pm$ 0.136 & $^*$0.708 $\pm$ 0.060 & $^*$0.449 $\pm$ 0.012 & 0.256  & ---\\
        Genie                       & 0.581 $\pm$ 0.064   & 2.968 $\pm$ 0.344 & $^*$0.556 $\pm$ 0.093& $^*$0.434 $\pm$ 0.016 & 0.228 & ---\\
        FrameDiff-ICML              & 0.402 $\pm$ 0.062 & 3.885 $\pm$ 0.415 & 0.176 $\pm$ 0.124 & 0.542 $\pm$ 0.046 & 0.237 & ---\\
        FrameDiff-Improved          & 0.555 $\pm$ 0.071 & 2.929 $\pm$ 0.354 & 0.296 $\pm$ 0.112  & 0.457  $\pm$ 0.026 & 0.278  & ---\\
        \midrule
        FrameDiff-Retrained   & 0.612 $\pm$ 0.060 & 2.990 $\pm$ 0.307 & 0.108 $\pm$ 0.083 & 0.684 $\pm$ 0.032 & 0.403 & 1.278\\
        \foldflowbase,              & 0.657 $\pm$ 0.042 & 3.000 $\pm$ 0.271 & 0.432 $\pm$ 0.074  & 0.452  $\pm$ 0.024 & 0.264  & 2.674 \\
        \foldflowot,                & 0.820 $\pm$ 0.037 & 1.806 $\pm$ 0.249 & 0.484 $\pm$ 0.068 & 0.460 $\pm$ 0.020 & 0.247 & 2.673 \\
        
        \foldflowsfm,         & 0.716 $\pm$ 0.040 & 2.296 $\pm$ 0.391 & 0.544 $\pm$ 0.061  & 0.411  $\pm$ 0.023 & 0.248 & 2.647 \\
        \bottomrule
        \end{tabular}
        }
    \label{tab:main}
    \vspace{-10pt}
\end{table}

\looseness=-1
We evaluate \foldflow, models in generating valid, diverse, and novel backbones by training on a subset of the Protein Data Bank (PDB) with $22{,}248$ proteins. We compare \foldflow, to pretrained versions of  FrameDiff~\citep{yim2023se} (FrameDiff-ICML), the improved version on the authors' GitHub (FrameDiff-Improved), Genie~\citep{lin2023generating}, and RFDiffusion, which is the gold standard~\citep{watson_novo_2023}. We also retrain FrameDiff (FrameDiff-Retrained) on our dataset, which contains $\sim$10\% more admissible structures, while inheriting the majority of the hyperparameters of \textsc{FoldFlow}. We provide a detailed description of all the metrics in~\S\ref{app:protein_metrics}.
\cut{but briefly, we consider designability to be the primary metric as diversity/novelty are intuitively only evaluated on designable proteins; for models that have low designability, this may result in a larger variance of diversity/novelty.}
Figures~\ref{fig:sfm-refold} and~\ref{fig:sfm-refold_appendix} visualize generated samples and ESM-refolded structures.


\looseness=-1
We report our findings in table \ref{tab:main} and observe that \foldflow, outperforms FrameDiff-Retrained on all three metrics. We identify FrameDiff as the most comparable baseline as it is the current SOTA model that does not utilize pre-training while using comparable resources. In contrast to \foldflow, we highlight that RFDiffusion uses a pre-trained backbone and a significantly larger model ($60m$ vs.\ $17m$ parameters), training set, and compute resources ($1800$ vs.\ $10$ GPU days). We also note that Genie is trained on a larger dataset ($195k$ vs.\ $22k$), which hinders rigorous comparisons with \foldflow,. Next, we analyze the performance of \foldflow, on each metric in detail.


\looseness=-1
\xhdr{Designability} We measure designability using the \emph{self-consistency} metric with ProteinMPNN~\citep{dauparas_robust_2022} and ESMFold~\citep{lin2022language}, counting the fraction of proteins that refold ($C_\alpha$-RMSD (scRMSD) < $2.0\angstrom$ and mean scRMSD) over $50$ proteins at lengths $\{100, 150, 200, 250, 300\}$. 
In \cref{tab:main}, we find that all \foldflow, models achieve significantly higher Frac. designability score than all FrameDiff models, and appreciably close the gap to RFDiffusion, e.g.\ $\Delta = 0.149 \text{ vs.\ }\Delta = 0.357$ for \foldflowot, and FrameDiff-Retrained respectively. When retrained on our dataset with 10\% more samples, we find that FrameDiff is more designable, but is still below all \foldflow, models.
We also note that while \foldflowot, creates the highest fraction of designable proteins (excluding RFDiffusion), it has relatively low diversity and novelty. We find that adding stochasticity with \foldflowsfm, results in a model that beats FrameDiff-Improved on every metric and can dramatically improve novelty at the cost of worse designability (\cref{tab:main}). In \cref{fig:protein_lengths} in \S\ref{app:protein_experiments}, we plot designability versus sequence length and observe the largest gains on sequence lengths $<300$. 

\looseness=-1
\xhdr{Diversity} We use the average pairwise TM-score of the \textit{designable} generated samples averaged across lengths as our diversity metric (lower is better). We find an inverse correlation between performance on designability and diversity metrics for FrameDiff models, which interestingly does not hold for \foldflow, models. We note that \foldflow, models have comparable diversity to the baselines with \foldflowot, and \foldflowsfm, being the most diverse. 

\begin{figure}[t!]
\captionsetup[figure]{aboveskip=-2pt,belowskip=-1pt}
\captionsetup[subfigure]{aboveskip=-2pt,belowskip=-1pt}
    \centering
    \begin{subfigure}{0.34\textwidth}
        \centering
        \includegraphics[width=1\linewidth]{    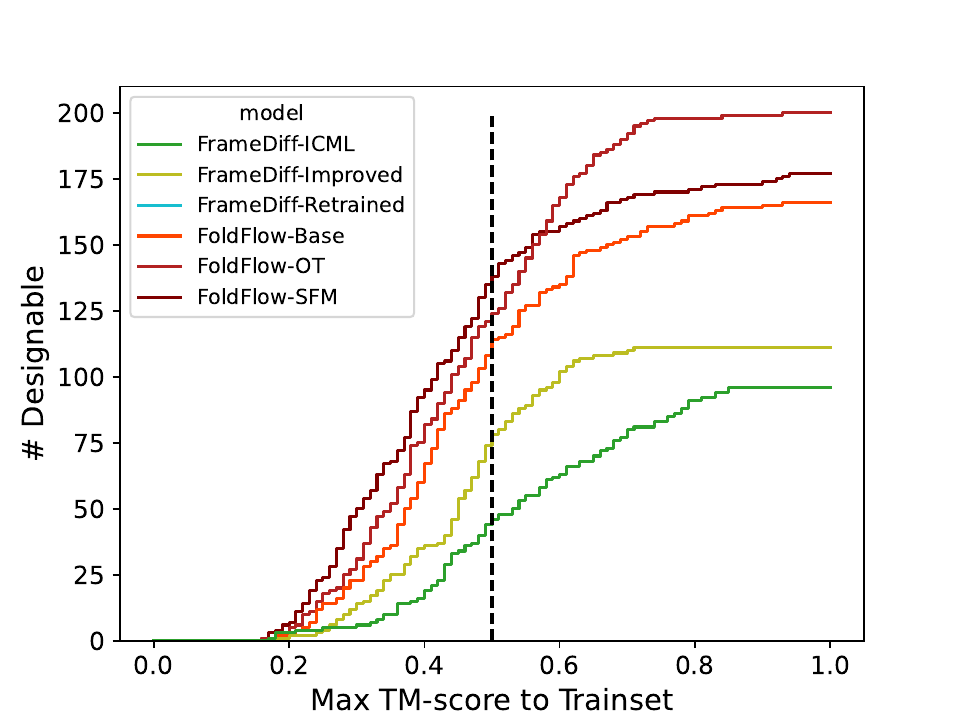}
        \caption{}
        \label{fig:designability_and_novelty}
    \end{subfigure}%
    \hspace{-5mm}
    \begin{subfigure}{0.30\textwidth}
        \centering
        \includegraphics[width=1\linewidth]{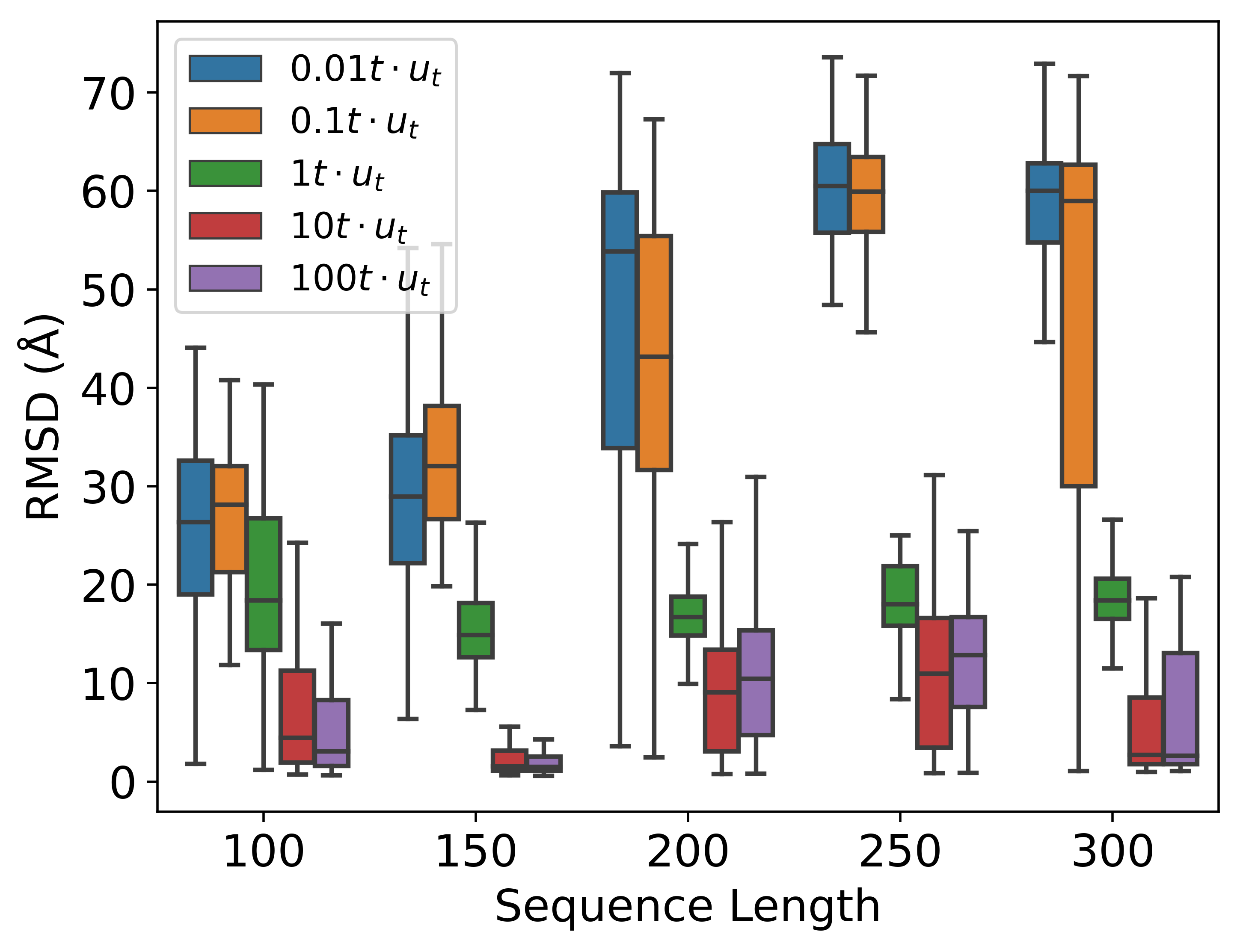}
        \caption{}
        \label{fig:inference_2}
    \end{subfigure}%
    \begin{subfigure}{0.32\textwidth}
        \centering
        \resizebox{1.0\textwidth}{!}{
        \begin{tabular}{P{1.5cm}|P{0.8cm} P{0.3cm}P{0.5cm}P{1.25cm}}
            \toprule
            \small{Designability} ($\uparrow$) & Stochas.  & OT & Aux. Loss & Inf. annealing \\
            \midrule
            0.716 & \cmark & \cmark & \cmark & \cmark \\
            0.820 & \xmark & \cmark & \cmark & \cmark \\
            0.657 & \xmark & \xmark  & \cmark & \cmark \\
            0.648 & \xmark & \xmark  & \xmark  & \cmark \\
            0.228 & \xmark & \xmark  & \xmark  & \xmark \\
            \bottomrule
            \end{tabular}
            }
            \vspace{5mm}
        \caption{}
        \label{tab:ablation}
    \end{subfigure}
    \vspace{-4mm}
    \caption{\small (a) Count of novel and designable (scRMSD < 2) proteins out of $250$ at various novelty thresholds with TM-score to the training set < $x$ for models trained on subsets of PDB. (b) scRMSD of designed proteins vs.\ ESMFold under flow scaling (c) Table showing an ablation study of \foldflow, features against designability.}
    \label{fig:inference}
    \vspace{-10pt}
\end{figure}

\looseness=-1
\xhdr{Novelty} Designing novel but realistic protein structures compared to the training data is also an important goal. Unlike conventional generative modeling problems, e.g. images, the novelty of proteins is particularly important since
the entire premise of ML-driven drug discovery requires developing \emph{original} drugs that may be vastly different than current human knowledge (training data) but also synthesizable (designable)~\citep{marchand2022computational,schneider2020rethinking,schneider2018automating}.
We measure novelty using two metrics: 1.) the fraction of designable proteins with TM-score $<0.5$ as used in~\citet{lin2023generating} (higher is better) and 2.) the average maximum TM-score of designable generated proteins to the training data (lower is better). In \cref{fig:designability_and_novelty} count the number of designable proteins as a function of the Max TM-score to the training set. We see that \foldflowsfm, designs the most novel structures against all methods including RFDiffusion and Genie. This substantiates the hypothesis that the stochasticity of the learned SDE is crucial to the improved robustness in high dimensions of \foldflowsfm, versus \foldflowbase,, and \foldflowot, which allows it to sample designable proteins far outside of the support of the training set. 

\xhdr{Inference Annealing} We now describe a numerical trick during inference that greatly improves the designability of \foldflow, which we term \textit{inference annealing}. Instead of following the theoretical ODE or SDE for generating rotations, we use a multiplicative scaling of the velocity---e.g.\ $\ddd \rm R_t = i(t) v_\theta(t, R_t) \dt + \gamma(t) \ddd \rm B_t$ for some positive function $i(t)$. In practice, we use $i(t) = c t$ for some constant $c$.
This annealing removes an unwanted increase in the flow norm during the end of inference (\cref{fig:inference_fig1}). We observe that larger $c$ drastically increases the designability of \foldflow, (\cref{fig:inference_2}). In practice, we use values of $c\approx 10$, which leads to designable yet diverse structures.

\looseness=-1
\xhdr{Ablation study of \foldflow,} Next, we ablate various additions to the \foldflowbase, model in \cref{tab:ablation} and report the full extended ablation in~\cref{tab:full_ablation}. We find \foldflowot, creates the most designable model, but adding stochasticity helps increase novelty and diversity. We also find that inference annealing is critical to the performance of the model in terms of achieving higher designability.


\subsection{Equilibrium Conformation Generation}
\label{exp:eq_conform}
\begin{figure}[htbp]
    \vspace{-8pt}
    \captionsetup[subfigure]{aboveskip=-1pt,belowskip=-1pt}
    \centering
    \begin{subfigure}[b]{0.33\textwidth}
        \includegraphics[width=\textwidth]{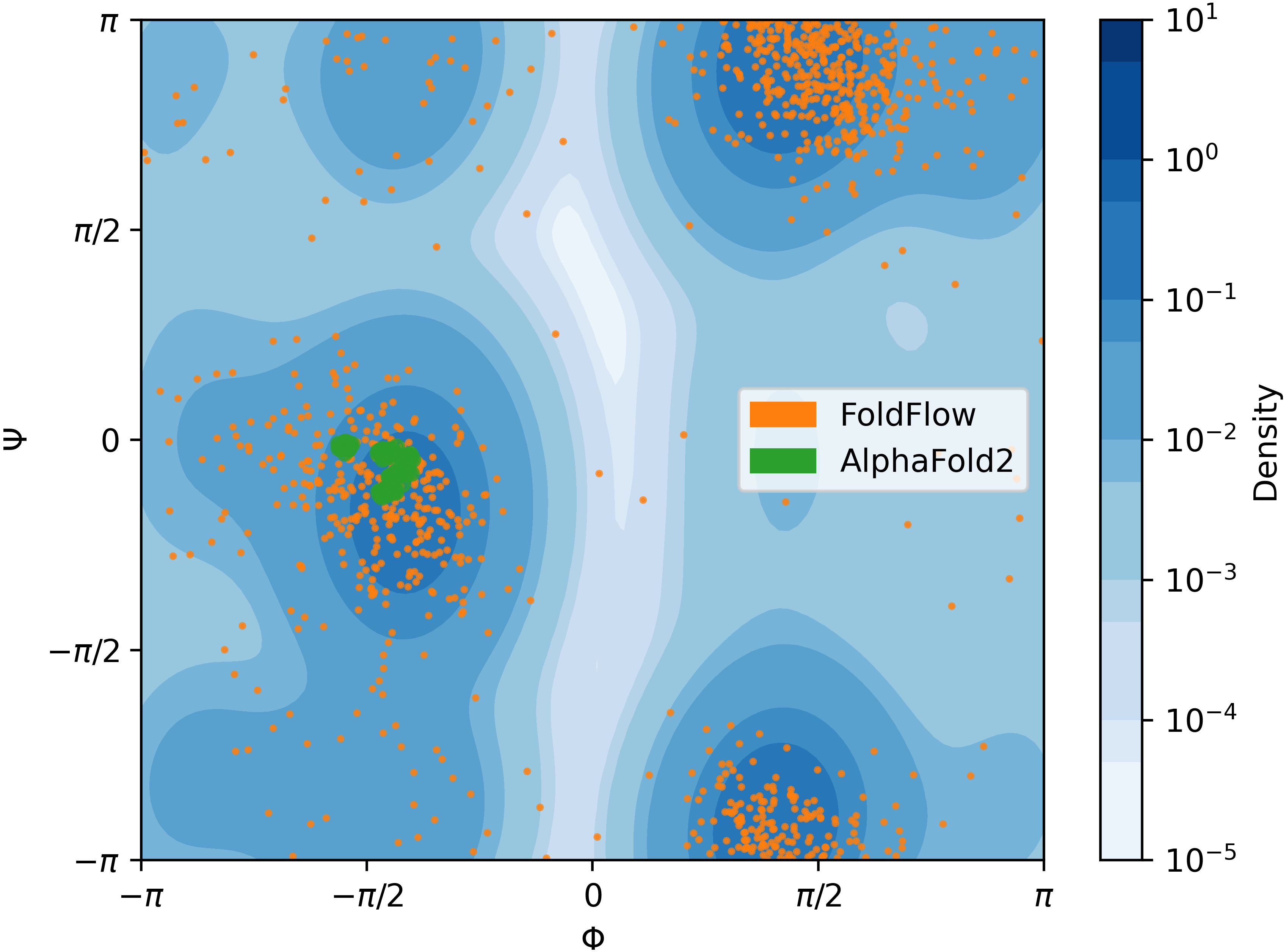}
        \caption{}
        \label{fig:md_fig:a}
    \end{subfigure}
    \begin{subfigure}[b]{0.315\textwidth}
        \includegraphics[width=\textwidth]{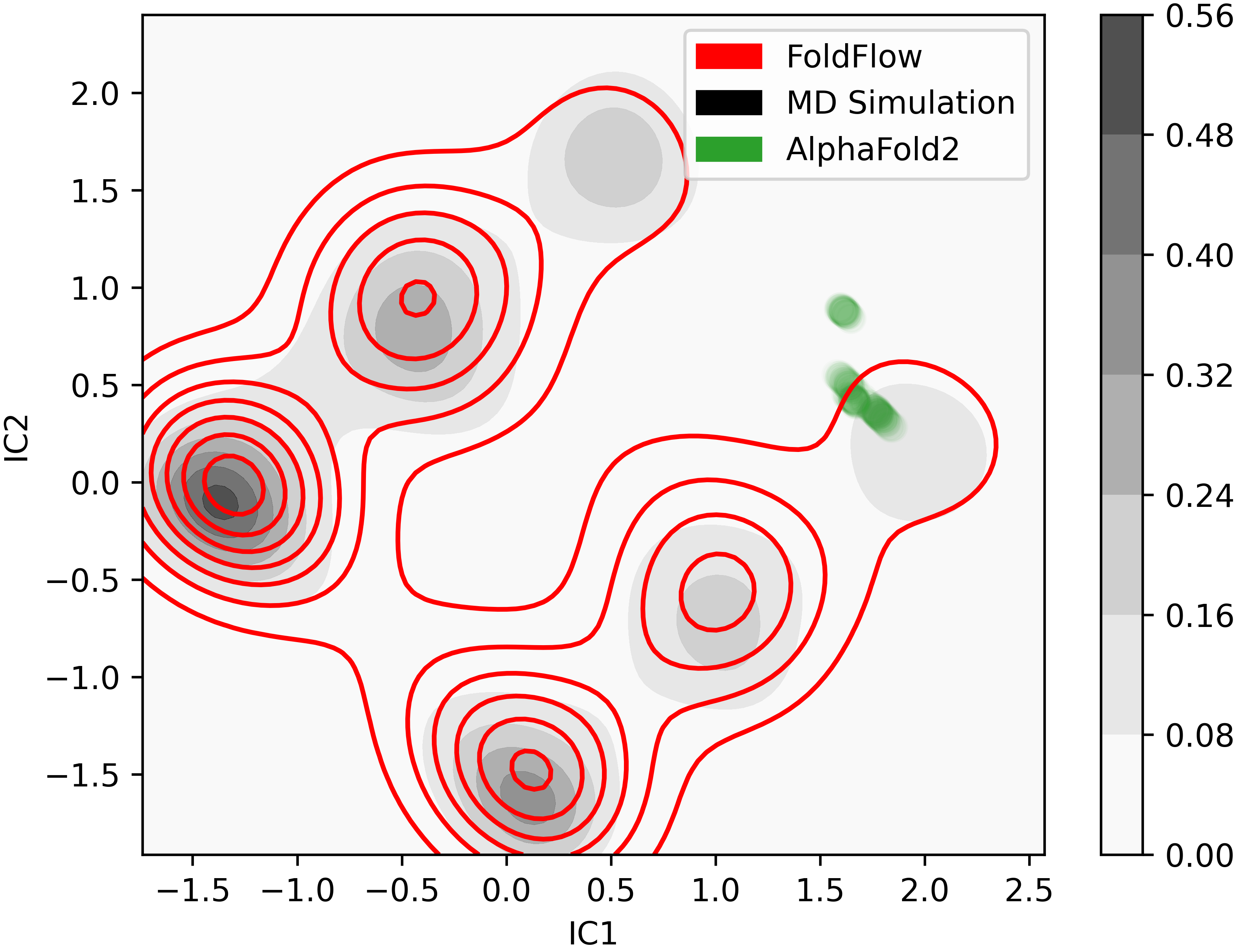}
        \caption{}
        \label{fig:md_fig:b}
    \end{subfigure}
    \begin{subfigure}[b]{0.315\textwidth}
        \includegraphics[width=\textwidth]{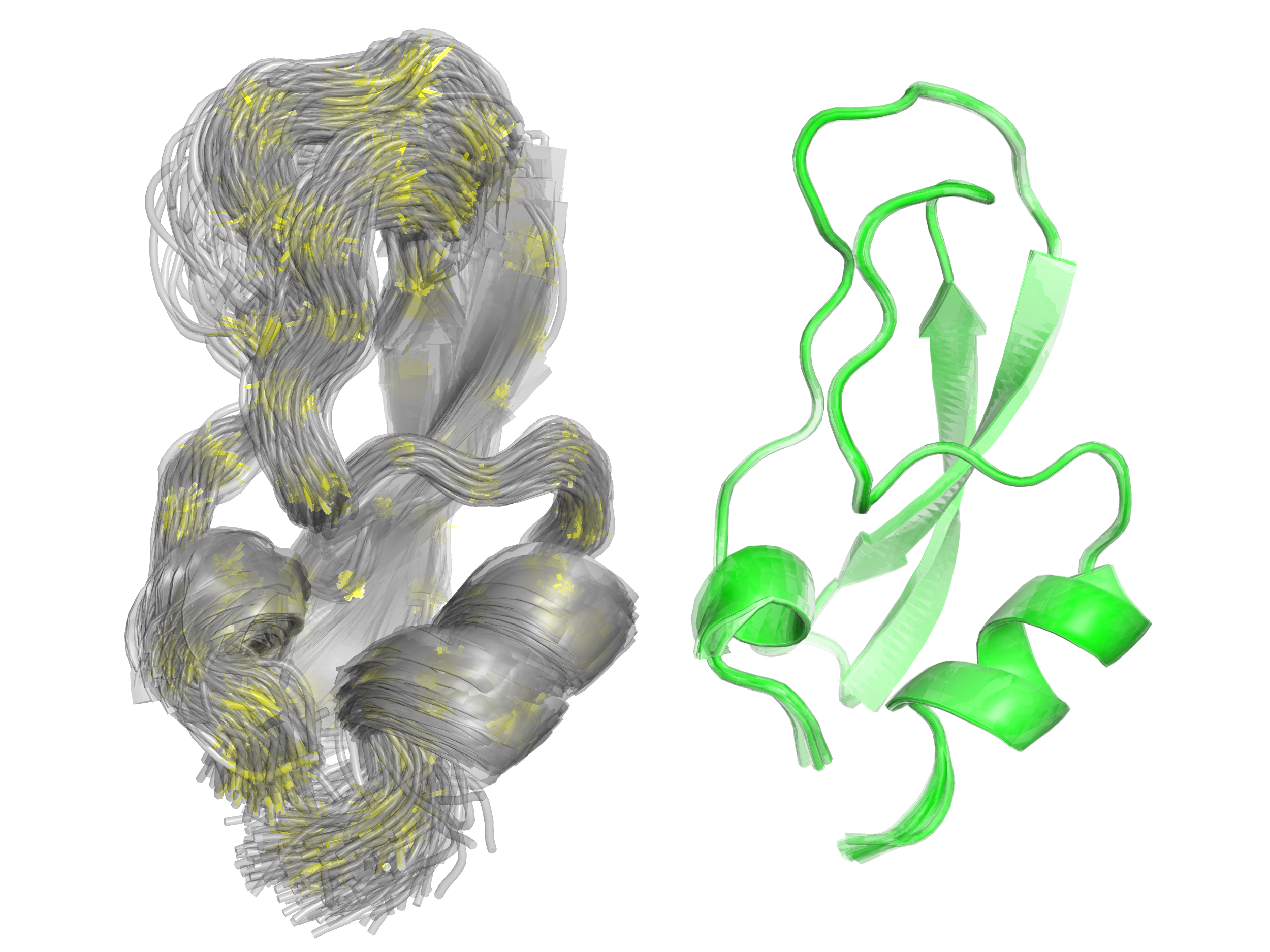}
        \caption{}
        \label{fig:md_fig:c}
    \end{subfigure}
    \vspace{-10pt}
    \caption{\small(a) Ramachandran plot of $\Phi$ and $\Psi$ of the most flexible residue (56) in BPTI (b) ICA of all dihedral angles of BPTI (c) $1000$ BPTI conformations sampled by \foldflow, with $C_{\alpha}$ alignment highlighted in yellow and AlphaFold2 samples in green.  \foldflow, reproduces test MD frames while AlphaFold2  samples do not.}
    \vspace{-10pt}
    \label{fig:md_fig}
\end{figure}

\begin{wraptable}{r}
{0.3\textwidth}
\vspace{-12.5mm}
    \centering
    \caption{\small $\mathcal{W}_2$ in angle space between generated and test samples.}
    \vspace{-2.5mm}
    \resizebox{0.3\textwidth}{!}{
    \begin{tabular}{lrr}
    \toprule
    {} &     $\mathcal{W}_2$ &  $\mathcal{W}_2 @ 56$ \\
    \midrule
    \foldflow,          &  4.379 &  0.406 \\
    \foldflow,-Rand     &  4.446 &  0.557 \\
    FrameDiff           &  4.844 &  0.800 \\
    \bottomrule
    \end{tabular}
        }
    \vspace{-10pt}
    \label{tab:equilibrium}
\end{wraptable}
\looseness=-1
Modeling various protein conformations is crucial in determining biological behaviours such as mechanisms of actions or binding affinity to other proteins. Unlike diffusion models, \foldflow, can easily be instantiated from any sampleable source distribution. To test this, we model the equilibrium distribution of a protein given initial predicted structures from pre-trained folding models including OmegaFold~\citep{OmegaFold} and ESMFold~\citep{lin2022language}. The training target distribution consists of 200,000 frames at $5 ns$ intervals of a $1 ms$ molecular dynamics trajectory of the BPTI protein~\citep{shaw2010atomic};  the inference of \foldflow, is tested against 20,000 unseen frames in that trajectory. \foldflow, can successfully model both the general set of conformations, as indicated by ICA of the dihedral angles in \cref{fig:md_fig:b}, as well as the highly flexible residues, as seen in the $2\rm D$ Ramachandran plot in \cref{fig:md_fig:a}. We note that our approach can capture all the modes of distribution in contrast to AlphaFold2, which does not model the flexibility well (\cref{fig:md_fig:c}). In \cref{tab:equilibrium}, we observe under the 2-Wasserstein ($\gW_2$) metric for all angles and only residue $56$, \foldflow, with an informed prior outperforms a random prior, and FrameDiff which can only use an uninformed prior which further highlights a key advantage of \foldflow, over diffusion-based approaches.

\vspace{-1mm}
\section{Related Work}
\label{sec:related_work}
\vspace{-1mm}
\xhdr{Protein design approaches}
\looseness=-1
The field of protein design has evolved over the course of several decades with many useful libraries~\citep{mccafferty1990phage,winter1994making,romero2009exploring,wang2021directed} with subfields being impacted by ML assitance~\citep{yang2019machine}. Sequence-based machine learning approaches resulted in multiple successful protein design cases~\citep{madani2023large,verkuil2022language,alamdari2023protein, hie2022high}. Structure-based biophysics approaches resulted in several drug candidates ~\citep{rothlisberger2008kemp,fleishman2011computational,cao2020novo}. Moreover, diffusion-based approaches have risen in prominence~\citep{wu2022protein,yim2023se}, including significantly improved biological experiment success compared to previous SOTA~\citep{watson_novo_2023}. \rev{$\sethree$ diffusion has also seen applications in protein-ligand binding~\citep{jin2023unsupervised}, docking~\citep{somnath_aligned_2023}, as well as robotics~\citep{brehmer2023edgi}. Lastly, FrameFlow~\citep{yim2023fast} concurrently investigates a model similar to \foldflowbase,.} 

\xhdr{Equivariant generative models}
\looseness=-1
There have been several efforts to incorporate symmetry constraints in generative models. These include building equivariant vector fields for CNFs~\citep{kohler2020equivariant, katsman2021equivariant, garcia2021n, klein2023equivariant} and finite flows using the affine coupling transform~\citep{dinh2016density,bose2021equivariant,midgley2023se}. Applications in theoretical physics have also been impacted by equivariant flows~\citep{boyda2020sampling,kanwar2020equivariant,abbott2305normalizing}. Lastly, beyond flows a new genre of models coming into prominence is based on the idea of equivariant score matching~\citep{de2022riemannian,brehmer2023edgi} and diffusion models~\citep{hoogeboom2022equivariant, xu2022geodiff,igashov2022equivariant}.

\vspace{-1mm}
\section{Conclusion}
\vspace{-1mm}
\label{sec:conclusion}
\looseness=-1
In this paper, we tackle the problem of protein backbone generation using $\sethreen$-invariant generative models. In pursuit of this objective, we introduce \foldflow,, a family of simulation-free generative models under the flow matching framework. Within this model class, we introduce \foldflowbase, which learns deterministic dynamics over $\sethree$, \foldflowot, which learns more stable flows using Riemannian OT. To learn stochastic dynamics, we propose \foldflowsfm, which learns an SDE over $\sethreen$, and is motivated by learning Brownian bridges over $\sothree$, but in a simulation-free manner. We investigate the empirical caliber of \foldflow, models on PDBs that contain up to $300$ amino acids and find that our proposed models are competitive with RFDiffusion while significantly outperforming the current non-pretrained SOTA approach, FrameDiff-Improved, on all metrics. Finally, \foldflow, is more amenable for equilibrium conformation sampling which is an important subtask in protein design. Beyond generating $3\rm D$ structures, a natural direction for future work is to extend \foldflow, to conditional generation by using target sequence and structure during training. 
\clearpage
\section*{Acknowledgements}
AJB was supported by the Ivado Phd fellowship. 
KF is supported by NSERC Discovery grant (RGPIN-2019-06512), CIFAR AI Chairs, and a Samsung grant. CHL is supported by the Vanier scholarship. The authors would like to thank Cl\'ement Bonet and Gauthier Gidel for fruitful discussions on Riemannian geometry and optimal transport. We also thank Riashat Islam for providing feedback on early versions of this work. We thank Alexander Stein and the entire DreamFold team for providing a vibrant workspace that enabled this research. The authors would like to thank the restaurant Le Don Donburi for keeping our stomachs full and sustaining this research. Finally, the authors would like to acknowledge Anyscale and Google GCP for providing computational resources for the protein experiments.

\bibliography{tidy}
\bibliographystyle{iclr2024_conference}

\clearpage
\appendix
\section{Theoretical Preliminaries}

\subsection{$\sothree$ Lie Group}
\label{app:sothree}
The Special Orthogonal group in $3$ dimensions, $\sothree$ consists of the $3\rm D$ rotation matrices:
\begin{equation}
    \sothree = 
    \left \{ r \in \R^{3\times3} : r^\top r = rr^\top = I, \det{r} =1
     \right \}
\end{equation}
It is a matrix Lie group with the lie algebra given by:
\begin{equation}
    \sothreelie = 
    \left \{ \mathfrak{r} \in \R^{3\times3} : \mathfrak{r}^\top = -\mathfrak{r}
     \right \}
\end{equation}

\xhdr{Parametrizations of $\sothree$}
The skew-symmetric matrices $\mathfrak{r} \in \sothree$ can be uniquely identified with a vector $\boldsymbol{\omega} \in \R^3$ such that $\forall \mathbf{v} \in \R^3$, $\mathfrak{r} \mathbf{v} = \boldsymbol{\omega} \times \mathbf{v}$, where $\times$ indicates the cross product. This vector is known as the \textit{rotation vector}. The magnitude of this vector, $\omega = || \boldsymbol{\omega} ||$ is the angle of rotation its direction, $e_{\boldsymbol{\omega}} = \frac{\boldsymbol{\omega}}{|| \boldsymbol{\omega} ||}$ is the axis of rotation.

Mapping the $\R^3$ vector to the skew-symmetric matrix is known as the \textit{hat} operation, $\hat{(\cdot)}$.

Another parametrization of $\sothree$ is with \textit{Euler angles}, described using three angles $(\phi, \theta, \psi)$. A common convention is to use the $x$-convention, where the rotation is given by: a rotation about the $z$-axis by $\phi$, a second rotation about the former $x$-axis by $\theta$, and a last one about the former $z$-axis by $\psi$.

\xhdr{Metric on $\sothree$}
First, we recall that a metric is a bilinear function $\langle \cdot, \cdot \rangle: \mathbb{R}^n \times \mathbb{R}^n \rightarrow \mathbb{R}$ that is both symmetric and positive definite. Additionally, we recall that a quadratic form on a manifold $\mathcal{M}$ is a bilinear map $\mathcal{T}_x \mathcal{M} \times \mathcal{T}_x \mathcal{M} \rightarrow \mathbb{R}$ that is smooth and symmetric. A positive-definite quadratic form is, therefore, a metric. Let us consider the following symmetric positive definite quadratic form defined:
\begin{equation}
\label{eq:quadratic_form}
    Q = 
    \begin{pmatrix}
       A & B^\top \\
        B & C
    \end{pmatrix}.
\end{equation}
A canonical choice for the metric of $\sothree$ is obtained by taking $Q=1/2I$, resulting in a bi-invariant metric on $\sothree$. Therefore, the metric is given by:
\begin{equation}
    \langle \mathfrak{r}_1, \mathfrak{r}_2 \rangle_{\sothree} = \text{tr}(\mathfrak{r}_1^\top Q \mathfrak{r}_2) = \frac{1}{2} \text{tr}(\mathfrak{r}_1^\top \mathfrak{r}_2)
\end{equation}
Note that the inner product on Lie groups consumes elements of the Lie algebra and, because the left action is transitive, this inner product is well-defined for all tangent spaces of the group elements.

The distance induced by this metric is given by:
\begin{equation}
\label{eq:dist_so3}
    d_\sothree(r_1, r_2) = \| \log{(r_1^\top r_2)}\|_F
\end{equation}
for $r_1, r_2 \in \sothree$ and where the Frobenius matrix norm is used. 

\xhdr{The exponential and logarithmic maps on $\sothree$}
Generally speaking, the \textit{exponential} and \textit{logarithmic} maps of a Lie group $G$ relate the elements in the group to the lie algebra, $\mathfrak{G}$. In the case of matrix Lie groups, these coincide with the \textit{matrix exponential}:
\begin{equation}
\label{eq:matrix_exp}
    g = \exp({\mathfrak{g}}) = \sum_{n=0}^N \frac{1}{n!} \mathfrak{g}^n
\end{equation}
and the \textit{matrix logarithm}:
\begin{equation}
    \mathfrak{g} = \log({g)} = \sum_{n=1}^N \frac{(-1)^{n-1}}{n} (g - I)^n
\end{equation}

For $\sothree$, since the elements of the lie algebra are skew-symmetric matrices, \cref{eq:matrix_exp} for the matrix exponential can be simplified significantly to obtain a closed-form, known as Rodrigues formula. Given $\boldsymbol{\omega}$ a rotation vector, and  $\hat{\boldsymbol{\omega}}\in \sothreelie$, the corresponding element of the lie group, $r \in \sothree$ is given by:
\begin{equation}
\label{eq:rodrigues}
    r = \exp{\hat{\boldsymbol{\omega}}} = \cos({\omega}) I + \sin({\omega}) e_{\boldsymbol{\omega}}+ (1 - \cos({\omega})) e_{\boldsymbol{\omega}} e_{\boldsymbol{\omega}}^\top
\end{equation}
where $\omega$ and $e_{\boldsymbol{\omega}}$ are the angle and axis of rotation for $\boldsymbol{\omega}$.

Similarly, the matrix logarithm can be expressed using the rotation angle:

\begin{equation}
    \log(r) = 
    \begin{cases} 
    \frac{\omega}{2 \sin(\omega)} (r - r^\top) & \text{if } r \neq I, \\
    0 & \text{if } r = I.
    \end{cases}
\end{equation}

\subsection{$\sethree$ Lie Group}
\label{app:sethree}

The special Euclidean group, $\sethree$
is used to represent rigid body transformations in $3$ dimensions:
\begin{equation}
    \sethree = 
    \left \{ 
    \begin{pmatrix}
       r & s \\
        0 & 1 
    \end{pmatrix}
     : r \in \sothree, s \in (\R^3, +)
     \right \}
\end{equation}
Represented by this $4 \times 4$ matrix and with the group operation defined by matrix multiplication, this group can be seen as a subgroup of the general linear group $\mathrm{GL}(4, \mathbb{R})$.
The lie algebra of the group $\sethreelie$ is given by: 
\begin{equation}
    \sethreelie = 
    \left \{ 
    \mathfrak{x} = 
    \begin{pmatrix}
       \mathfrak{r} & s \\
        0 & 0
    \end{pmatrix}
     : \mathfrak{r} \in \sothreelie, s \in \mathbb{R}^3
     \right \}
\end{equation}
Note that the tangent space of $\R^3$ is isomorphic to the space itself so we can simply use the notation $s$ instead of $\mathfrak{s}$.
This lie algebra is isomorphic to $\mathbb{R}^6$ using the map: $\mathfrak{x} \mapsto (\boldsymbol{\omega}, s)$, where we have identified the skew-symmetric matrix $\mathfrak{r} \in \sothreelie$ with its axis-angle representation, $\boldsymbol{\omega} \in \mathbb{R}^3$. As the group of translations, $(\R^3, +)$ is a normal subgroup of $\sethree$, the group can be understood as a semi-direct product: $\sethree = \sothree \ltimes (\R^3, +)$.

\xhdr{Metric on $\sethree$}
Although there are many possible choices for metrics on $\sethree$, none of them are bi-invariant. Instead, one can choose to build a left-invariant or right-invariant metric. A simple choice for the quadratic form $Q$ from \cref{eq:quadratic_form} is setting the matrices $A = C = I_3$ and $B = 0$ \citep{park1994kinematic}, which gives:
\begin{equation}
    Q = 
    \begin{pmatrix}
       I_3 & 0 \\
        0 & I_3
    \end{pmatrix}.
\end{equation}
Using this metric we can define an inner product on $\sethree$ as $ \langle \mathfrak{x}_1 , \mathfrak{x}_2 \rangle_{\sethree} = \text{tr}(\mathfrak{x}_1^\top Q \mathfrak{x}_2)$, where $\text{tr}$ is the trace operation. Writing out the inner product explicitly for $\mathfrak{x}_1, \mathfrak{x}_2 \in \sethreelie$ we get,
\begin{equation}
    \text{tr}(\mathfrak{x}_1^\top Q \mathfrak{x}_2)= \text{tr}
    \begin{pmatrix}
       \mathfrak{r}^\top_1 \mathfrak{r}_2 & \mathfrak{r}^\top_1 \mathfrak{s}_2 \\
        \mathfrak{s}^\top_1 \mathfrak{r}_2 & \mathfrak{s}_1^\top \mathfrak{s}_2
    \end{pmatrix}.
\end{equation}
Thus, we have $\text{tr}(\mathfrak{x}_1^\top Q \mathfrak{x}_2) = \mathfrak{r}^\top_1 \mathfrak{r}_2 +  s^\top s_2$. Therefore, the metric on $\sethree$ decomposes into the metric on $\sothree$ and $\R^3$:
\begin{equation}
\label{eq:sethree_metric}
    \langle \mathfrak{x}_1 , \mathfrak{x}_2 \rangle_{\sethree}  =
    \langle \mathfrak{r}_1 , \mathfrak{r}_2 \rangle_{\sothree} + \langle s_1 , s_2 \rangle_{\R^3}
\end{equation}
This means that we can obtain the geodesics on $\sethree$ from the geodesics on the product manifold $\sothree \times \R^3$:
\begin{equation}
\label{eq:sethree_distance}
    d_\sethree(x_1, x_2) = \sqrt{d_\sothree(r_1, r_2)^2 + d_{\R^3}(s_1, s_2)^2}
\end{equation}
where $x_1=(r_1, s_1), x_2=(r_2, s_2) \in \sethree$, $d_\sothree$ is defined in \cref{eq:dist_so3} and $d_{\R^3}$ is the usual Euclidean distance.

\subsection{The isotropic Gaussian distribution on $\sothree$}
\label{app:isgo}
\xhdr{$\igso$ density} 
The isotropic Gaussian distribution on $\sothree$ is parametrized by a mean, $r \in \sothree$ and a concentration parameter, $\epsilon \in \R$. It can be parametrized in axis-angle, where the axis of rotation is sampled uniformly and the angle of rotation $\omega$ has probability density function (pdf) given by:
\begin{equation}
\label{eq:isgo3}
    f(\omega_x, \eps) = \sum_{l=0}^\infty (2l+1) e^{-l(l+1) \epsilon} \frac{\sin{\big((l+1/2) \omega_x} \big)}{\sin{(\omega_x/2)}}
\end{equation}

Although this expression contains an infinite sum, \citet{Matthies1900} has shown that for $\epsilon \leq 1$, it can be approximated by a closed-form equation:

\begin{equation}
\label{eq:igso3_closed_form}
    f(\omega_x, \eps) = \sqrt{\pi} \epsilon^{-3/2} 
    e^{\frac{\epsilon - \omega^2/\epsilon}{4}} \frac{\left(\omega - e^{- \pi^2 / \epsilon}\left( (\omega - 2\pi) e^{\pi \omega/\epsilon}+ (\omega + 2\pi) e^{-\pi \omega / \epsilon} \right) \right)}{ 2 \sin\left(\frac{\omega}{2}\right)}
\end{equation}

\xhdr{Sampling from $\igso$} 
Sampling from $\igso$ is done following \citet{leach2022denoising}. The angle of rotation is obtained by inverse transform sampling, where the cumulative density function is approximated using the pdf above, scaled by uniform density on $\sothree$ with density $f(\omega) = \frac{1 - \cos{\omega}}{\pi}$; the axis is sampled uniformly from $\mathbb{S}^2$. We note that the closed-form approximation of \cref{eq:igso3_closed_form} makes the computation of the cdf, and hence the sampling process very efficient.

\section{Flow matching in $\R^d$} \label{app:fm_euc}
\label{app:flow_matching_rd}
To perform \foldflow, on $\sethree$, we consider two different flows. One on $\sothree$ that we described in the main paper and another one on $\R^9$ that we describe depending on the consider \foldflow, method.

Riemannian Flow Matching is a generalization of Flow Matching on Riemannian manifold. Therefore, the setting as well as the main ideas are similar and are straightforward to adapt to the Euclidean case. This means that the objective is also to regress a conditional vector field built from conditional probability paths. In this section, we described the conditional probability paths and conditional vector fields that were used respectively by \cite{lipman_flow_2022} and \cite{tong2023improving}.

The main difference is that the conditional probability path is now a Gaussian conditioned on a latent variable $z \sim q(z)$ with variance $\sigma_t$, $\rho_t(s) = \gN(s|z, \sigma_t)$. The conditional vector field has a closed form derived from the following Theorem:

\begin{theorem}[Theorem 3 of \cite{lipman_flow_2022}]\label{thm:gaussian_flow}
The unique vector field whose integration map satisfies $\rho_t(s) = \mu_t + \sigma_t s$ has the form
\begin{equation}\label{eq:gaussian_ut}
u_t(s) = \frac{\sigma_t'}{\sigma_t} (s - \mu_t) + \mu_t',
\end{equation}
\end{theorem}

We now describe the Flow Matching method \cite{lipman_flow_2022} and OT-CFM \cite{tong2023improving, tong2023simulationfree, pooladian_2023_multisample}.

{\bfseries Flow Matching.} In the context of data living in the Euclidean space $\R^d$. Identifying the condition $z$ with a single datapoint $z \coloneqq s_1$, and choosing a smoothing constant $\sigma>0$, one sets
\begin{align}\label{eq: fm_density_vectorfield}
p_t(s | z) &= \mathcal{N}(s\mid t s_1, (t \sigma - t + 1)^2), \\
u_t(s | z) &= \frac{s_1 - (1 - \sigma) s}{1 - (1 - \sigma) t},
\end{align}
which is a probability path from the standard normal distribution ($p_0(x|z)=\gN(x;0,1)$) to a Gaussian distribution centered at $x_1$ with standard deviation $\sigma$ ($p_1(x|z)=\gN(x;x_1,\sigma^2)$). If one sets $q(z)=q(x_1)$ to be the uniform distribution over the training dataset, the objective introduced by \cite{lipman_flow_2022} is equivalent to the CFM objective (\ref{eq:CFM}) for this conditional probability path. 

{\bfseries OT-Conditional Flow Matching \citep{tong2023improving}.} As explained in the main paper, the probability path used in FM is not the optimal transport probability paths between the distributions $\rho_0$ and $\rho_1$. Therefore, we want to get straighter flows for faster inference and more stable training. To achieve that, we leverage the optimal transport theory and want the probability path to be the Euclidean McCann interpolants defined as $\rho_t = t \Psi(s_0) + (1 - t) s_0$. However, as the map $\Psi$ is intractable in practice, we rely on the Brenier theorem which makes a connection between the map $\Psi$ and the optimal transport plan $\pi$. Therefore we set the mean of Gaussian conditional probability path as $\mu_t = t s_1 + (1-t) s_0$ and the latent distribution $q(s_0, s_1) = \pi(s_0, s_1)$.

\begin{align}
p_t(s | s_0, s_1) &= \mathcal{N}(s\mid t s_1 + (1 - t) s_0, \sigma^2),\label{eq:cfm:ptc}\\
p_t(s) &= \int \mathcal{N}(s\mid t s_1 + (1 - t) s_0, \sigma^2) \pi(s_0, s_1) ds_0 ds_1,\label{eq:cfm:pt}\\
u_t(s | z) &= s_1 - s_0. \label{eq:cfm:ut}
\end{align}

In the case of the Euclidean space, the FM loss is equal to $\gL_{\foldflow,-\R^3} = \|v_\theta(t, s) - u_t(s|z)\|$. This can be simplified to down to  $\gL_{\foldflow,-\R^3} = \|v_\theta(t, s) - (s_1 - s_0)\|$. This method is the main inspiration to develop \foldflowot,.


\section{Riemannian Optimal transport}
\xhdr{Optimal transport in generative models}
\looseness=-1
OT has been used in generative models for several approaches. For GANs, it was used as a loss function~\citep{genevay_2018, fatras2021minibatch, salimans2018improving, arjovsky2017wasserstein}. More recently, it was used to speeding up training and inference for continuous normalizing flows~\citep{finlay_how_2020,liu_flow_2023,lipman_flow_2022, tong_trajectorynet_2020, tong2023improving, tong2023simulationfree}, Schr\"odinger bridge models~\citep{shi_diffusion_2023,liu_sb_2023, de_bortoli_diffusion_2021}. In this section, we recall its basic definition over a Riemannian manifold. Then we discuss its empirical computation and we finish this section by proving Proposition 1.

\subsection{Background on Riemannian Optimal transport}\label{app:riemannianOT}

Optimal transport on Riemannian manifold was first studied in the seminal work of \cite{McCann2001PolarFO} and we refer to \cite{villani2003topics, villani2008optimal} for a review of all results. Recently, optimal transport has also drawn attention from the machine learning community, and we now give a longer introduction on this topic.

The (static) Kantorovich optimal transport problem seeks a mapping from one measure to another that minimizes a displacement cost. Formally, we define the $2$-Wasserstein distance between distributions $\rho_0$ and $\rho_1$ on $\gM$ with respect to the cost $c(x,y) = \frac12 d(x,y)^2$ as:
\begin{equation}\label{eq:ot}
W(\rho_0, \rho_1)^2_2 = \inf_{\pi \in \Pi(\rho_0, \rho_1)} \int_{\gM^2} c(x, y)\,d\pi(x, y),
\end{equation}
where $\Pi(\rho_0, \rho_1)$ denotes the set of all joint probability measures on $\gM\times\gM$ whose marginals are $\rho_0$ and $\rho_1$. To compute the optimal transport plan, we rely on the POT library \cite{flamary2021pot}. This problem is a relaxation of the well-known Monge formulation described in the main paper and that we recall now for the sake of readability.

The Monge optimal transport problem is defined as
\begin{equation}\label{app:eq:monge}
\text{OT}(\rho_0, \rho_1) = \inf_{\Psi: \Psi_\#\rho_0=\rho_1} \int_{\gM} c(x, \Psi(x))\,d\rho_0(x).
\end{equation}

When $\gM$ is a smooth compact manifold with no boundary and $\rho_0$ has a density, \cite[Proposition 9]{McCann2001PolarFO} shows that the map $T$ exists and is unique. This is an extension to Riemannian manifold of the well-known Brenier Theorem \citep{Brenier1991PolarFA}. The optimal transport map $\Psi$ and the McCann interpolation  have then the following form:
\begin{equation}
    \Psi(x) = \exp_{x}(-\nabla \phi (x)), \qquad \Psi_t(x) = \exp_{x}(-t\nabla \phi (x)),
\end{equation}
where $\phi$ is a $c$-concave function. Furthermore, we have that the optimal transport plan 
is supported on the graph of the Monge map, \emph{i.e.,} $\pi = (\rm id, \Psi)_\#\rho_0$. Therefore, knowing the transport plan leads to the Monge map. The connection between the two formulations for $\sethree^N_0$ is explicitly stated in Proposition 1 which is proved in~\S\ref{app:proof_prop_1}. However, we first discuss the computation of the OT plan $\pi$.




\paragraph{Minibatch OT approximation.} For empirical distributions, the Kantorovich problem is a linear program and can be efficiently solved with the simplex algorithm. We refer to \cite[Chapter 3]{peyre_computational_2019} for a review on how to solve the Kantorovich problem. However, when we deal with large datasets, computing and storing the transport plan $\pi$ for Optimal Transport (OT) can be challenging due to its cubic time and quadratic memory complexity with respect to the number of samples. To address this, a minibatch OT approximation is often employed. While this approach introduces some error compared to the exact OT solution \citep{fatras20a}, it has been proven effective in various applications such as domain adaptation and generative modeling~\citep{Damodarandeepjdot2018, genevay_2018}. Specifically, during training, for each source and target minibatch, pairs of points are sampled from the optimal transport plan computed between the pair $(x,y) \sim \pi_{\rm batch}$. We empirically show that the batch size can be small compared to the full dataset size and still give a good performance, which aligns with prior studies~\citep{fatras2021minibatch,fatras_unbalanced_2021}. This strategy is also at the heart of the OT-CFM methods~\citep{tong2023improving, tong2023simulationfree, pooladian_2023_multisample}.
\clearpage
\subsection{Proof of Proposition 1}
\label{app:proof_prop_1}
We recall the proposition statement here for convenience and then prove it below.

\begin{mdframed}[style=MyFrame2]
\newriemannianotprop*
\end{mdframed}
\begin{proof}
    The manifold $\sethree$ is a connected, complete, ($\gC^\infty$) smooth manifold without boundary.  $\sethreen$ (equipped with the usual product distance) is a finite Cartesian product of connected, complete, smooth manifolds without boundary and therefore it is itself a connected, complete, smooth manifold without boundary. We only need to check these assumptions are also satisfied by $\sethreenzero$. We do so by noting that $\sethreenzero$ can be written as $N-1$ copies of $\sethree$ where the $\R^3$ component is mean subtracted---i.e. $s^c = s - 1/N \sum_{i=1}^N s^i$, and the final $N$th element in the product is the mean, $1/N \sum_{i=1}^N s_i$. Certainly, the first $N-1$ components satisfy connectedness, and completeness, and are manifolds that are smooth without boundary. Furthermore, the disintegration of measures on $\sethreenzero$ (Proposition 3.5~\citep{yim2023se}) allows us to define a measure $\bar{\mu}$ proportional to $\R^3$ for the final component $N$th component. Therefore, by our assumptions on the measures $\rho_0, \rho_1$, we can apply the following Theorem from \citet{villani2003topics} to get the desired results.
\begin{theorem}[Theorem 2.47, \cite{villani2003topics}]\label{thm:mccan_villani}
    Let $\gM$ be a connected, complete and smooth ($\gC^3$) Riemannian manifold without boundary, equipped with its standard volume measure. Let $\rho_0, \rho_1$ be two compactly supported distributions and set the ground cost $c(x,y) = \frac12 d(x,y)^2$ with $d$ the geodesic distance on $\gM$. Further, assume that $\rho_0$ is absolutely continuous with respect to the volume measure on $\gM$. Then the Kantorovich and Monge problems admit a unique solution that is connected as follows $\pi = (id \times \Psi)_\#\rho_0$, where $\Psi$ is almost uniquely determined everywhere $\rho_0$. Furthermore we have that $\Psi(x) = \exp_x(\nabla \phi(x))$ for some $d^2$-concave function $\phi$.
\end{theorem}
\end{proof}

\section{Stochastic Riemannian flow matching}
\label{app:sfm}

\subsection{Brownian Bridge on $\sothree$}
\label{app:brown_bridge_so3}
We follow the presentation in \cite{jensen2022bridge} to define the Brownian bridge on a Lie group $G$ endowed with a metric. We note that $\log$ is the inverse of the Riemannian exponential map. However, if the metric is bi-invariant, which is the case for $\sothree$, it coincides with the Lie group logarithm. We can simulate a bridge on $G$ via the guided diffusion SDE (using $\circ$ for the Stratonovich integral), for a process conditioned to reach $v$ at $t=1$.
\begin{equation}
    \ddd \rm R_t = -\frac{1}{2} V_0(\rm R_t)\dt + V_i(\rm R_t) \circ \left( d\rm B_t^i - \frac{\log_{\rm R_t}(v)^i}{1-t}\dt \right) \quad \rm R_0 = r_0,
\end{equation}
where $V_i(xr = (dL_r)_e v_i$ with $\{v_1,\dotsc,v_d\}$ an orthonormal basis of $T_e G$, and where $\rm B_t$ is a Brownian motion on $G$. On $\sothree$, since the metric is bi-invariant, we have $V_0 = 0$. In this work, we model the guided bridge with a diffusion that does not depend on the process $\rm R_t$. In this case, the Stratonovich and Itô formulations are the same, yielding the reversed process defined in \eqref{eq:r_bridge}. 

\subsection{Simulation-Free Approximation of Brownian Bridges on $\sothree$}
\label{app:numerical_approximation_of_bridges}
\begin{figure}[htbp]
    \centering
    \begin{subfigure}[b]{0.3\textwidth}
        \includegraphics[width=\textwidth]{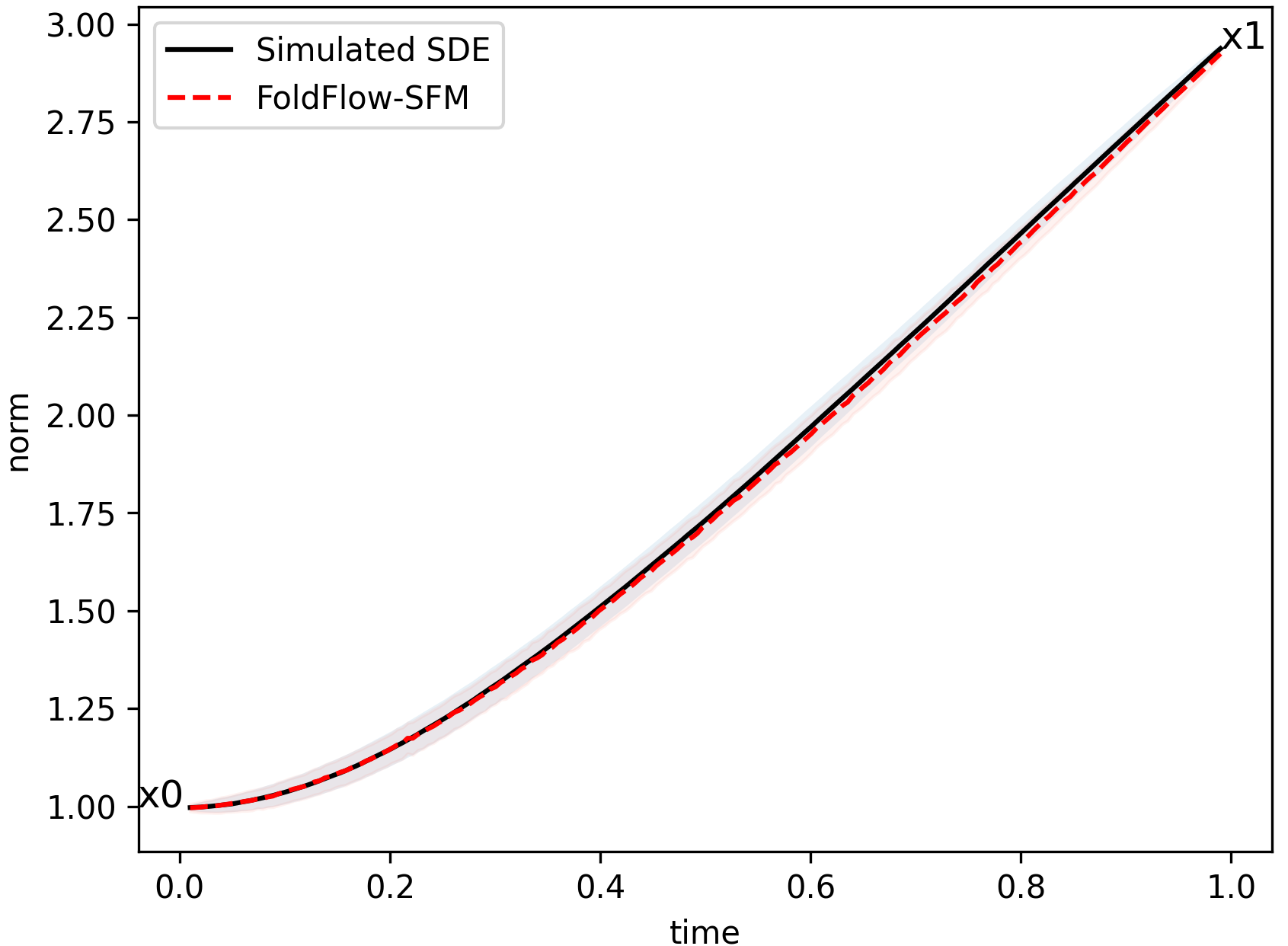}
        \caption{$\gamma=0.1$}
        \label{fig:bridge_approximation:a}
    \end{subfigure}
    \hspace{-2pt}
    \begin{subfigure}[b]{0.3\textwidth}
        \includegraphics[width=\textwidth]{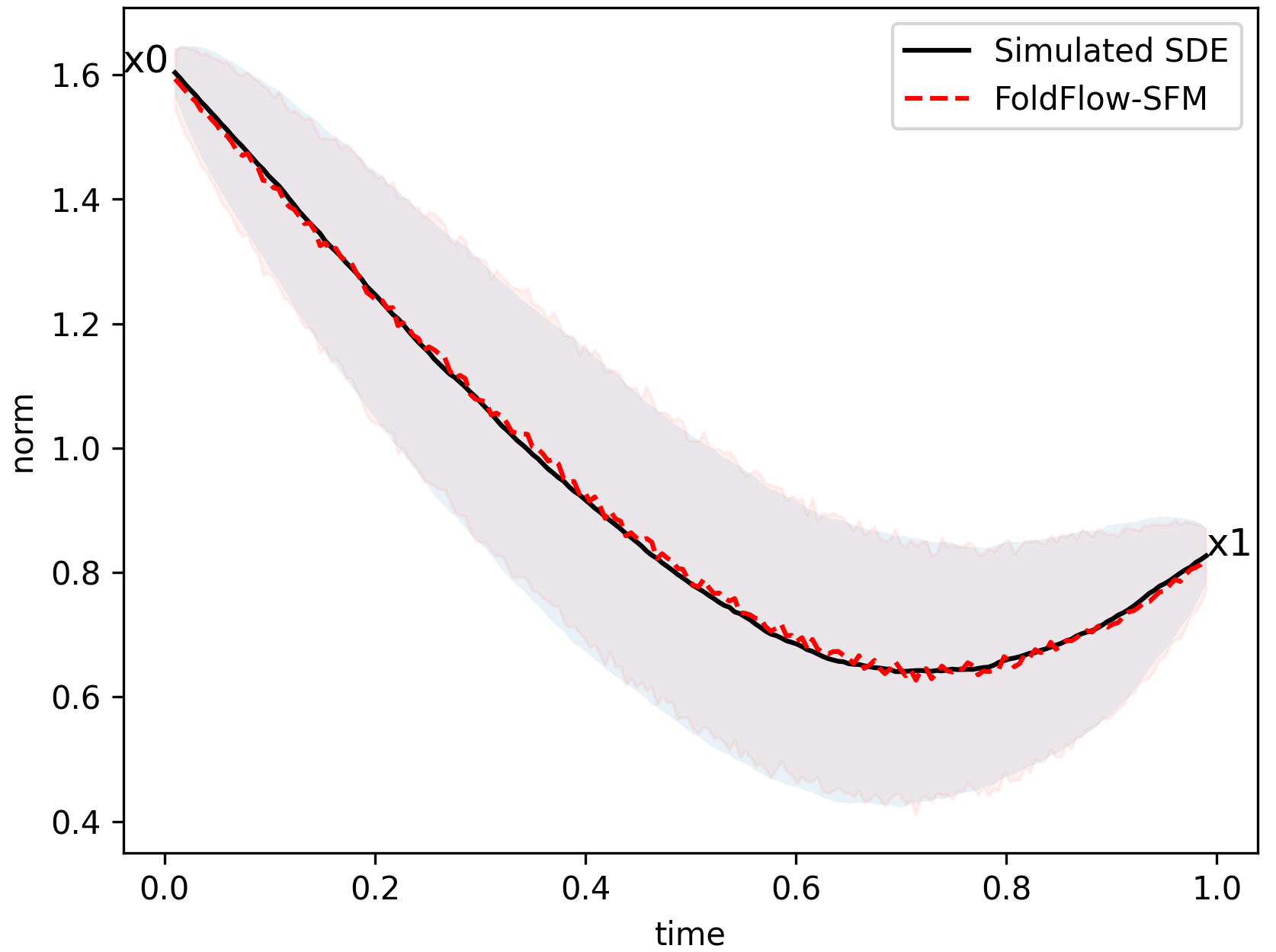}
        \caption{$\gamma=0.5$}
        \label{fig:bridge_approximation:b}
    \end{subfigure}
    \hspace{-1pt}
    \begin{subfigure}[b]{0.3\textwidth}
        \includegraphics[width=\textwidth]{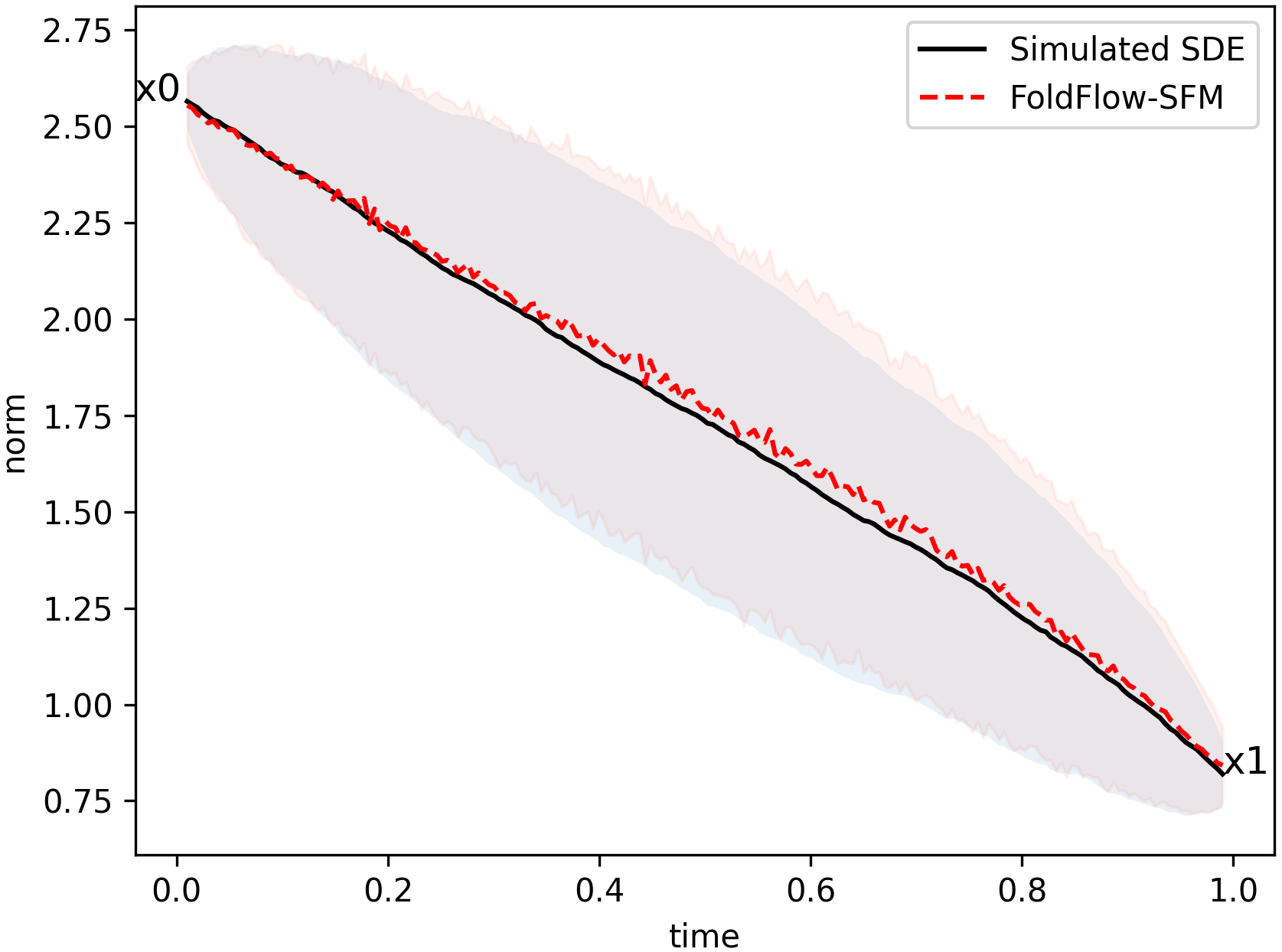}
        \caption{$\gamma=1.0$}
        \label{fig:bridge_approximation:c}
    \end{subfigure}
    \caption{\small Numerical comparison between the simulation-free of the SDE in \foldflowsfm, vs. simulated Brownian bridge on $\sothree$, for different values of the diffusion coefficient, $\gamma$.}
    \label{fig:bridge_approximation}
\end{figure}

We now numerically investigate the fidelity of our simulation-free SDE which is employed in \foldflowsfm, in relation to the guided drift SDE in \cref{eq:sfm-sothree}. In \cref{fig:bridge_approximation} we plot the mean and the standard deviation (over 1024 data points) of the distribution of the $\sothree$-norm along the trajectory against time, for three different values of the diffusion coefficient, $\gamma$. We find the true simulated Brownian bridge (bold black line) is in close proximity to the simulation-free \foldflowsfm, SDE (red dotted lines). We further note that this holds for the entire trajectory and leads to overlapping shaded regions that correspond to the standard deviation of the norm. This result adds empirical substantiation to using the \foldflowsfm, as a drop-in and fast approximation to the Brownian bridge SDE on $\sothree$.

\subsection{Proof of Proposition 2}\label{app:proof_proptwo}
\begin{mdframed}[style=MyFrame2]
\proptwo*
\end{mdframed}
\vspace{-20pt}
\proof{
Let $u_t = \mathbb{E}_{\rho(z)} \left[\frac{\rho_t(x | z)}{\rho_t(x)}u_t(x | z)\right]$, for $x \in \sethreen_0$. We claim that:

\begin{equation}
     \nabla_\theta \mathbb{E}_{z \sim \rho(z), x \sim \rho_t(x|z)} \left[ || v_\theta(t, x) - u_t(x | z) ||^2_{\sethreen_0} \right] = 
     \nabla_\theta \mathbb{E}_{x \sim \rho_t(x)} \left[ || v_\theta(t, x) - u_t(x) ||^2_{\sethreen_0} \right]
\end{equation}

From disintegration of measures (\citet{pollard2002user} and Proposition 3.5 form \citet{yim2023se}), we know the probabilities $\rho_t(x) \propto \rho_t(r^1) \cdots \rho_t(r^N) \rho_t(s^1) \cdots \rho_t(s^N)$, and similar for the conditional probability $\rho_t(x|z)$. 
Given that by \cref{eq:sethree_metric}, the metric on $\sethree$ also factorizes into metric on $\sothree$ and $\R^3$, it suffices to prove the claim for $\sothree$ and $\R^3$. The claim can therefore be stated as follows, where we have written $r^i$ as $r$ and $s^i$ as $s$ for conciseness. 

\begin{equation}
     \nabla_\theta \mathbb{E}_{z \sim \rho(z), x \sim \rho_t(r|z_r)} \left[ || v_\theta(t, r) - u_t(r | z_r) ||^2_{\sothree} \right] = 
     \nabla_\theta \mathbb{E}_{r \sim \rho_t(r)} \left[ || v_\theta(t, r) - u_t(r) ||^2_{\sothree} \right]
\end{equation}

\begin{equation}
\label{eq:prop2_so3}
     \nabla_\theta \mathbb{E}_{z \sim \rho(z), s \sim \rho_t(s|z_s)} \left[ || v_\theta(t, s) - u_t(s | z_s) ||^2_{\R^3} \right] = 
     \nabla_\theta \mathbb{E}_{s \sim \rho_t(s)} \left[ || v_\theta(t, s) - u_t(s) ||^2_{\R^3} \right]
\end{equation}

The proof of this claim follows a similar structure to \citet{chen2023riemannian}. We proceed by proving \cref{eq:prop2_so3}. Dropping the distributions for conciseness, we have:

\begin{equation}
    \begin{split}
        &\nabla_\theta \left( \mathbb{E}_{z_r, r} [ || v_\theta(t, r) - u_t(r | z_r)||^2 ] - \mathbb{E}_{z_r, r} [ || v_\theta(t, r) - u_t(r)||^2 ] \right) \\
        &= \nabla_\theta \left( -2\mathbb{E}_{z_r, r} \langle v_\theta(t, r), u_t(r | z_r)\rangle_\sothree - 2\mathbb{E}_{r} \langle v_\theta(t, r), u_t(r)\rangle_\sothree \right)
    \end{split}
\end{equation}
}
Now:

\begin{equation}
    \begin{split}
        \mathbb{E}_{r} \langle v_\theta(t, r), u_t(r)\rangle &= \int_0^1 \int_\sothree  \langle v_\theta(t, r), u_t(r)\rangle_\sothree \rho_t(r) d\mathrm{vol}_r \\
        &= \int_0^1 \left \langle v_\theta(t, r) \int_\sothree \frac{\rho_t(r | z_r)}{\rho_t(r)} u_t(r | z_r) \rho(z_r) d\mathrm{vol}_{z_r} \right \rangle \rho_t(r) d_r \\
        &=  \int_0^1  \int_\sothree \langle v_\theta(t, r), u_t(r|r_z) \rangle \rho_t(r | r_z) \rho(r_z) d\mathrm{vol}_{r} d\mathrm{vol}_{z_r} \\
        &= \mathbb{E}_{r, z_r} \langle v_\theta(t, r) u_t(r | z_r) \rangle
    \end{split}
\end{equation}

The proof for $\R^3$ directly follows Theorem 3.2 from \cite{tong2023simulationfree}.

\section{Extended Figure Information}
\subsection{\Cref{fig:simulation_on_sphere}} depicts the probability paths of \foldflowbase,, \foldflowot,, and \foldflowsfm, projected onto $\mathbb{S}^2$. Where \foldflowbase, paths may cross, \foldflowot, conditional paths do not cross reducing the variance in the objective stabilizing training as studied in \cite{pooladian_2023_multisample,tong2023improving}. \foldflowsfm, adds in stochasticity which improves novelty in our protein generation task. \Cref{fig:simulation_on_sphere} also contains a table summarizing the differences between methods. Showing whether or not they can map from a general source distribution, can perform optimal transport under some conditions, are stochastic or deterministic, and do not require calculation of the score. We note that there is a $^*$ for \foldflowsfm, performing OT, as it only achieves OT when noise goes to zero and it recovers \foldflowot. However, this bias may still be helpful in reducing the variance of the objective function even if OT is not achieved.

\section{$\sothree$ Toy Experiment}

\subsection{Toy Model Parameterization}
\label{app:toy_vfield}
For the vector-field parametrization, the goal is to create a function that by construction lies on the tangent space of the manifold. For the toy experiments, this is done by using a $3$-layer MLP, and projecting the output of the network to the tangent space of the input. That is, similar to \citet{chen2023riemannian}, we have:
\begin{equation}
    u_\theta(t, r) = \Pi_r \mathrm{MLP}(t, r),
\end{equation}
where $\Pi_r(M)$ projects a $3 \times 3$ matrix onto $\mathcal{T}_r\sothree$. This operation essentially computes the skew-symmetric component of $M$, given by $\frac{M - M^\top}{2}$ and parallel transports it to the tangent space of $r$ using left matrix multiplication which is the group operation on $\sothree$.

\subsection{Additional Results for $\sothree$ Toy Experiments}\label{app:toy_para}
\begin{figure}[htbp]
    \centering
    \begin{subfigure}[b]{0.24\textwidth}
        \includegraphics[width=\textwidth]{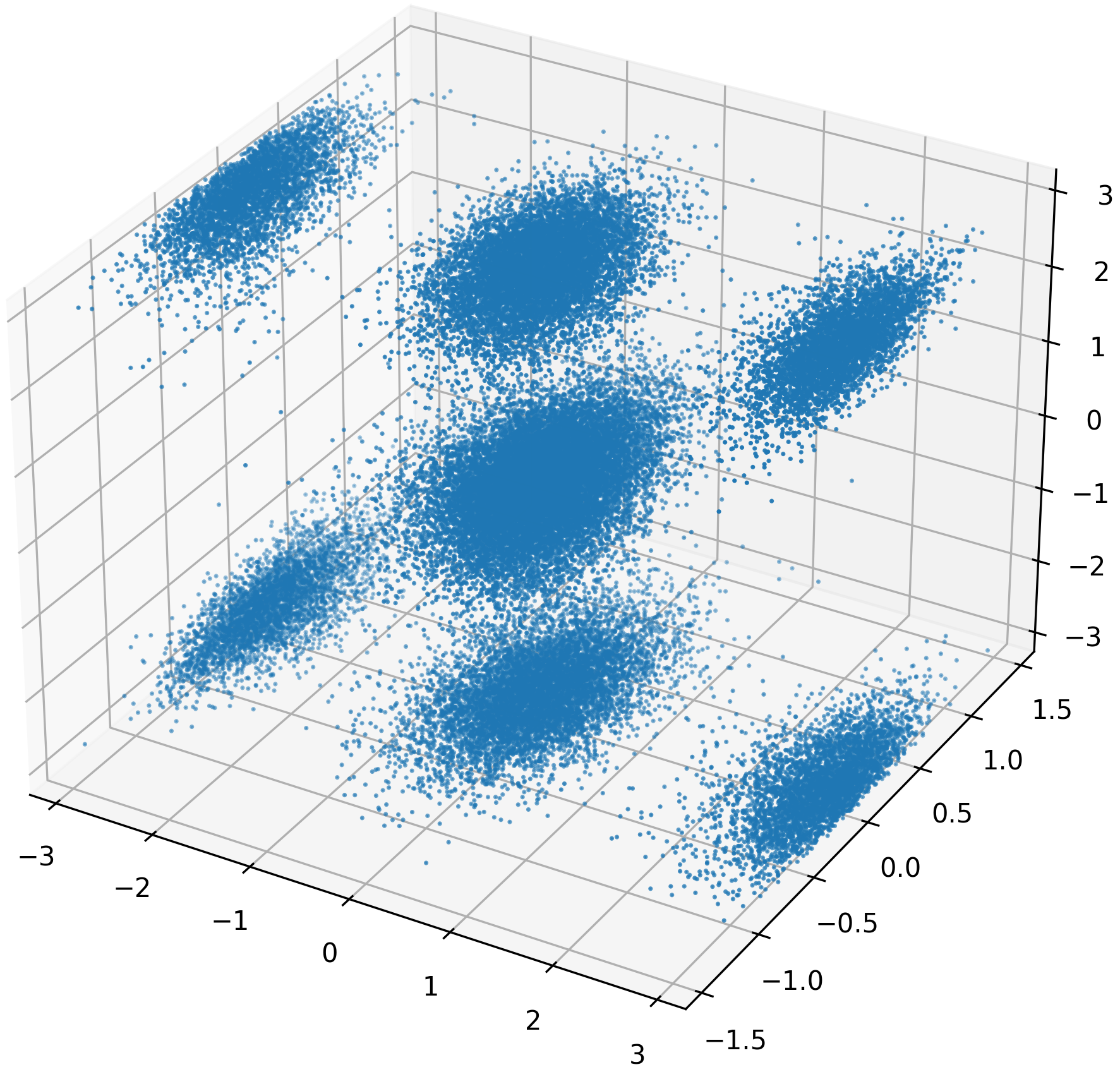}
        \caption{}
        \label{fig:1a}
    \end{subfigure}
    \hspace{-2pt}
    \begin{subfigure}[b]{0.24\textwidth}
        \includegraphics[width=\textwidth]{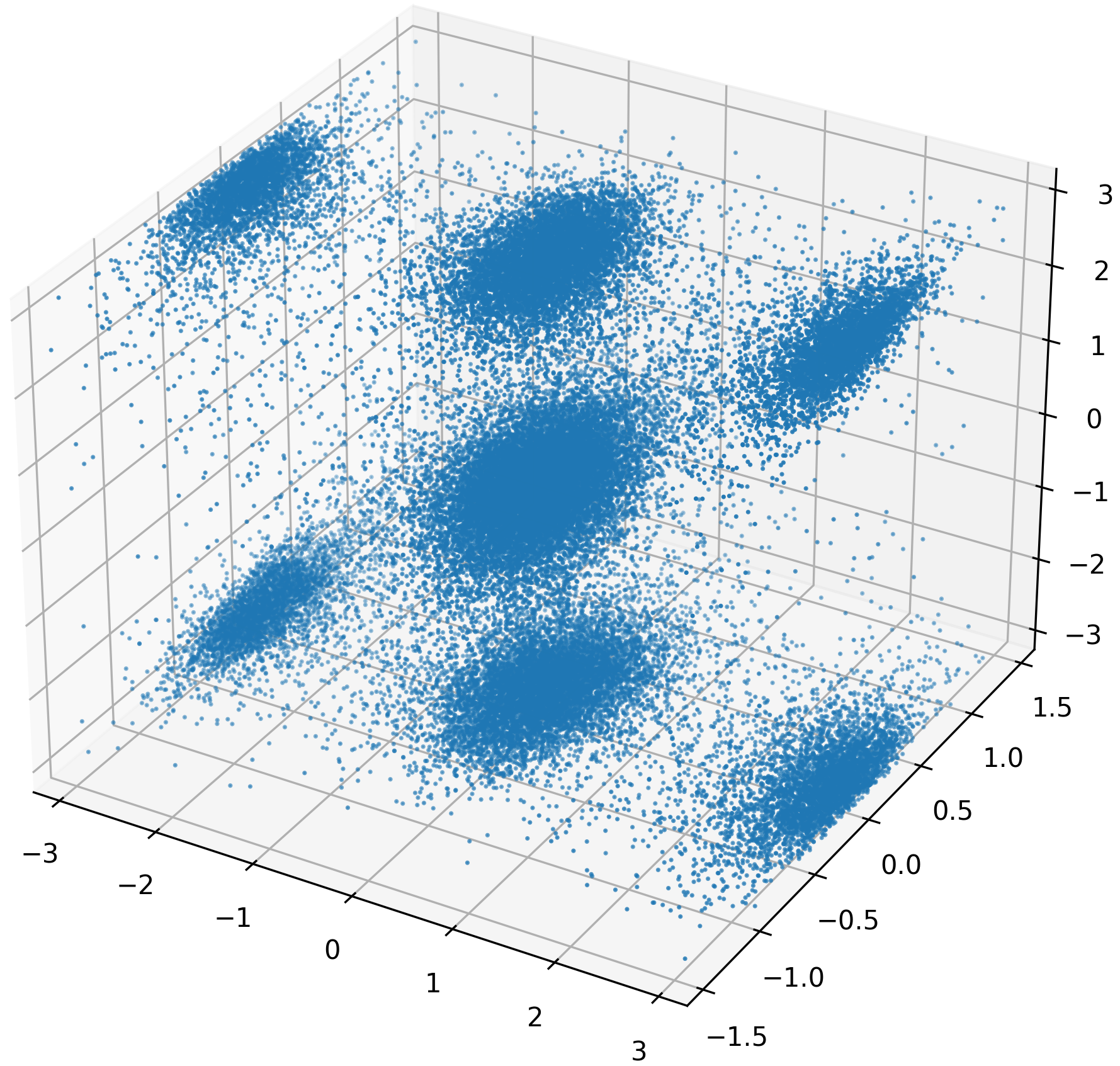}
        \caption{}
        \label{fig:1b}
    \end{subfigure}
    \hspace{-1pt}
    \begin{subfigure}[b]{0.24\textwidth}
        \includegraphics[width=\textwidth]{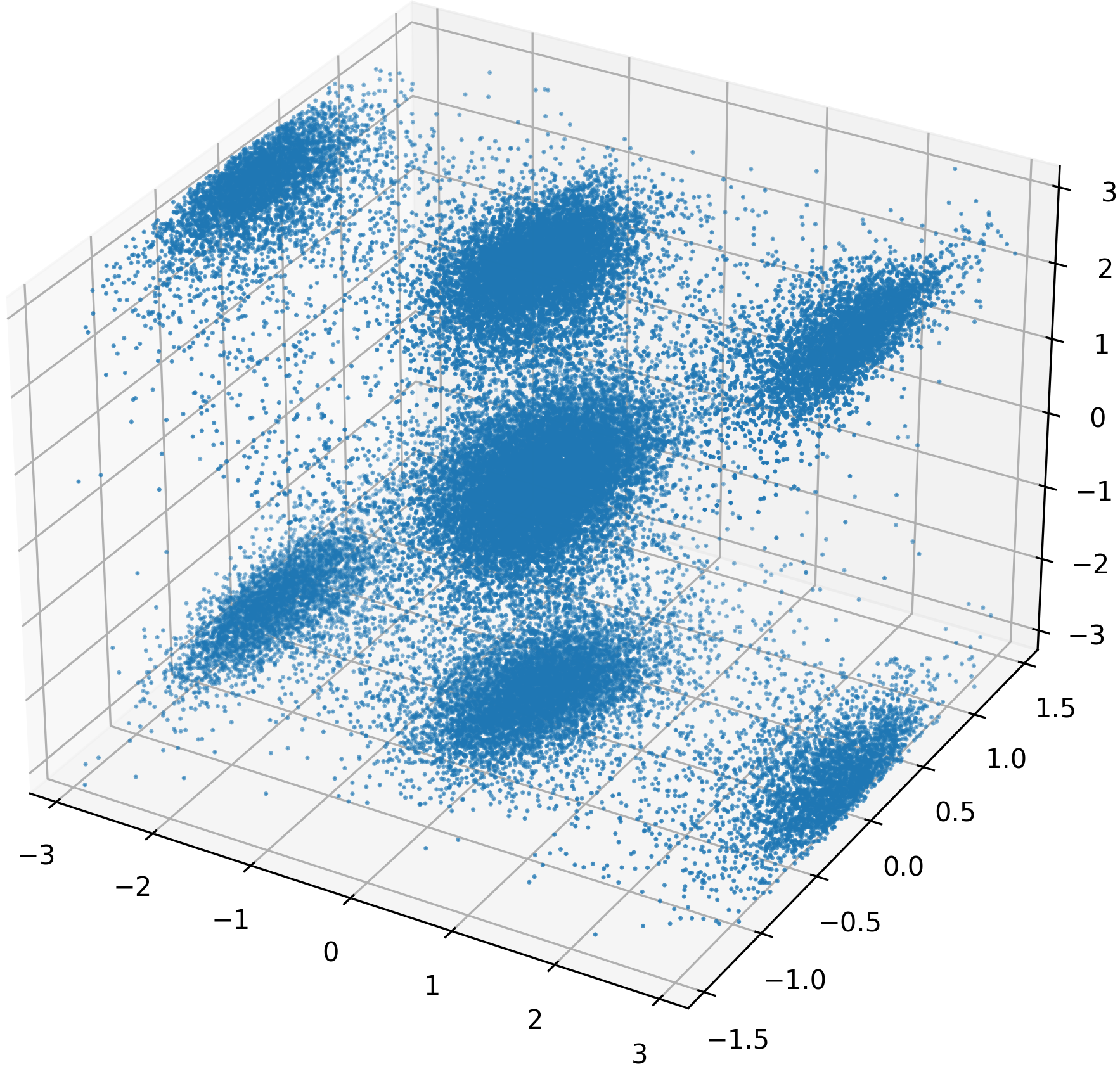}
        \caption{}
        \label{fig:1c}
    \end{subfigure}
    \hspace{-1pt}
    \begin{subfigure}[b]{0.24\textwidth}
        \includegraphics[width=\textwidth]{Figures/inference_sfm.png}
        \caption{}
        \label{fig:1d}
    \end{subfigure}
    \caption{(a) Data distribution (b) \foldflowbase, (c) \foldflowot, (d) \foldflowsfm,. The data is visualized using the Euler-angle representation of the rotation matrices.}
    \label{fig:toy_results}
\end{figure}

In this section, we present the qualitative results of our toy experiments. In \cref{fig:toy_results}, we can see that all the three models, \foldflowbase,, \foldflowot, \foldflowsfm, learn to correctly model the modes of the ground-truth distribution with a slight model shrinkage in the \foldflowbase,.

\section{Additional Results and Analysis for the Protein Experiments}

\subsection{Protein Backbone Generation Experiment Additional Results}
\label{app:protein_experiments}

\xhdr{Empirical investigation of rotation norms for Inference Annealing}
\begin{figure}
    \centering
    \includegraphics[width=0.5\linewidth]{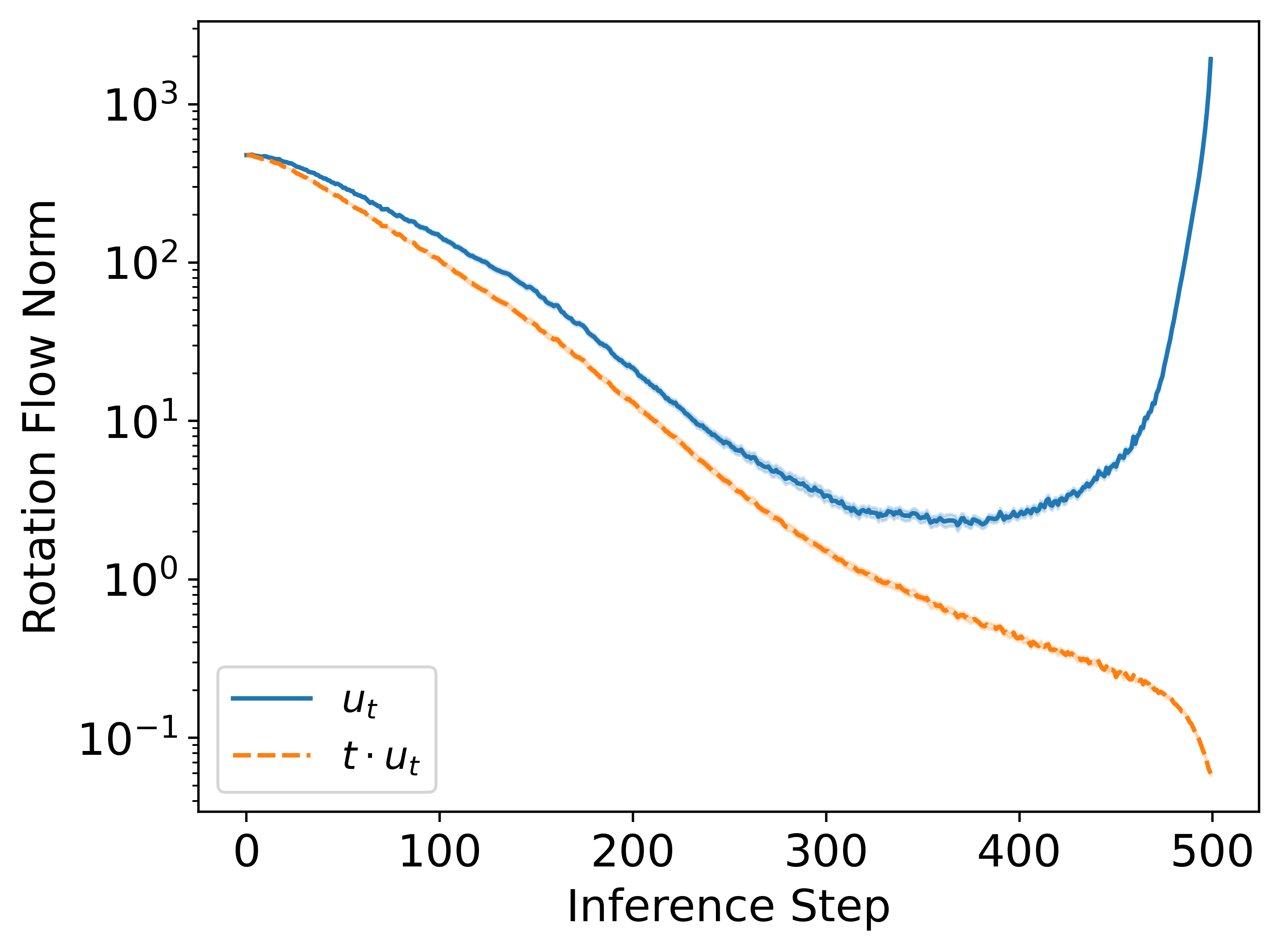}
    \caption{Norm of the rotation flow with and without  $t$ scaling.}
    \label{fig:inference_fig1}
\end{figure}

\begin{figure}[htbp]
\captionsetup[subfigure]{aboveskip=-2pt,belowskip=-2pt}
    \centering
    \begin{subfigure}[b]{0.32\textwidth}
        \includegraphics[width=\textwidth]{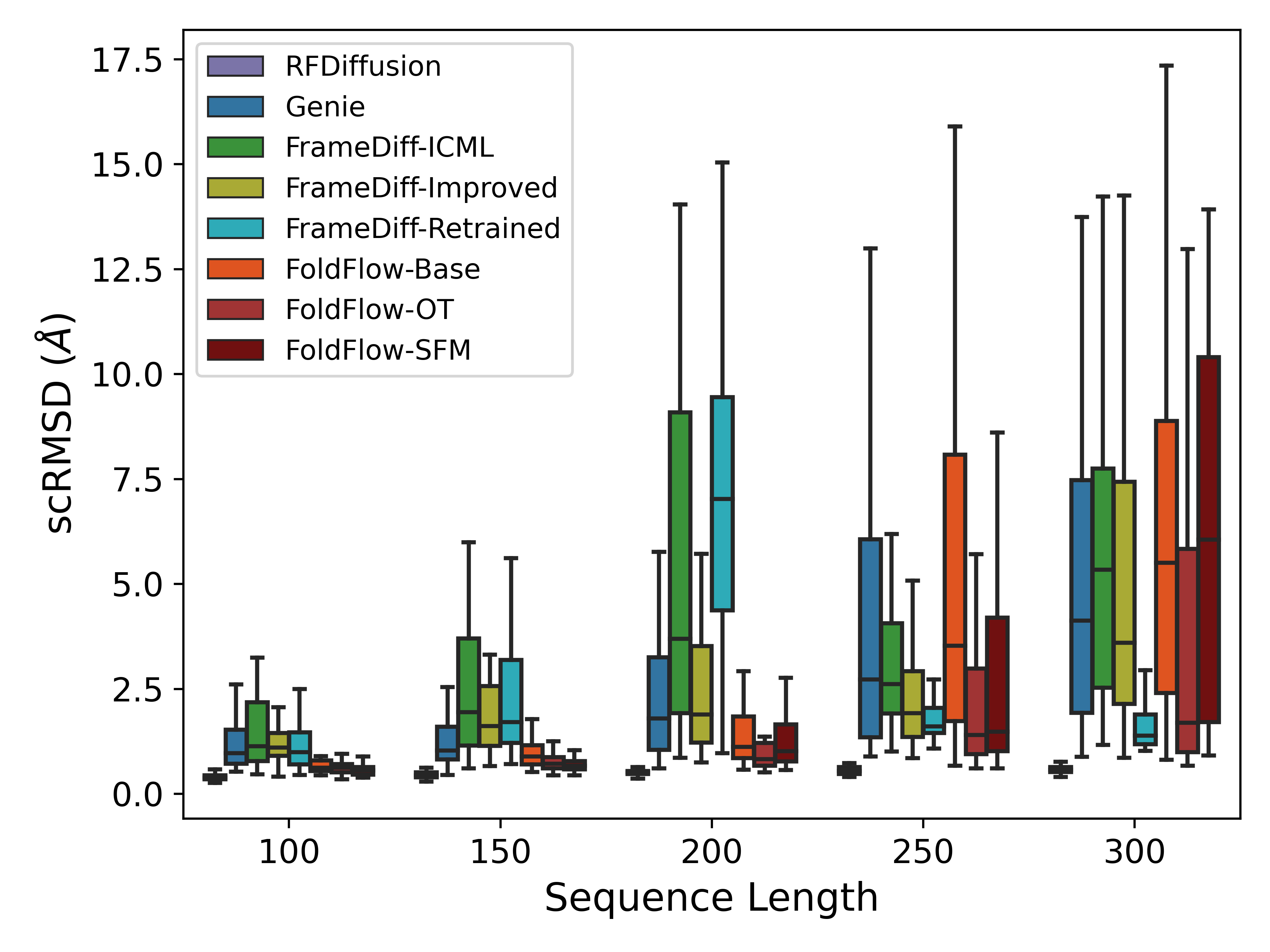}
        \caption{}
        \label{fig:protein_lengths_designability}
    \end{subfigure}
    \hspace{-2pt}
    \begin{subfigure}[b]{0.32\textwidth}
        \includegraphics[width=\textwidth]{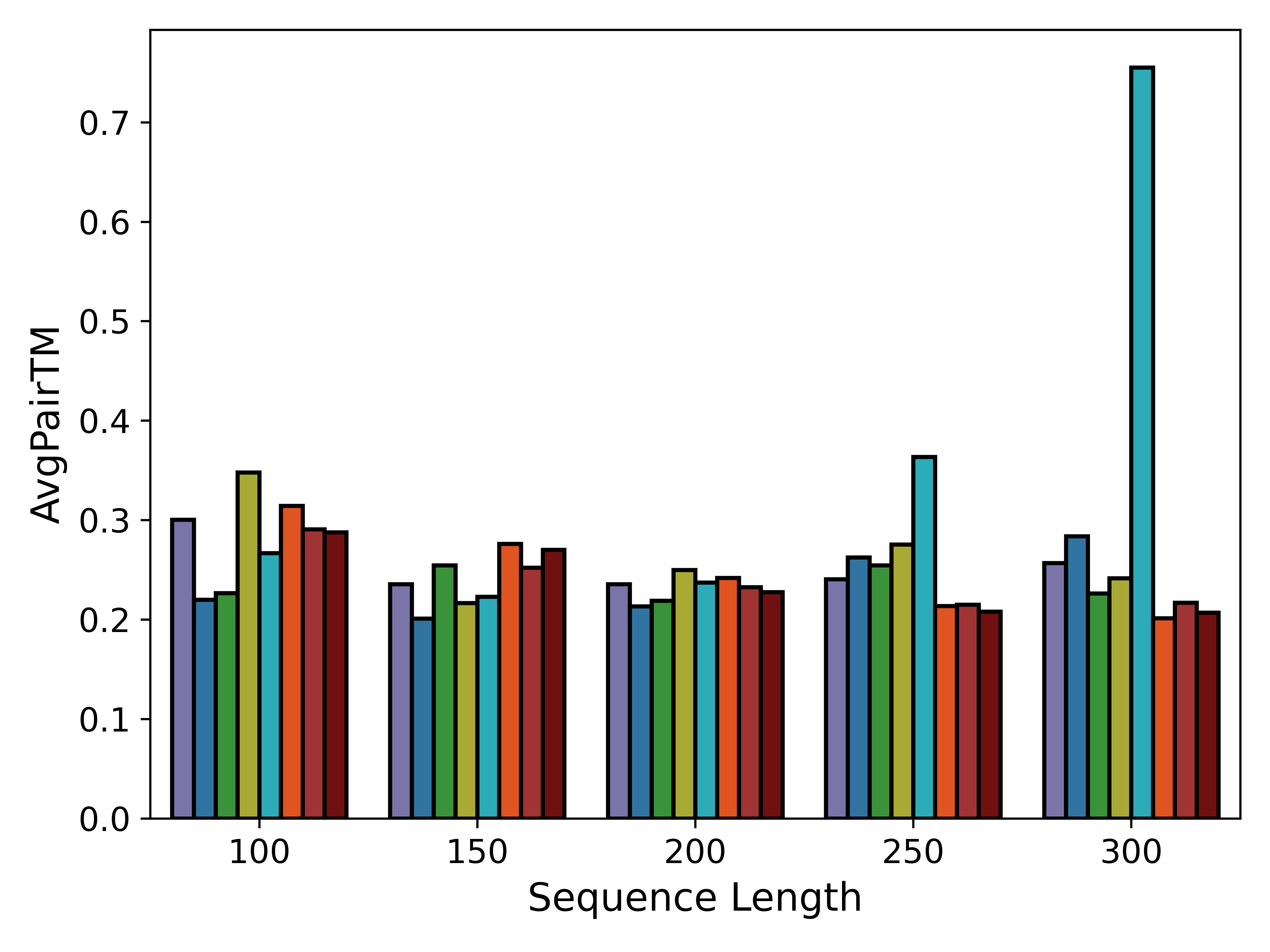}
        \caption{}
        \label{fig:protein_lengths_diversity}
    \end{subfigure}
    \begin{subfigure}[b]{0.32\textwidth}
        \includegraphics[width=\textwidth]{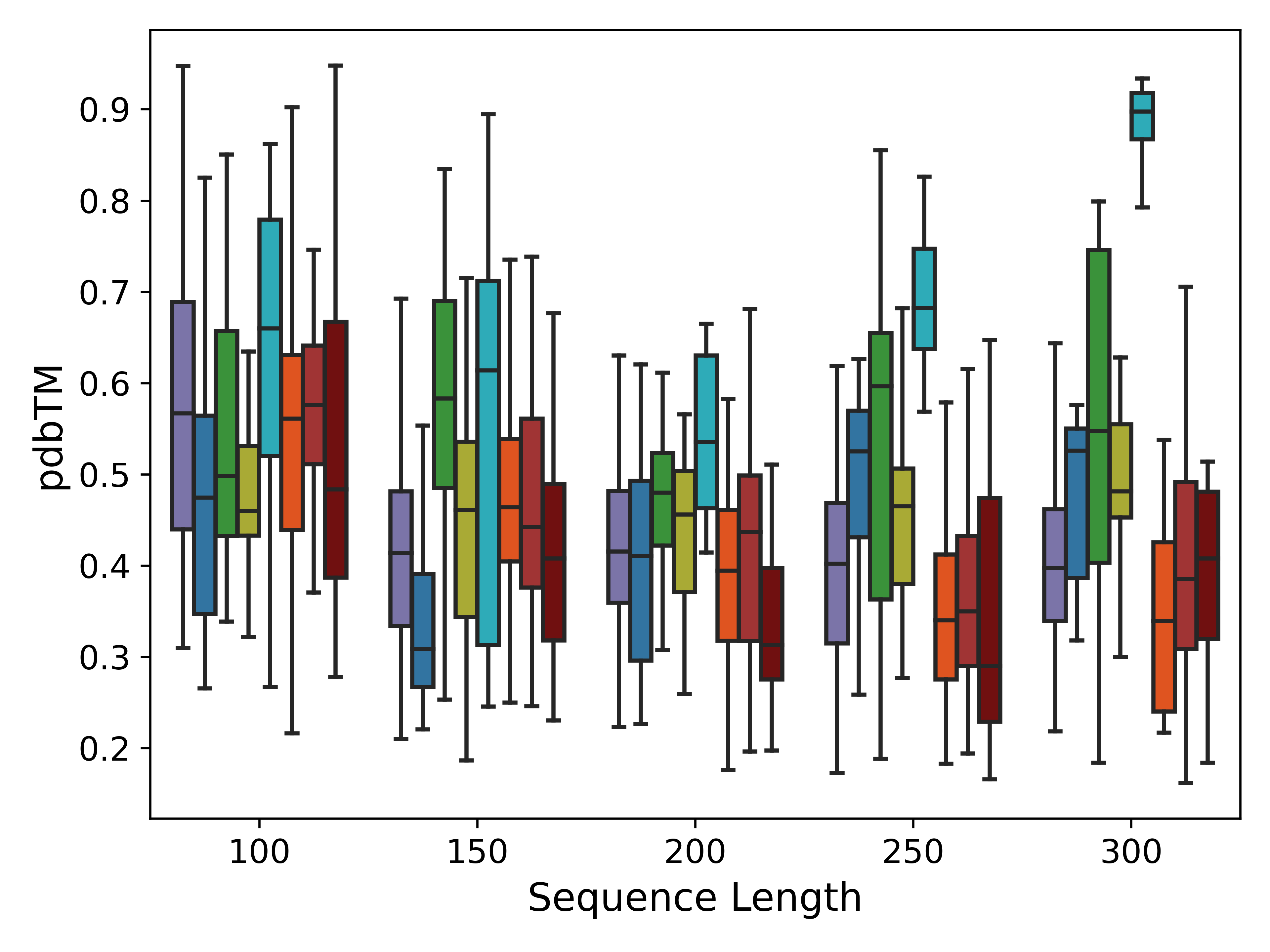}
        \caption{}
        \label{fig:protein_lengths_novelty}
    \end{subfigure}
    \hspace{-1pt}
    \vspace{-5pt}
    \caption{\small (a) Designability as quantified by scRMSD (lower is better), (b) Diversity as quantified by average pairwise TMScore (lower is better), and (c) Novelty of proteins as quantified by maximum TMScore to PDB (lower is better), designed across lengths for \foldflow, models and previous state of the art models.}
    \vspace{-5pt}
    \label{fig:protein_lengths}
\end{figure}

\begin{figure}[htbp]
    \centering
    \includegraphics[width=1\textwidth]{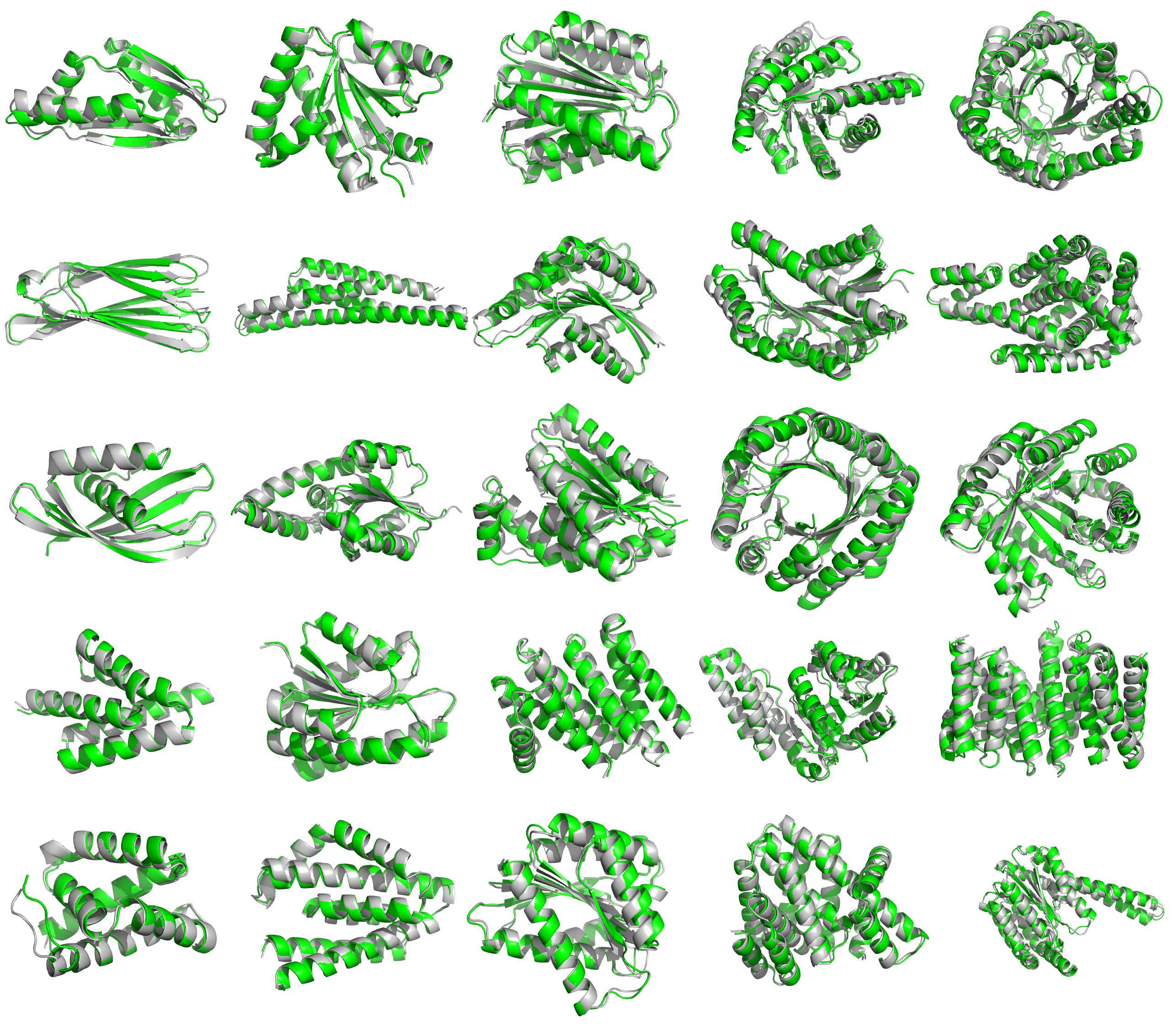}
    \caption{More \foldflowsfm, generated structures in green compared to ProteinMPNN --> ESMFold refolded structures in grey. 5 samples all with RMSD $<2 \angstrom$ for lengths 100, 150, 200, 250, 300 from left to right. \foldflowsfm, generates designable diverse proteins.}
    \label{fig:sfm-refold_appendix}
\end{figure}

In \cref{fig:sfm-refold_appendix}, we show five proteins generated by \foldflowsfm, from each backbone length. Here we show the generated structure in green and the best ESM-refolded structure out of eight sequences generated using ProteinMPNN. We can see that \foldflowsfm, generates diverse folds that refold with diversity in secondary structure and overall 3D conformation. 

In \cref{fig:protein_lengths} we compare the performance of models on designability, diversity, and novelty tasks for different backbone lengths. In particular, we can see that \foldflow, closes the gap between models without pretraining (Genie, FrameDiff, \foldflow,) and RFDiffusion in terms of designability, particularly on shorter sequences ($\le 200$).

We note there is a trade-off between designability and diversity/novelty, both at the short sequence lengths and as sequence length increases. For longer sequences (250, 300), while \foldflow, models are comparable in terms of designability, they generate significantly more diverse and novel structures as compared to all other models, even RFDiffusion (although RFDiffusion still generates significantly more designable proteins at these lengths.

\subsubsection{Timing comparison in \Cref{tab:main} and \Cref{tab:training_time}}

In \Cref{tab:main} we compare the number of steps per second for each model, where a step corresponds to a forward and backwards pass on the effective batch size as defined in \Cref{eq:batch_size} on a single GPU. Here we find that \foldflow, is over 2x faster than \rev{FrameDiff} per step. This drastic improvement is due to a number of optimizations, with the largest being that we can avoid the costly $\igso$ score computation which is necessary for their method. 

We train our model in Pytorch using distributed data-parallel (DDP) across four NVIDIA A100-80GB GPUs for roughly 2.5 days. We note that this is substantially less than comparable models (\cref{tab:training_time}). RFDiffusion requires the use of pre-trained weights from RosettaFold which trained for 4 weeks on 64 V100 GPUs~\citep{watson_novo_2023}.

\begin{table}[htb]
\small
    \centering
    \caption{Training resources for protein generation models.}
    \begin{tabular}{l|rrrr}
        \toprule
        Model & Training time & Optimization Steps & \#gpus & Distributed Training\\
        \midrule
        RFDiffusion & 28 + 3 days  & --- & 64 + 8 & ---\\
        RFDiffusion w/o pretraining & 3 days & --- & 8 & ---\\
        Genie (SwissProt) & $\sim$8 days & $\sim$800k & 6 & DP\\
        FrameDiff-ICML & $\sim$7 days & $\sim$1.9m & 2 & DP\\
        FrameDiff-Improved &  +7 days & +1.9m & 2 & DP \\
        \rev{FrameDiff-Retrained}& \rev{10 days} & \rev{2.2m} & \rev{2} & \rev{DP} \\ 
        \foldflow, \rev{(BASE, OT and SFM)} & $\sim$2.5 days & 600k & 4 & DDP \\
        \bottomrule
    \end{tabular}

    \label{tab:training_time}
\end{table}

\subsection{Equilibrium Conformation Generation Experiment}
\label{app:eq_conformation_gen}
\begin{figure}[htbp]
\captionsetup[subfigure]{aboveskip=-2pt,belowskip=-2pt}
    \centering
    \begin{subfigure}[b]{0.48\textwidth}
        \includegraphics[width=\textwidth]{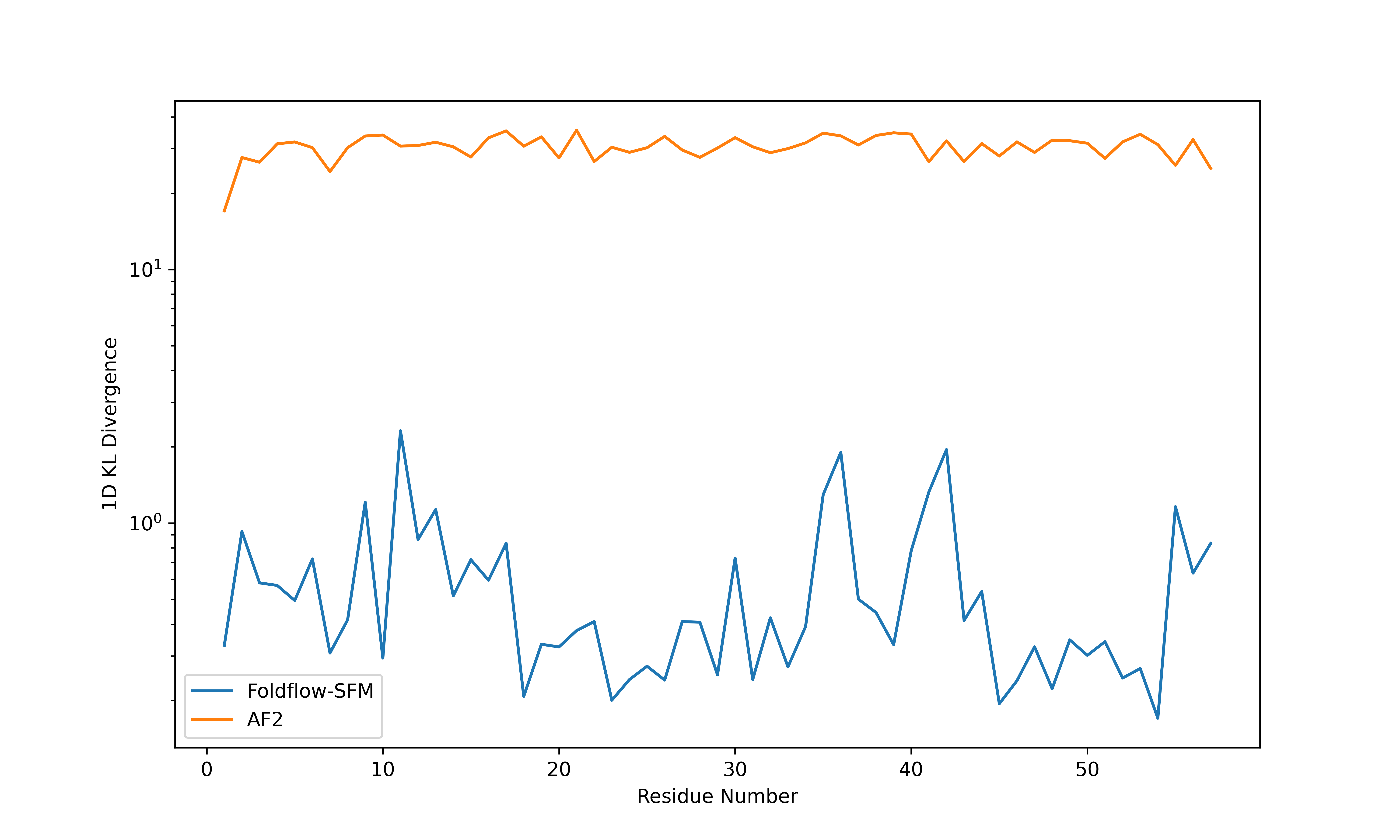}
        \caption{}
        \label{fig:md_figs:kld_1d}
    \end{subfigure}
    \hspace{-2pt}
    \begin{subfigure}[b]{0.48\textwidth}
        \includegraphics[width=\textwidth]{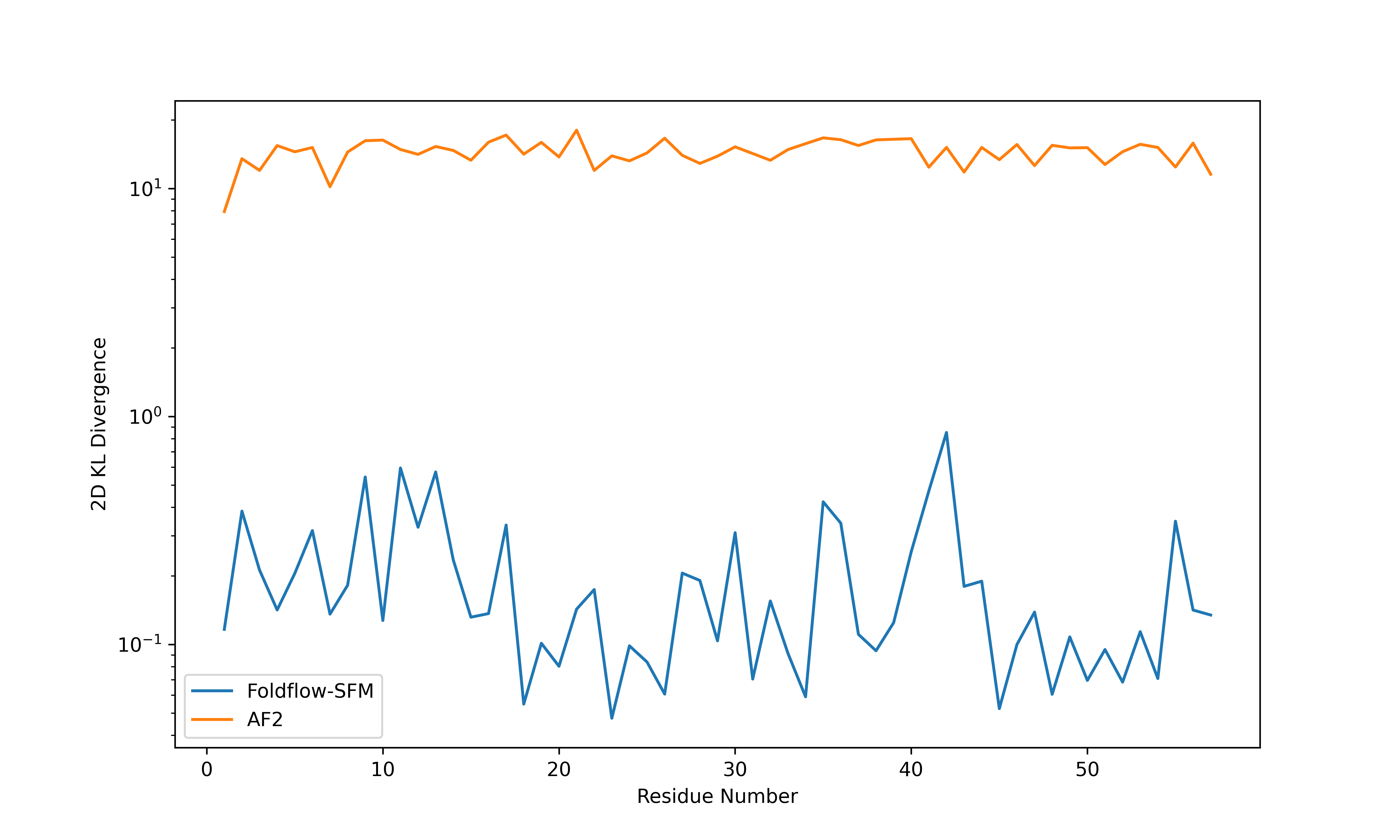}
        \caption{}
        \label{fig:md_figs:kld_2d}
    \end{subfigure}
    \hspace{-1pt}
    \vspace{-5pt}
    \caption{\small KL divergence per residue of the 2D dihedral angle ($\Phi$ and $\Psi$) distributions between the samples from \foldflow, and test MD frames (blue) and AlphaFold 2 and the test MD frames (orange).}
    \vspace{-5pt}
    \label{fig:app_md_kl}
\end{figure}

\begin{figure}[htbp]
\captionsetup[subfigure]{aboveskip=-2pt,belowskip=-2pt}
    \centering
    \begin{subfigure}[b]{0.32\textwidth}
        \includegraphics[width=\textwidth]{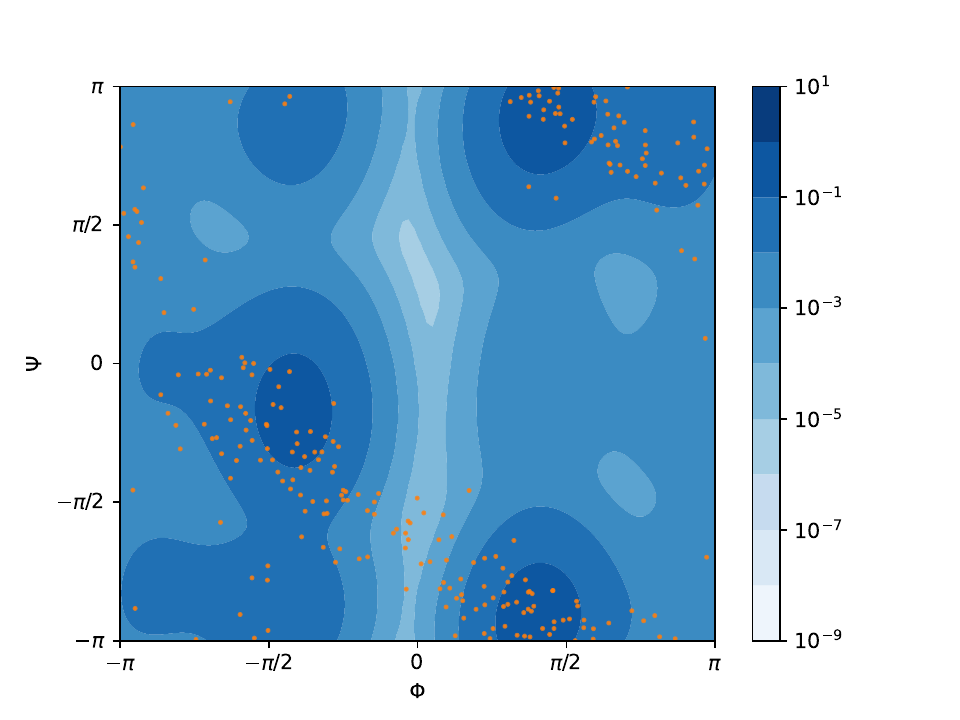}
        \caption{\foldflow,}
        \label{fig:rama1}
    \end{subfigure}
    \hspace{-2pt}
    \begin{subfigure}[b]{0.32\textwidth}
        \includegraphics[width=\textwidth]{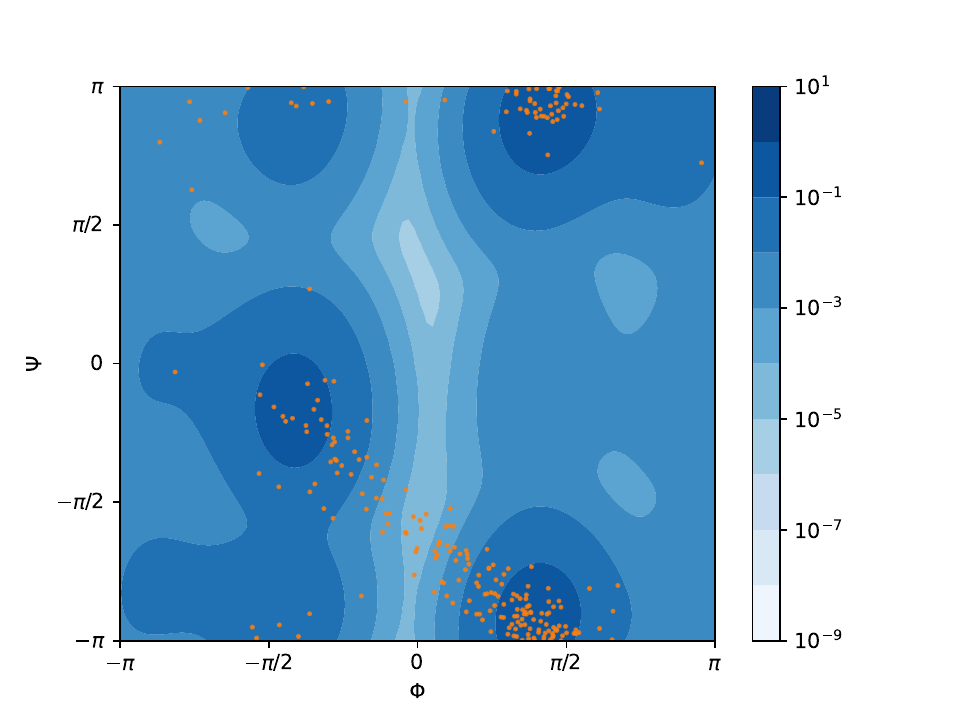}
        \caption{\foldflow,-rand}
        \label{fig:rama2}
    \end{subfigure}
    \begin{subfigure}[b]{0.32\textwidth}
        \includegraphics[width=\textwidth]{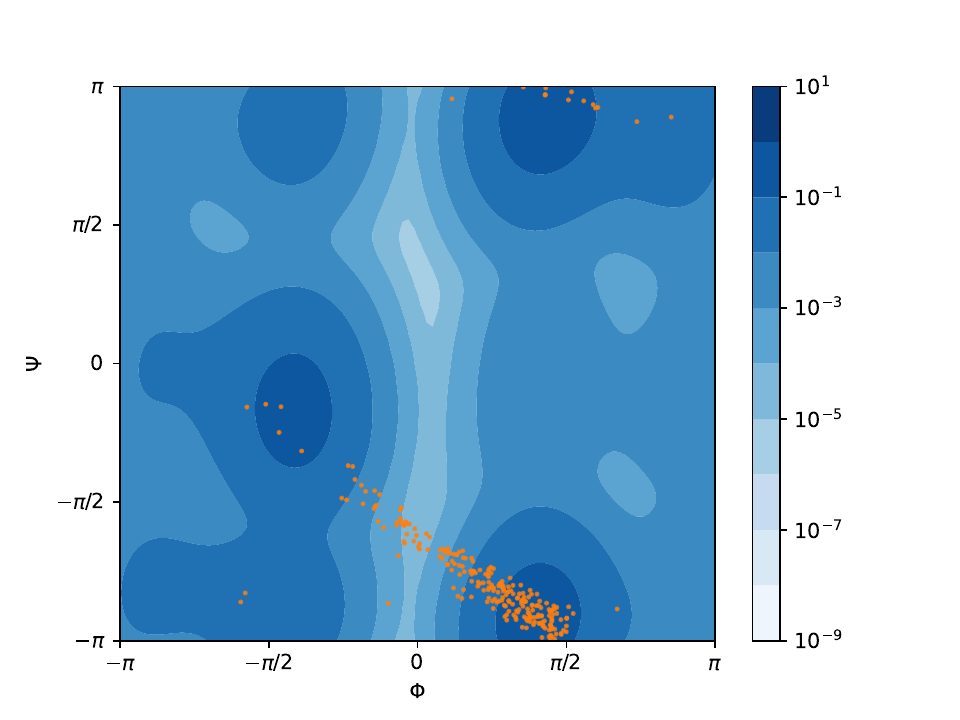}
        \caption{FrameDiff}
        \label{fig:rama3}
    \end{subfigure}
    \hspace{-1pt}
    \vspace{-5pt}
    \caption{\small Ramachandran plot of $\Phi$ and $\Psi$ for (a) \foldflow, with informed prior (b) \foldflow,-rand with uniformed prior and (c) FrameDiff at 10k steps. With the informed prior, \foldflow, is able to better capture the mode on the left.}
    \vspace{-5pt}
    \label{fig:rama}
\end{figure}

\begin{table}[htb]
    \centering
    \caption{\small \rev{Quantitative performance on the equilibrium conformation generation task on the BPTI protein. Measures the 2-Wasserstein $\mathcal{W}_2$ in angle space between generated and test samples over all residues ($\mathcal{W}_2$), the most flexible residue ($\mathcal{W}_2 @ 56$), and the Kullback-Leibler divergence also at the most flexible residue (KL @ 56). We also measure the distance of a distribution of AlphaFold2 structures (AlphaFold2) as well as the distance to samples from the trainset.}}
    \begin{tabular}{lrrr}
    \toprule
    {} &     \rev{$\mathcal{W}_2 (\downarrow)$} &  \rev{$\mathcal{W}_2 @ 56 (\downarrow)$} & \rev{$\text{KL} @ 56 (\downarrow)$} \\
    \midrule
    \rev{\foldflow,}          &  \rev{4.379} &  \rev{0.406} & \rev{0.441}\\
    \rev{\foldflow,-Rand}     &  \rev{4.446} &  \rev{0.557} & \rev{1.205}\\
    \rev{FrameDiff}           &  \rev{4.844} &  \rev{0.800} & \rev{3.051}\\
    \midrule
    \rev{RandomPrior}         &  \rev{18.752}&  \rev{1.993} & \rev{0.746} \\
    \rev{AlphaFold2}          &  \rev{7.298} &  \rev{1.917} & \rev{5.724} \\
    \rev{Trainset}            &  \rev{4.140} &  \rev{0.198} & \rev{0.487}\\
    \bottomrule
    \end{tabular}
    \label{tab:equilibrium_large}
\end{table}

As described in \Cref{exp:eq_conform} proteins take on many different physical conformations in the real world.  These conformations dictate many important attributes of a protein's behaviour, e.g., how one protein might bind to another.  As a protein's conformations generally do not deviate greatly from one another, a desirable approach would be to start from a noised version of a known conformation of the protein to generate another conformation. We hypothesize that the flows required to do this are easier to learn than starting from an uninformed source distribution. We find \foldflow, is an ideal candidate for this setting, and we show its efficacy in fig.~\ref{fig:md_fig}.

We chose bovine pancreatic trypsin inhibitor (BPTI) to study in this experiment. The 58-residue protein is the first protein whose dynamics were studied experimentally by nuclear magnetic resonance (NMR) and was the first protein that was simulated by molecular dynamics for 1ms. On timescales ranging from nanoseconds to milliseconds, the dynamics of BPTI involve protein backbone structural changes that, for example, accommodate water molecule exchange and disulfide isomerization~\citep{persson2008nanosecond}. We used the $1-ms$ MD simulation trajectory at a temperature of 300 K~\citep{shaw2010atomic} to reproduce and interpret the kinetics of folded BPTI. To construct our source distribution, we first generated four folded conformations from each of AlphaFold2, ESMFold, RoseTTAFold, and Unifold~\citep{li2022uni-fold}. Our source distribution was then added a small amount of noise from the standard Gaussian and $\igso$. We trained \foldflow, for one day on 4 A100 GPUs.

Results in \Cref{exp:eq_conform} show that \foldflow, generates conformations covering all modes of the true conformation distribution. Moreover, we sample different conformations of BPTI from AlphaFold2 and plot them on the Ramachandran and ICA plots, observing while \foldflow, can capture all modes of the distribution AlphaFold2 only captures one. Further \cref{fig:app_md_kl} shows that the KL divergence between the distribution of angles generated by \foldflow, is low and uniform, conveying it has learned the distribution of the target. We believe this is an exciting direction meriting larger experiments on more proteins in the future.

\rev{\xhdr{Comparison to Uninformed Prior and FrameDiff} Next we further describe the setup of the comparison experiment in \cref{tab:equilibrium} and \cref{tab:equilibrium_large} which compares \foldflow, with an informed prior, \foldflow, with a uniform random prior (\foldflow,-Rand) and FrameDiff. We measure the 2-Wasserstein ($\mathcal{W}_2$) distance between generated and test samples either on all 58 residues (denoted $\mathcal{W}_2$) or just on residue 56 (denoted $\mathcal{W}_2 @56$), which is the most flexible residue. The $\mathcal{W}_2$ is calculated using the $\sotwo^{\scriptscriptstyle 2N}$ distance on $\Phi$ and $\Psi$. We also compare the generated and test samples over the Ramachandran plot of residue 56 with a Kullback-Leibler (KL) divergence (depicted in \cref{fig:rama}). Here we compute an empirical histogram from samples over a 100 by 100 grid in angle space. We then compute the KL divergence between the smoothed empirical distributions where the minimum of each bin is clipped at $10^{-10}$ for stability.}

\rev{We train each model for 10k steps on 2 A100 GPUs using either a random prior (uniform over rotations and Gaussian translation) or a mixture of Gaussians / $\igso$ distributions using centers defined by the four folded prior conformations and with standard deviation and $\igso$ concentration 0.5. We do not use inference annealing for this experiment. We generate 250 samples from the model and test against 1000 samples from the test set for computational efficiency reasons.}

\rev{To contextualize these results, we compare the performance of these models with various baselines such as 250 samples from a random prior (used as the prior in \foldflow,-Rand and FrameDiff where each residue is sampled from $\mathcal{N}(0, 10)$), 160 conformations sampled from AlphaFold 2, and 250 samples from the training set (Trainset). Results are averaged over 10 seeds for the random prior and the train set. The trainset represents a well-trained model as the Wasserstein distance is not zero even for empirical distributions drawn from the same distribution. The RandomPrior and the AlphaFold2 represent the random and informed priors respectively. All models are significantly better than these two priors, and \foldflow, approaches the performance of samples from the training set.}

\rev{\Cref{fig:rama} depicts the Ramachandran plots for residue 56 with scatter plots for \foldflow, \foldflow,-Rand and FrameDiff against a kernel density estimate (KDE) of the test set. We see that \foldflow, with the informed prior is able to model both modes where \foldflow,-Rand and FrameDiff both focus on the mode in the bottom right, centred at $\Phi=\pi/2$.}

\rev{We have two major findings from this experiment:
\begin{itemize}
    \item An informed prior helps improve performance both overall and on the most flexible residue as seen by comparing the performance of \foldflow, and \foldflow,-Rand in \cref{tab:equilibrium}.
    \item \foldflow, (with both informed and random priors) improve over FrameDiff on this task. 
\end{itemize}
The equilibrium conformation generation task, studied here, is an example of a setting where an informed prior may be useful. Recent work has explored other applications of starting from an informed prior, such as protein docking~\citep{somnath_aligned_2023,stärk2023harmonic}, single-cell~\citep{tong2023improving} and image-to-image translation~\citep{liu_flow_2023}.}

\section{Further Discussion of \foldflow, and Related Models}

\xhdr{Symmetries as an inductive bias for flow matching}
Leveraging symmetries as an inductive bias in deep learning models (for example by data augmentation or design equivariant models) has been shown to improve data efficiency and lead to better generalization. 
In the context of flow matching for proteins, the goal is to learn the vector field generating the flow, which maps an invariant source to an invariant target distribution, guaranteeing the existence of an equivariant vector field \citep{kohler2020equivariant,bose2021equivariant}. Therefore, one way to exploit this symmetry would be to parameterize the vector field with an equivariant network, taking as input the $3\rm D$ coordinates of the protein. Alternatively, since protein backbones can be parametrized by elements of $\sethreen$, we can directly construct the vector field by taking an intrinsic perspective by using charts on the manifold and their coordinate system. In this case, as the vector field lies on the tangent space of $\sethree$ it is equivariant by construction. 

\rev{\xhdr{Comparison between flow matching and diffusion approaches}
While flow matching and diffusion models bear many similarities they also have key differences which we highlight in this appendix.\begin{enumerate}
    \item Flow matching based approaches enjoy the property of transporting any source distribution to any target distribution. This is in contrast to diffusion where one typically needs a Gaussian-like source distribution. 
    \item Flow matching approaches are readily compatible with optimal transport due to the same property of being able to transport and source to any target. Optimal transport which itself has the advantage of providing faster training with a lower variance training objective and reducing the numerical error in inference due to straighter paths. Diffusion models by themselves are not amenable to optimal transport but instead one can do entropic regularized OT. In Euclidean space, this corresponds to a Schrodinger bridge but this is not an optimal transport path is it stochastic.
    \item In general, simulating an ODE is much more efficient than simulating an SDE during inference. Conditional flow-matching and OT-conditional flow matching~\citep{lipman_flow_2022, tong2023improving} both learn ODEs as the learned flow corresponds to a continuous normalizing flow. Diffusion models on the other hand are SDEs and while being more robust to noise in higher dimensions require more challenging inference.
\end{enumerate}}

\xhdr{Comparison to FrameDiff} While our model uses a similar setup to FrameDiff, we introduce a number of improvements that help to stabilize training and improve performance. Indeed our additions lead to improvements on all metrics over the FrameDiff-Improved model released on GitHub which substantially improves on the designability over FrameDiff-ICML. We first recap the improvements made in FrameDiff-Improved over FrameDiff-ICML as detected in the code:
\begin{enumerate}
    \item A bug in the score calculation for rotations means that there is a stop gradient in the rotation score calculation and FrameDiff-ICML is not trained to match the rotation score, which makes its performance quite impressive given this limitation. This bug is fixed in the FrameDiff-Improved model which uses a different score calculation.
    \item The dataloader was switched from sampling uniform over proteins in the dataset, to uniform over clusters, then uniform within clusters. As we explore in \Cref{sec:data_and_sample}, this changes the distribution of proteins but overall increases diversity as there are many similar proteins in a small number of clusters \Cref{fig:dataset_analysis_c}. 
    \item The rotation loss was changed to use a separate axis and angle component to reduce variance in the loss.
\end{enumerate}
While these items improve the performance of FrameDiff, especially in terms of designability, there are still a few potential areas for improvement. 

One area we focus on is the costly loss function of FrameDiff which relies on calculating the pdf of $\igso$ for sampling and computing the score. In the setting used Frame-Diff, the infinite-sum formulation of the density from \cref{eq:isgo3} had to be used, leading to an expensive score loss. 

We also noticed that FrameDiff does not exactly follow theory in that the model is not exactly translation invariant: As mentioned in \cref{sec:se3_flow_matching} in order to obtain $\sethree$ invariant distributions, the center of mass has to be removed. However, in the FrameDiff code, this was only done at inference and not during training. It is unclear to what extent this impacts the performance, as the model remains translation invariant in expectation and during inference.


\section{Implementation Details and Experimental Setup}
\label{app:experiment_setup}

\subsection{Training and Inference}
To describe the precise algorithm for training \foldflow, models over distributions in $\sethreen$. Our starting distribution in $\sethreen$ is $r_1 \sim \mathcal{U}_\sothree$ i.e.\ uniform over rotations and $s_1 \sim \mathcal{N}(0,I)$. After centering (i.e.\ subtracting the mean) this distribution will be uniform over rotations and with translations distributed according to the centered normal $(r_1, s_1^c) \in \sethreenzero$, with $s_1^c \sim \mathcal{N}^c(0,1)$. In \cref{alg:foldflow_training}, we also slightly abuse the notation and denote the output of the rotation part of $v_\theta$ as $v_\theta(t, r_t)$ and similarly the translation part of $v_\theta$ as $v_\theta(t, s_t)$. We do not include separate algorithms for \foldflowbase, and \foldflowot, as they are simple modifications to \foldflowsfm,. If we set $\gamma_r(t) = 0$ and $\gamma_s(t) = 0$, then we recover the \foldflowot, algorithm. If in addition we remove the resampling in lines 4 and 5 then we recover the \foldflowbase, algorithm. 
\begin{algorithm}
\caption{\foldflowsfm, training on $\sethreen$}
\label{alg:foldflow_training}
\begin{algorithmic}[1]
\State \textbf{Input:} Source and target $\rho_1(x_1), \rho_0(x_0)$, flow network $v_\theta$, and diffusion scalings $\gamma_r(t)$, $\gamma_s(t)$.
\While{Training}
    \State $t, x_0, x_1 \sim \gU(0, 1), \rho_0, \rho_1$
    
    \State $\bar{\pi} \leftarrow \text{OT}(x_0, x_1)$ \Comment{OT resampling step to obtain \foldflowot,}
    \State $(r_0, s_0), (r_1, s_1) \sim \bar{\pi}$
    \State $s_0^c, s_1^c \leftarrow s_0 - \frac{1}{N} \sum_i s_0^i,\quad s_1 - \frac{1}{N} \sum_i s_1^i$ 
    \Comment{mean subtract: $(s_0^c, r_0),(s_1^c, r_1) \in \sethreenzero$}
    \State $r_t \leftarrow \exp_{r_0}(t \log_{r_0}(r_1))$ \Comment{geodesic interpolant from~\cref{eq:sampling_r_t}}
    \State $s_t \leftarrow t s_1^c + (1 - t) s_0^c$
    \Comment{interpolant (Euclidean)}
    
    \State $\tilde{r}_t \sim\igso(r_t, \gamma_r^2(t) t(1-t))$ \Comment{simulation-free approximation from \cref{eq:igso_approx}}
    \State $\tilde{s}_t \sim \mathcal{N}(s_t, \gamma_s^2(t) t(1-t))$
    \State $\gL_{\foldflow,} \leftarrow \left \|v_\theta(t, \tilde{r}_t) - \frac{\log_{\tilde{r}_t}(r_0)}{t} \right \|_{\sothree}^2 + \left \| v_\theta(t, \tilde{s}_t) - \frac{\tilde{s}_t - s^c_0}{t} \right\|^2 $
    \State $\theta \leftarrow \text{Update}(\theta, \nabla_\theta \gL_{\foldflow,})$
\EndWhile
\State \textbf{return} $v_\theta$
\end{algorithmic}

\end{algorithm}

\subsection{SDE Training and Inference}\label{app:sec:sde_train_inf}
In this section, we outline our training and inference algorithms for the $\sothree$ component of \foldflowsfm. The training algorithm is detailed in \cref{alg: sfm_training} while the inference algorithm is provided in \cref{alg:sfm_inference}.


\begin{algorithm}
\caption{\foldflowsfm, training on $\sothree$}
\label{alg: sfm_training}
\begin{algorithmic}[1]
\State \textbf{Input:} Source and target $\rho_1, \rho_0$, diffusion schedule $\gamma(\cdot)$, flow network $v_\theta$
\While{Training}
    \State $t, x_0, x_1 \sim \gU(0, 1), \rho_0, \rho_1$
    \State $\bar{\pi} \leftarrow \text{OT}(x_0, x_1)$
    \State $r_0, r_1 \sim \bar{\pi}$
    \State $r_t \leftarrow \exp_{r_0}(t \log_{r_0}(r_1))$
    \Comment{geodesic interpolant from~\cref{eq:sampling_r_t}}
    \State $\tilde{r}_t \sim\igso(r_t, \gamma^2(t) t(1-t))$ \Comment{simulation-free approximation from \cref{eq:igso_approx}}
    \State $u_t(\tilde{r}_t| r_0, r_1) \leftarrow  \frac{\log_{\tilde{r}_t}(r_0)}{t}$
    \State $\gL_{\foldflowsfm,} \leftarrow ||v_\theta(t, r_t) - u_t(\tilde{r}_t|r_0, r_1)||_{\sothree}^2$
    \State $\theta \leftarrow \text{Update}(\theta, \nabla_\theta \gL_{\foldflowsfm,})$
\EndWhile
\State \textbf{return} $v_\theta$
\end{algorithmic}

\end{algorithm}

\begin{algorithm}
\caption{FoldFlow-SFM Inference}
\label{alg:sfm_inference}
\begin{algorithmic}[1]
\State \textbf{Input:} Source distribution \( \rho_1 \), flow network \( v_\theta \), diffusion schedule \( \gamma(\cdot) \), inference annealing \( i(\cdot) \), noise scale, \( \zeta \), integration step size \( \Delta t \).
\State Sample \( r_1 \sim \rho_1 \)
\For{ \( s\) in \( [0, 1 / \Delta t) \)}
    \State $t \gets 1 - s \Delta t$
    \State Sample \( z \sim \mathcal{N}(0, 1) \)
    \State \( dB_t \leftarrow \zeta \gamma_t \cdot \sqrt{dt} \cdot z \)
    \State \( d\hat{B}_t \leftarrow \mathrm{hat}(dB_t)\)
    \Comment {map rotation vector to $\sothreelie$}
    \State \( u_t \leftarrow r_t^\top v_\theta(t, r_t) \)
    \Comment {parallel-transport the vector field to $\sothreelie$}
    \State \( r_{t+\Delta t} \leftarrow r_t \exp(u_t i_t dt + d\hat{B}_t) \)
\EndFor
\State \textbf{return} \( r_0 \)
\end{algorithmic}
\end{algorithm}

\subsection{Vector Field Parametrization}
\label{app:vector_field_parameterization}
Similar to the toy experiment, for the protein modelling case, the architecture is constructed such that the output vector lies on the tangent space.

\subsubsection{Protein Model Parameterization}
\label{app:protein_vfield}

For proteins, we use the FrameDiff architecture~\citep{yim2023se} over $\sethreenzero$, which is based on the structure module of AlphaFold2 \citep{jumper2021highly} following the initial work on diffusion models with AF2-like architectures~\citep{anand2022protein}. As described in the main text, this architecture $w_\theta$ outputs a predicted $\hat{x}_0$, which we can then deterministically transform into a vector located at the tangent space of $x_t$. This transformation can be split into $2N$ components, the $N$ $\R^3$ components, and the $N$, $\sothree$ components. For the $\sothree$ components we calculate
\begin{equation}
    v_\theta(t, r_t) = \frac{\log_{r_t} \hat{r}_0}{t}, \quad 
\end{equation}
and for the $\R^3$ components we calculate after centering,
\begin{equation}
    v_\theta(t, s_t) = \left(\frac{s_t - \hat{s}_0}{t}\right) - \frac{1}{N} \sum_i^N (\frac{s_t - \hat{s}_0}{t})_i,
\end{equation}
where $(\hat{r}_0, \hat{s}_0) = \hat{x}_0 = w_\theta(t, x_t)$. For $\R^3$, $v_\theta(t, s_t)$ is clearly on the tangent space as $\R^3$ is isomorphic to its own tangent space. This is because Euclidean space is a flat space. For $\sothree$, $v_\theta(t, r_t)$ is also on the tangent space of $r_t$ due to the definition of the $\log$ map. Since all components of the product space $\sethreenzero$ are on the tangent space, $v_\theta(t, x_t)$ is on the tangent space of $\sethreenzero$.

\subsection{Protein Task Hyperparameters}
\foldflow, is implemented in Pytorch, and uses the invariant point attention (IPA) implementations from OpenFold~\citep{ahdritz_openfold_2022} in the backbone.  We use the Adam optimizer with constant learning rate $10^{-4}$, $\beta_1 = 0.9$, $\beta_2 = 0.99$. The batch size depends on the length of the protein to maintain roughly constant memory usage. In practice, we set the effective batch size to 
\begin{equation}\label{eq:batch_size}
    \text{eff\_bs} = \max(\text{round}( \# GPUs \times 500,000 / N^2), 1)
\end{equation} 
for each step. We set $\lambda_{aux} = 0.25$ and weight the rotation loss with coefficient $0.5$ as compared to the translation loss which has weight $1.0$. 

We also used a trick from FrameDiff-Improved to stabilize the rotation loss. Instead of the $L^2$ loss on the rotation vector, we separate the loss into two components: one on the axis and one on the angle for the rotation vector. This seemed to reduce variance and numerical instability in the training.

\subsection{Data and Data Sampling}
\label{sec:data_and_sample}

We use a subset of PDB filtered with the same criteria as FrameDiff, specifically, we filter for monomers of length between 60 and 512 (inclusive) with resolution $<5 \angstrom$ downloaded from PDB~\citep{berman_protein_2000} on July 20, 2023. After filtering out any proteins with $>50\%$ loops we are left with 22248 proteins. To support diversity, we sample uniformly over clusters with similarity of 30\% as suggested in FrameDiff-Improved model\footnote{\url{https://cdn.rcsb.org/resources/sequence/clusters/clusters-by-entity-30.txt}}. Our model functions most efficiently with batches of proteins of the same length, so we each batch contains proteins of a single length. There are 4268 clusters in our dataset. 

\begin{figure}[htbp]
    \centering
    \begin{subfigure}[b]{0.32\textwidth}
        \includegraphics[width=\textwidth]{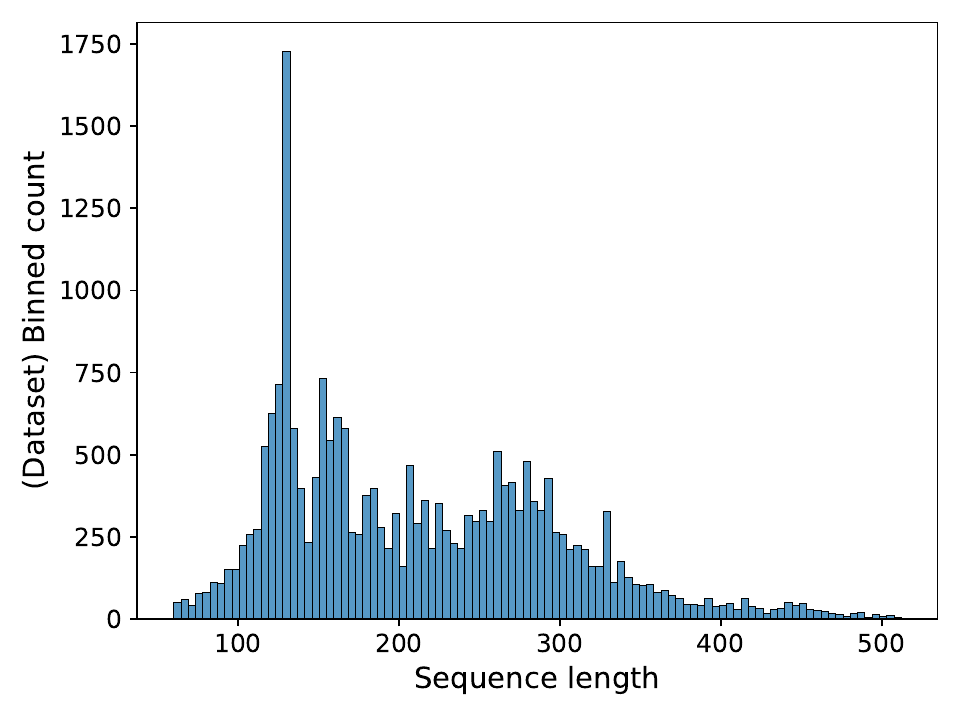}
        \caption{}
        \label{fig:dataset_analysis_a}
    \end{subfigure}
    \hspace{-2pt}
    \begin{subfigure}[b]{0.32\textwidth}
        \includegraphics[width=\textwidth]{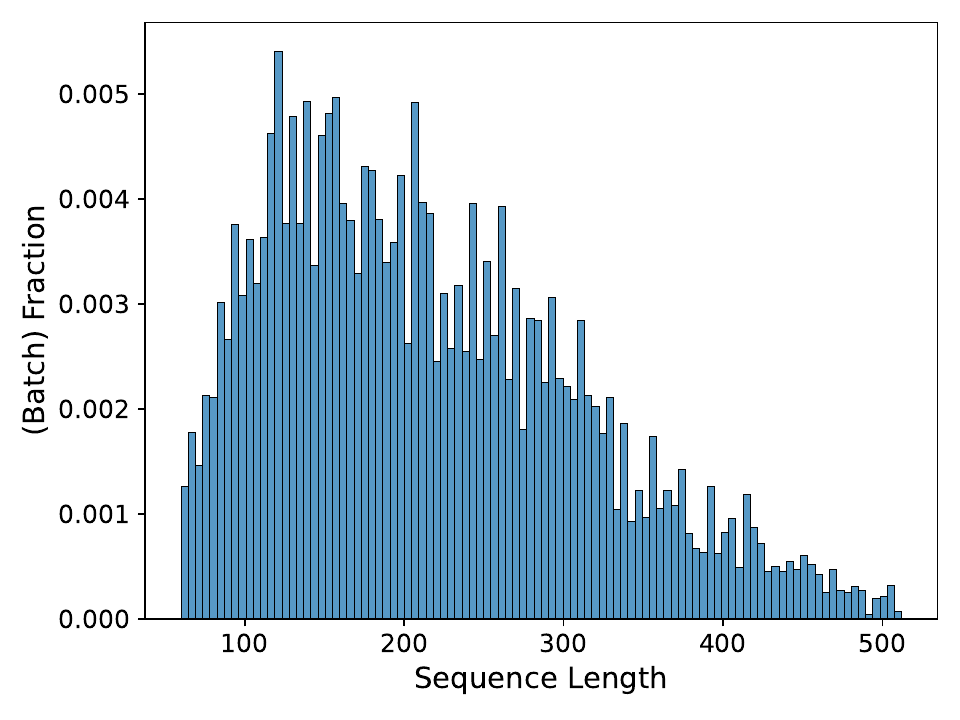}
        \caption{}
        \label{fig:dataset_analysis_b}
    \end{subfigure}
    \begin{subfigure}[b]{0.32\textwidth}
        \includegraphics[width=\textwidth]{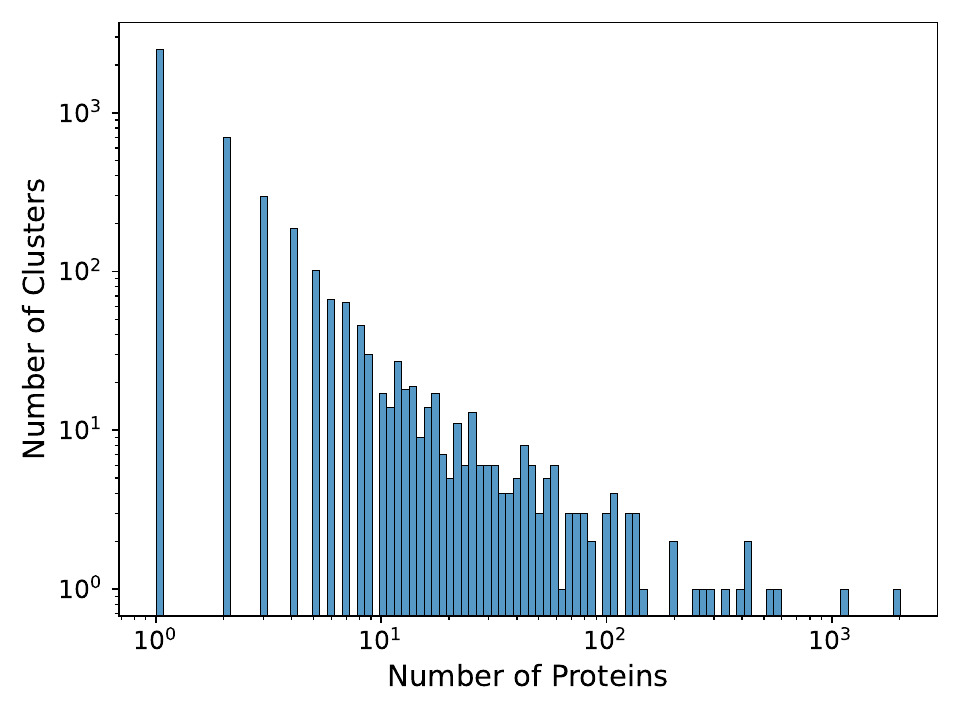}
        \caption{}
        \label{fig:dataset_analysis_c}
    \end{subfigure}
    \caption{\small (a) Distribution of protein lengths in our dataset. (b) Distribution of protein lengths in a batch (when sampling uniformly by cluster) (c)  Distribution of number of proteins per cluster.}
    \label{fig:dataset_analysis}
\end{figure}

To assess the effects of our sampling methods on protein diversity and length distribution, we present three plots. \cref{fig:dataset_analysis_a} illustrates the variability and range of protein lengths in the dataset, giving an overview of available lengths for sampling. \cref{fig:dataset_analysis_b} shows the batch fraction per length, highlighting alterations in sequence length distribution during training due to uniform cluster sampling. Fig.~\ref{fig:dataset_analysis_c}, which uses a log scale on both axes, unveils the variation in cluster sizes and the skewness in protein distribution across clusters. Uniform cluster sampling enhances batch diversity, aiding model generalization over various protein sequences and structures. However, as observed in \cref{fig:dataset_analysis_b}, it slightly modifies the sequence length distribution during training. \cref{fig:dataset_analysis_c} reveals an unevenness in protein distribution across clusters, with two bins containing approximately 14\% of proteins.

\subsection{Protein Metrics}\label{app:protein_metrics}
\xhdr{The TM-score} The template modeling score (TM-score) measures the similarity between two protein structures. The TM score can be expressed for two protein backbones $x_0, x_1 \in \sethreen$ as 
\begin{equation}
    \text{TM-score}(x_0, x_1) = \max \left [ \frac{1}{N_{\rm target}} \sum_i^{N_{\rm common}} \frac{1}{1 + \left ( \frac{ d_i }{d_0(N_{\rm target})}\right )^2} \right]
\end{equation}
where $N_{\rm target}$ is the length of the target sequence, $N_{\rm common}$ is the length of the common sequence after 3D structural alignment, $d_i$ is the distance (post alignment) of the $i^{th}$ residues in $x_0$, and $x_1$, and $d_0(N) = 1.24 (N - 15)^{1/3} - 1.8$ is a scaling factor to normalize across protein lengths. The TM-score ranges between $(0, 1]$ with a TM-score of $1$ indicating perfectly aligned structure. In general a TM-score $>0.5$ are considered roughly similar folds, with TM-score $<$ 0.2 corresponding to randomly chosen unrelated proteins.

\xhdr{The RMSD metric} The root-mean-square deviation (RMSD) is a simple metric over paired residues expressed as
\begin{equation}
    \text{RMSD}(x_0, x_1) = \sqrt{ \sum_{i=1}^L \frac{d_i^2}{L}}
\end{equation}
where $d_i$ is again the distance between the $i^{th}$ residues heavy atoms $[\text{N},\text{C}_{\alpha},\text{C},\text{O}]$. The RMSD score is length dependent unlike the TM-score, but has been shown to be a more stringent filtering step then TM-score $>0.5$ for designability \citep{watson_novo_2023}. In general, as compared to TM-score the RMSD metric is more sensitive local errors and less sensitive to global misalignments. 


\xhdr{Designability} A generated protein structure is considered designable if there exists an amino acid sequence which refolds to that structure. We first generate 50 proteins at lengths \{100, 150, 200, 250, 300 \}, then apply ProteinMPNN with $\text{sampling\_temp} = 0.1$ 8 times to generate 8 sequences for every generated structure. Finally we apply default ESMFold and aligned RMSD of the $C_\alpha$ backbone atoms to calculate alignment of each ESMFold-refolded structure with the generated structure. We determine a protein designable if at least one of the 8 refolded structures has an scRMSD score $< 2.0$. While a threshold of $< 2.0$ for designability is standard, this threshold may be unreasonably strict for longer backbones. However, it is unclear how this threshold should decay with increasing sequence length.

Finally, we note the imperfection of the self-consistency designability metric: when ESMFold does not produce the same structure as \foldflow, it does not imply \foldflow,'s structure is wrong, especially for longer sequences where protein folding models are known to perform worse. Both ProteinMPNN and ESMFold are imperfect, and the failure cases of these models has not been well characterized. While the false positive rate of this metric appears to be low, the false negative of this metric has not been quantified.

\xhdr{Diversity} We calculate all pairwise TM-scores for all generated structures that achieve the designability threshold of scRMSD $<2$ for each length of protein. We then compute the mean over all of these pairwise TM-scores as our diversity metric. For this metric, a lower score is better. We choose to compare diversity on designable proteins as we do not want the designability score to be inflated by models which produce poor, proteins that may be very dissimilar to the space of refoldable proteins at that length.

\xhdr{Novelty} We calculate novelty \rev{using two metrics. The first is the} minimum TM-score of designable generated proteins to the training data as described in \Cref{sec:data_and_sample}. \rev{The second metric is motivated by previous research \citep{lin2023generating} and is the fraction of proteins that are both designable (scRMSD < 2 $\angstrom$) and novel (avg. max TM-score < 0.5). We note that all models are not trained on the same dataset: \foldflow, and FrameDiff-Retrained share their dataset and\foldflow, and FrameDiff-ICML use very similar training datasets (only differing in about 10\% of structures),}. However, Genie and RFdiffusion use substantially larger datasets. Genie is trained on the Swissprot database~\citep{jumper2021highly, varadi2021alphafold} and RFdiffusion is at least pretrained on high-confidence AlphaFold2 structures. These larger training sets may cause novelty to be overestimated for these models as there are structures in their training sets that are far from the training set we use to test novelty against.

\rev{\xhdr{Error bounds in \Cref{tab:main}} We also report the standard error of the novelty and designability metrics in \cref{tab:main}. This is calculated by taking the standard error for each metric per sequence length, and then taking the mean over sequence lengths. We note that as the diversity is calculated as the averaged pairwise distances of designable proteins, each estimate of the mean is correlated resulting in an invalid estimate of the standard error.}

\section{Extended Ablation of experiments}



\begin{table}[H]
  \centering
  \caption{Ablation study of \foldflow, features (stochasticity, optimal transport, auxiliary losses and inference annealing) against designability, diversity and novelty metrics.}
\begin{tabular}{P{2.1cm}P{1.7cm}P{1.7cm}P{1.7cm}|P{0.8cm} P{0.8cm}P{0.8cm}P{1.2cm}}
    \toprule
      \multicolumn{1}{c}{Designability($\uparrow$)}  & \multicolumn{1}{c}{Diversity($\downarrow$)} & \multicolumn{2}{c}{Novelty}& Stochas.  & OT & Aux. Loss & Inf. annealing \\
     \cmidrule(lr){3-4}
     & & \rev{max($\downarrow$)} & \rev{fraction($\uparrow$)} & & & & \\
    \midrule
    \rev{0.228} & \rev{0.230} & \rev{0.440} & \rev{0.172} &  \xmark & \xmark  & \xmark  & \xmark \\
    \rev{0.648} & \rev{0.267} & \rev{0.447} & \rev{0.412} & \xmark & \xmark  & \xmark  & \cmark \\
    \rev{0.132} & \rev{0.235} & \rev{0.432} & \rev{0.096} & \xmark & \xmark  & \cmark  & \xmark \\
    0.657 & 0.264 & 0.452 & \rev{0.432} &  \xmark & \xmark  & \cmark & \cmark \\
    \rev{0.112} & \rev{0.209} & \rev{0.414} & \rev{0.088} & \xmark & \cmark  & \xmark  & \xmark \\
    \rev{0.592} & \rev{0.247} & \rev{0.419}  & \rev{0.424} & \xmark & \cmark  & \xmark  & \cmark \\
    \rev{0.152} & \rev{0.190} & \rev{0.443} & \rev{0.108} & \xmark & \cmark  & \cmark  & \xmark \\
    0.820 & 0.247 & 0.460 & \rev{0.484} & \xmark & \cmark & \cmark & \cmark \\
    \rev{0.128} & \rev{0.198} & \rev{0.394} & \rev{0.100} & \cmark & \xmark  & \xmark  & \xmark \\
    \rev{0.580} & \rev{0.253} & \rev{0.439} & \rev{0.416} & \cmark & \xmark  & \xmark  & \cmark \\
    \rev{0.164} & \rev{0.196} & \rev{0.427} & \rev{0.120} & \cmark & \xmark  & \cmark  & \xmark \\
    \rev{0.684} & \rev{	0.253} & \rev{0.412} & \rev{0.500} & \cmark & \xmark  & \cmark  & \cmark \\
    \rev{0.188} & \rev{0.215} & \rev{0.449} & \rev{0.136} & \cmark & \cmark  & \xmark  & \xmark \\
    \rev{0.632} & \rev{0.257} & \rev{0.433} & \rev{0.432} & \cmark & \cmark  & \xmark  & \cmark \\
    \rev{0.268} & \rev{0.210} & \rev{0.446} & \rev{0.188} & \cmark & \cmark  & \cmark  & \xmark \\
    \rev{0.716} & \rev{0.251} & \rev{0.411} & \rev{0.544} & \cmark & \cmark & \cmark & \cmark \\
    \bottomrule
    \label{tab:full_ablation}
\end{tabular}
\end{table}
\vspace{-30pt}
\rev{We perform a complete ablation experiment for the four features of the \foldflow, models: stochasticity, optimal transport, auxiliary losses in training and inference annealing. For each of the experiments, we evaluate the performance of the model on the designability, diversity and novelty metrics. These results can be seen in \cref{tab:full_ablation}. Overall, we observe the following trends: 
\begin{itemize}
    \item Stochasticity improves robustness and the ability of the model to generate novel proteins in $7/8$ settings, as observed in the fraction of novel and designable proteins (Novelty-fraction).
    \item Optimal transport improves the designability of the model by reducing the variance in the training objective in $7/8$ settings.
    \item The auxiliary losses improve the designability of the models in $7/8$ settings.
    \item Inference annealing improves the performance of all \foldflow, models in all metrics.
\end{itemize}}


\end{document}